\documentclass{article}

\usepackage{times}
\usepackage{graphicx} % more modern
\usepackage{subfigure} 

\usepackage{natbib}

\usepackage{algorithm}
\usepackage{algpseudocode}

\usepackage{amssymb,amsmath}
\usepackage{mathtools}  
\mathtoolsset{showonlyrefs}  
\usepackage{fancyhdr}
\usepackage{titling}
\usepackage{authblk}
\usepackage{subfigure}
\usepackage{multirow}
\usepackage{float}
\usepackage{caption}
\usepackage{natbib}
\usepackage{color}

\usepackage{stmaryrd}

\usepackage{multirow}
\usepackage{booktabs}
\usepackage{array}
\newcolumntype{C}[1]{>{\centering\let\newline\\\arraybackslash\hspace{0pt}}m{#1}}

\newcommand{\dotprod}[2]{\ensuremath{\langle #1 , #2\,\rangle}}

\newcommand\partialfrac[2]{\tfrac{\partial #1}{\partial #2}}

\definecolor{greengray}{rgb}{0.05,0.50,0.15}
\definecolor{gold}{rgb}{0.0,0.7,0.7}
\newcommand{\ERED}[1]{\boldsymbol{\textcolor{red}{#1}}}
\newcommand{\EBLU}[1]{\boldsymbol{\textcolor{blue}{#1}}}
\newcommand{\EGGR}[1]{\boldsymbol{\textcolor{greengray}{#1}}}

\def\RR{\mathbb{R}}

\DeclareMathOperator{\dtw}{\mathbf{dtw}_0}
\DeclareMathOperator{\sdtw}{\mathbf{dtw}_\gamma}
\def\kg{k_{\text{GA}}^\gamma}

\def\Acal{\mathcal{A}}
\def\E{\mathbb{E}}
\def\bx{\mathbf{x}}
\def\by{\mathbf{y}}
\DeclareMathOperator{\mmin}{min}

\DeclareMathOperator{\mming}{\mmin^\gamma}

\def\In{\llbracket n \rrbracket}
\def\Im{\llbracket m \rrbracket}
\usepackage{hyperref}

\usepackage[accepted]{icml2017}
\icmltitlerunning{Soft-DTW: a Differentiable Loss Function for Time-Series}

\begin{document} 

\twocolumn[
\icmltitle{Soft-DTW: a Differentiable Loss Function for Time-Series}



% you can specify symbols, otherwise they are numbered in order
\icmlsetsymbol{equal}{*}

\begin{icmlauthorlist}
\icmlauthor{Marco Cuturi}{ensae}
\icmlauthor{Mathieu Blondel}{ntt}
\end{icmlauthorlist}

\icmlaffiliation{ensae}{CREST, ENSAE, Universit\'e Paris-Saclay, France}
\icmlaffiliation{ntt}{NTT Communication Science Laboratories, Seika-cho, Kyoto, Japan}

\icmlcorrespondingauthor{Marco Cuturi}{marco.cuturi@ensae.fr}
\icmlcorrespondingauthor{Mathieu Blondel}{mathieu@mblondel.org}

\icmlkeywords{dynamic time warping, time series, differentiable loss}

\vskip 0.3in
]



\printAffiliationsAndNotice{}  % leave blank if no need to mention equal contribution
%\footnotetext{hi}

\def\mytablefontsize{\footnotesize}

\begin{abstract} 
We propose in this paper a differentiable learning loss between time series, building upon the celebrated dynamic time warping (DTW) discrepancy.
Unlike the Euclidean distance, DTW can compare time series
of variable size and is robust to shifts or dilatations across the time dimension. To compute DTW, one typically solves a minimal-cost alignment problem between two time series
using dynamic programming. Our work takes advantage of a smoothed formulation
of DTW, called soft-DTW, that computes the soft-minimum of all alignment costs.
We show in this paper that soft-DTW is a \emph{differentiable} loss function,
and that both its value and gradient can be computed with quadratic
time/space complexity (DTW has quadratic time but linear space complexity). We
show that this regularization is particularly well suited to average and cluster
time series under the DTW geometry, a task for which our proposal
significantly outperforms existing baselines~\citep{petitjean2011global}. Next,
we propose to tune the parameters of a machine that outputs time series by
minimizing its fit with ground-truth labels in a soft-DTW sense.

\end{abstract} 

\section{Introduction}\label{introduction}
The goal of supervised learning is to learn a mapping that links
an input to an output objects, using examples of such pairs. This task is noticeably more difficult when the output objects have a structure, \emph{i.e.} when they are not vectors~\citep{4269}. We study here the case where each output object is a \emph{time series}, namely a family of observations indexed by time. While it is tempting to treat time as yet another feature, and handle time series of vectors as the concatenation of all these vectors, several practical issues arise when taking this simplistic approach: Time-indexed phenomena can often be stretched in some areas along the time axis (a word uttered in a slightly slower pace than usual) with no impact on their characteristics; varying sampling conditions may mean they have different lengths; time series may not synchronized.

\textbf{The DTW paradigm.} Generative models for time series are usually built
having the invariances above in mind: Such properties are typically handled through latent variables and/or Markovian assumptions~\citep[Part
I,\S18]{lutkepohl2005new}. A simpler approach, motivated by geometry, lies in
the direct definition of a discrepancy between time series that encodes these
invariances, such as the Dynamic Time Warping
(DTW) score \citep{SakoeChiba71,Sakoe78}. DTW computes the best possible alignment between two time series (the optimal alignment itself can also be of interest, see e.g.~\citealt{garreau2014metric}) of respective length $n$ and $m$ by computing first the $n\times m$ pairwise distance matrix between these points to solve then a dynamic program (DP) using Bellman's recursion with a quadratic $(nm)$ cost.

\begin{figure}[t]
\centering
\includegraphics[width=0.43\textwidth]{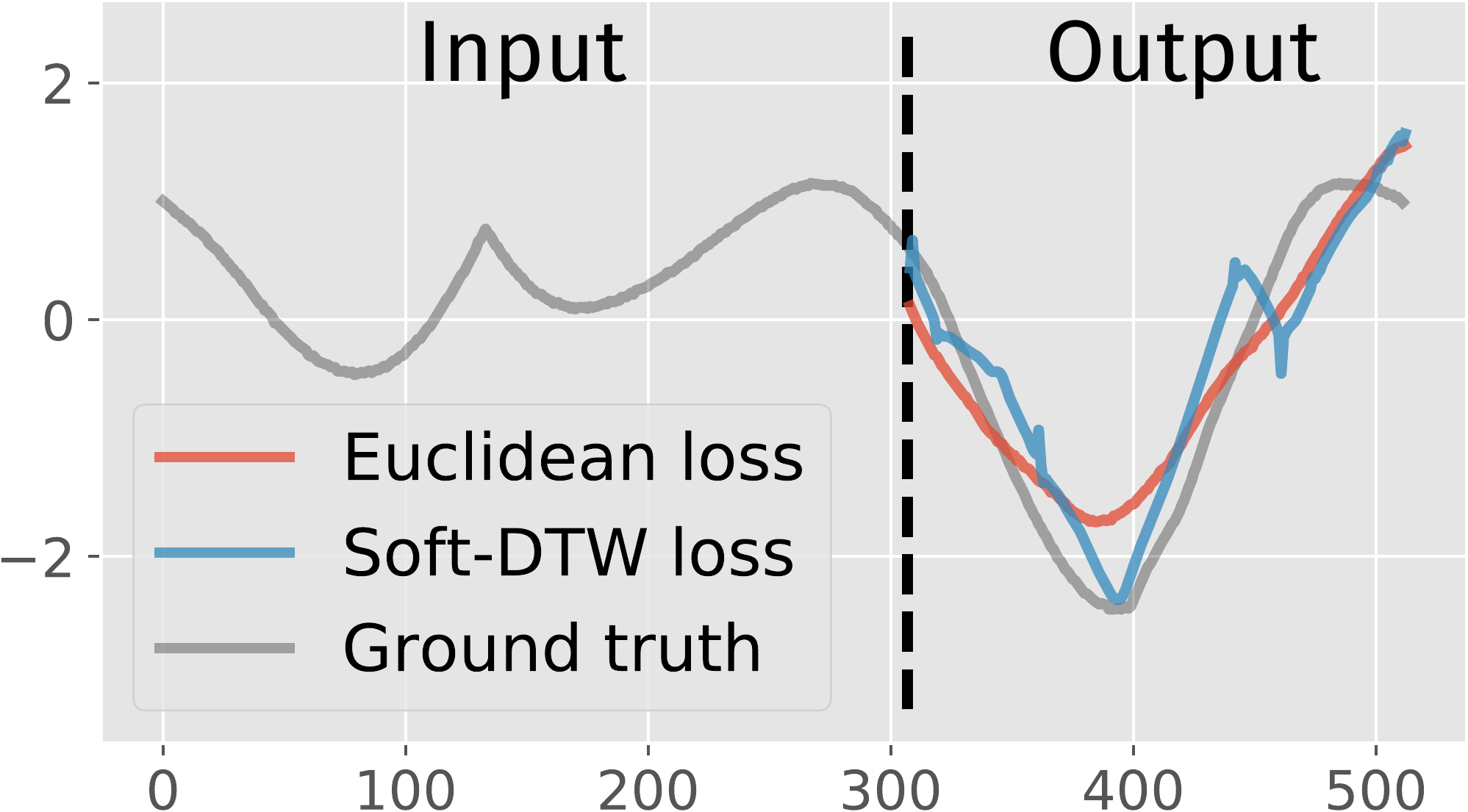}
\caption{Given the first part of a time series, we trained two multi-layer
perceptron (MLP) to predict the entire second part. Using the ShapesAll dataset, we used a Euclidean loss for the first MLP and the soft-DTW loss proposed in this paper for the second one. We display above the prediction obtained for a given test instance with either of these two MLPs in addition to the ground truth. Oftentimes, we observe that the soft-DTW loss enables us
to better predict sharp changes. More time series
predictions are given in Appendix \ref{appendix:more_ts_pred}.}
\label{fig:ts_pred_example}
\end{figure}

\textbf{The DTW geometry.} Because it encodes efficiently a useful class of
invariances, DTW has often been used in a discriminative framework (with a $k$-NN or SVM classifier) to predict a real or a class label output, and engineered to run faster in that context~\cite{yi1998efficient}. Recent works by \citet{petitjean2011global,petitjean2012summarizing} have, however, shown that DTW can be used for more innovative tasks, such as time series \emph{averaging} using the DTW discrepancy
(see~\citealt{schultz2017nonsmooth} for a gentle introduction to these ideas). More generally, the idea of synthetising time series centroids can be regarded as a first attempt to \emph{output} entire time series using DTW as a fitting loss. From a computational perspective, these approaches are, however, hampered by the fact that DTW is not differentiable and unstable when used in an optimization pipeline.

\textbf{Soft-DTW.} In parallel to these developments, several authors have considered smoothed modifications of Bellman's recursion to define smoothed DP distances~\citep{bahl1975decoding,ristad1998learning} or kernels \citep{saigo2004protein,cuturi_icassp}. When applied to the DTW discrepancy, that regularization results in a \emph{soft-DTW} score, which considers the \emph{soft-minimum} of the distribution of \emph{all costs} spanned by \emph{all} possible alignments between two time series. Despite considering all alignments and not just the optimal one, soft-DTW can be computed with a minor modification of Bellman's recursion, in which all $(\min,+)$ operations are replaced with $(+,\times)$. As a result, both DTW  and soft-DTW have quadratic in time \& linear in space complexity with respect to the sequences' lengths. Because soft-DTW can be used with kernel machines, one typically observes an increase in performance when using soft-DTW over DTW~\citep{cuturi_icml} for classification.
% PAS DE PLACE. Because evaluating a log-sum-exp operator is more costly than simply computing a minimum of values, DTW is usually a constant times faster that soft-DTW, giving DTW a small edge in terms of speed. However, this slower execution is often balanced, in practice, by the fact that soft-DTW typically outperforms DTW when cast as a kernel and paired with a kernel machine~\citep{cuturi_icml}

\textbf{Our contributions.} We explore in this paper another important benefit
of smoothing DTW: unlike the original DTW discrepancy, soft-DTW is
\emph{differentiable} in all of its arguments. We show that the gradients of
soft-DTW w.r.t to all of its variables can be computed as a by-product of the
computation of the discrepancy itself, with an added quadratic storage cost. We
use this fact to propose an alternative approach to the DBA (DTW Barycenter
Averaging) clustering algorithm of~\citep{petitjean2011global}, and observe that our smoothed approach significantly outperforms known baselines for that task. More generally, we propose to use soft-DTW as a \emph{fitting term} to compare the output of a machine synthesizing a time series segment with a ground truth observation, in the same way that, for instance, a regularized Wasserstein distance was used to compute barycenters~\cite{cuturi2014fast}, and later to fit discriminators that output histograms~\cite{zhang2015learning,rolet2016fast}. When paired with a flexible learning architecture such as a neural network, soft-DTW allows for a differentiable end-to-end approach to design predictive and generative models for time series, as illustrated in Figure \ref{fig:ts_pred_example}. Source code is available at \url{https://github.com/mblondel/soft-dtw}.

\textbf{Structure.} After providing background material, we show in
\S\ref{sec:techcontrib} how soft-DTW can be differentiated w.r.t the locations
of two time series. We follow in \S\ref{sec:learning} by illustrating how these
results can be directly used for tasks that require to output time series:
averaging, clustering and prediction of time series.  We close
this paper with experimental results in \S\ref{sec:experiments} that showcase each of these potential applications.

\textbf{Notations.} We consider in what follows multivariate discrete time
series of varying length taking values in $\Omega\subset\RR^p$. A time series
can be thus represented as a matrix of $p$ lines and varying number of columns.
We consider a differentiable substitution-cost function
$\delta:\RR^p\times\RR^p\rightarrow \RR_+$ which will be, in most cases, the
quadratic Euclidean distance between two vectors. For an integer $n$ we write
$\In$ for the set $\{1,\dots,n\}$ of integers. Given two series' lengths $n$ and
$m$, we write $\Acal_{n,m} \subset \{0,1\}^{n \times m}$ for the set of
(binary) alignment matrices, that is paths on a $n\times m$ matrix that connect the
upper-left $(1,1)$ matrix entry to the lower-right $(n,m)$ one using only
$\downarrow,\rightarrow,\searrow$ moves. 
% PAS DE PLACE This corresponds to the set of $n\times m$ binary matrices $A$ with the following constraints: $\{A|A_{11}=A_{nm}=1, A_{ij}\leq (A_{i+1j} \oplus A_{ij+1}) \vee A_{i+1j+1}, i\in\In, j\in\Im\}$ \MC{To check!} where $\oplus$ is the XOR operator. 
The cardinal of $\Acal_{n,m}$ is known as the $\text{delannoy}(n-1,m-1)$ number;
that number grows exponentially with $m$ and $n$. 
%Let $\bar{\mathcal{A}}(n,m)
%\subset [0,1]^{n \times m}$ be the convex hull of $\mathcal{A}(n,m)$.

\section{The DTW and soft-DTW loss functions}\label{sec:techcontrib}

We propose in this section a unified formulation for the original DTW
discrepancy~\citep{Sakoe78} and the Global Alignment kernel
(GAK)~\citep{cuturi_icassp}, which can be both used to compare two time series
$\bx=(x_1,\dots,x_n)\in\RR^{p\times n}$ and $\by=(y_1,\dots,y_m)\in\RR^{p\times
m}$.

\subsection{Alignment costs: optimality and sum}

\begin{figure}[t]
    \centering
	\includegraphics[width=.45\textwidth]{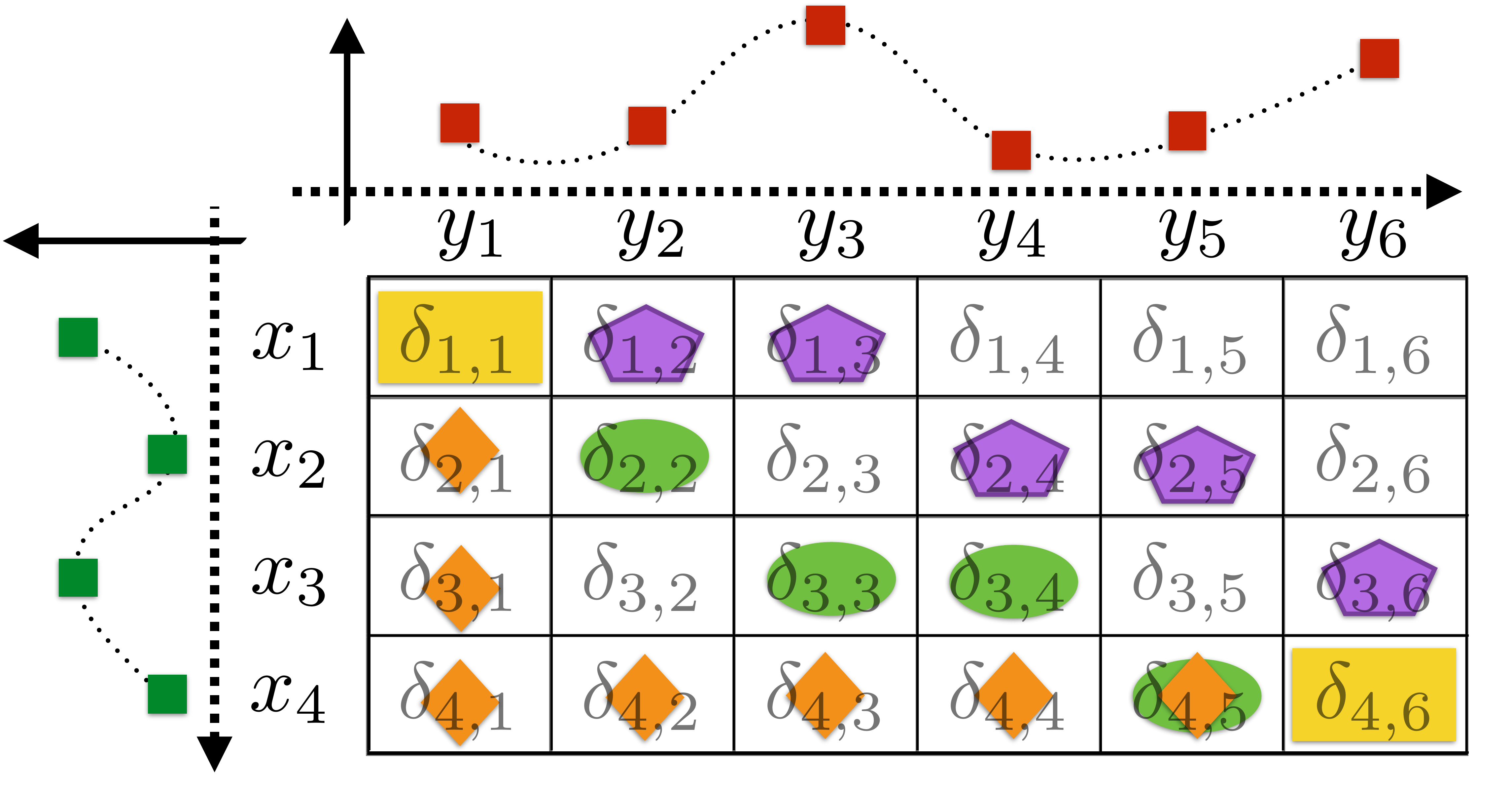}
	\caption{Three alignment matrices (orange, green, purple, in addition to the
    top-left and bottom-right entries) between two time series of length 4 and
6. The cost of an alignment is equal to the sum of entries visited along the
path. DTW only considers the optimal alignment (here depicted in purple
pentagons), whereas soft-DTW considers all $\text{delannoy}(n-1,m-1)$ possible alignment matrices.}\label{fig:delannoy}
\end{figure}

Given the cost matrix $\Delta(\bx,\by)\coloneqq \begin{bmatrix}
\delta(x_i,y_j)\end{bmatrix}_{ij}\in\RR^{n\times m}$, the inner product $\dotprod{A}{\Delta(\bx,\by)}$ of that matrix with an alignment matrix $A$ in $\Acal_{n,m}$ gives the score of $A$, as illustrated in Figure~\ref{fig:delannoy}. Both DTW and GAK consider the costs of all possible alignment matrices, yet do so differently:
\begin{equation}\begin{aligned}\text{DTW}(\bx,\by) &\coloneqq \min_{A\in\Acal_{n,m}}\dotprod{A}{\Delta(\bx,\by)},\\
\kg(\bx,\by)&\coloneqq\sum_{A\in\Acal_{n,m}}
e^{-\dotprod{A}{\Delta(\bx,\by)}/\gamma}.\end{aligned}\label{eq:sdtw}\end{equation}

%Note that $\lim_{\gamma\rightarrow 0}\sdtw(\bx,\by) =\dtw(\bx,\by)$.  
\textbf{DP Recursion.} \citet{Sakoe78} showed that the \citeauthor{bellman1952theory} equation \citeyearpar{bellman1952theory} can be used to compute $\text{DTW}$. That recursion, which appears in line~\ref{lst:line} of Algorithm~\ref{algo:gendtw} (disregarding for now the
exponent $\gamma$), only involves $(\min,+)$ operations. When considering kernel $\kg$ and, instead, its integration over all alignments (see e.g.~\citealt{lasserre2009linear}), \citet[Theorem 2]{cuturi_icassp} and the highly related formulation of \citet[p.1685]{saigo2004protein} use an old algorithmic appraoch~\citep{bahl1975decoding} which consists in \emph{(i)} replacing all costs by their neg-exponential; \emph{(ii)}  replace $(\min,+)$ operations with $(+,\times)$ operations. These two recursions can be in fact unified with the use of a soft-minimum operator, which we present below.

\textbf{Unified algorithm} Both formulas in Eq.~\eqref{eq:sdtw} can be computed with a single algorithm. That formulation is new to our knowledge. Consider the following generalized $\min$ operator, with a smoothing parameter $\gamma\geq 0$:
$$
\mming \{a_1,\dots,a_n\} \coloneqq \begin{cases} \min_{i\leq n} a_i, & \gamma=0,\\ -\gamma\log\sum_{i=1}^n e^{-a_i/\gamma},& \gamma>0.\end{cases}
$$
With that operator, we can define $\gamma$-soft-DTW:
\begin{equation}\label{eq:dtw}
    \boxed{\sdtw(\bx,\by) \coloneqq \mming \{\dotprod{A}{\Delta(\bx,\by)}, A \in \Acal_{n,m}\}.}
\end{equation}
The original DTW score is recovered by setting $\gamma$ to $0$. When $\gamma>0$, we recover $\sdtw=-\gamma \log \kg$. Most importantly, and in either case, $\sdtw$ can
be computed using Algorithm~\ref{algo:gendtw}, which requires $(nm)$ operations and $(nm)$ storage cost as well . That cost can be reduced to $2n$ with a more careful implementation if one only seeks to compute $\sdtw(\bx,\by)$, but the backward pass we consider next requires the entire matrix $R$ of intermediary alignment costs. Note that, to ensure numerical stability, the operator $\mming$ must be computed using the usual log-sum-exp stabilization trick, namely that
%\begin{equation}
$
\log \sum_i e^{z_i} = (\max_j z_j) + \log \sum_i e^{z_i - \max_j z_j}.
$
%\end{equation}

\begin{algorithm}[h]
	\caption{Forward recursion to compute $\sdtw(\bx,\by)$ and intermediate alignment costs\label{algo:gendtw}} 
	\begin{algorithmic}[1]
		\State \textbf{Inputs}: $\mathbf{x,y}$, smoothing $\gamma\geq 0$, distance function $\delta$
		\State $r_{0,0}=0; r_{i,0}=r_{0,j}=\infty; i\in\In, j\in\Im$
        \For{$j=1,\dots,m$} 
				\For{$i=1,\dots,n$}
				\State $\!\!\!\!\!\!r_{i,j}= \delta(x_i,y_j) +\mming\{r_{i-1,j-1},r_{i-1,j},r_{i,j-1}\}$\label{lst:line}
				\EndFor
		\EndFor
        \State \textbf{Output:} $(r_{n,m}, R)$
	\end{algorithmic}
\end{algorithm}

\subsection{Differentiation of soft-DTW} 
\label{subsec:differentiation}

A small variation in the input $\bx$
causes a small change in $\dtw(\bx,\by)$ or $\sdtw(\bx,\by)$. When considering
$\dtw$, that change can be efficiently monitored only when the optimal alignment
matrix $A^\star$ that arises when computing $\dtw(\bx,\by)$ in
Eq.~\eqref{eq:sdtw} is unique. As the minimum over a finite set of
linear functions of $\Delta$, $\dtw$ is therefore locally differentiable w.r.t. the cost matrix $\Delta$, with gradient $A^\star$, a fact that has been exploited in all algorithms designed to average time series under the DTW metric~\citep{petitjean2011global,schultz2017nonsmooth}.
To recover the gradient of $\dtw(\bx,\by)$ w.r.t. $\bx$, we only need to apply the chain rule, thanks to the differentiability of the cost function:

\begin{equation}\label{eq:nablanondiff}\nabla_\bx{\dtw}(\bx,\by) =  \left(\frac{\partial \Delta(\bx,\by)}{\partial \bx}\right)^T A^\star,\end{equation}

where $\partial \Delta(\bx,\by)/\partial \bx$ is the Jacobian of $\Delta$ w.r.t. $\bx$, a linear map from $\RR^{p\times n}$ to $\RR^{n\times m}$. When $\delta$ is the squared Euclidean distance, 
the transpose of that Jacobian applied to a matrix $B\in\RR^{n\times m}$ is ($\circ$ being the elementwise product):
$$(\partial \Delta(\bx,\by)/\partial \bx)^T B = 2 \left((\mathbf{1}_p\mathbf{1}_m^T B^T)\circ \bx - \by B^T\right).$$

With continuous data, $A^\star$ is almost always likely to be unique, and therefore the gradient in Eq.~\eqref{eq:nablanondiff} will be defined almost everywhere. However, that gradient, when it exists, will be discontinuous around those values $\bx$ where a small change in $\bx$ causes a change in $A^\star$, which is likely to hamper the performance of gradient descent methods. 

\paragraph{The case $\gamma>0$.} An immediate advantage of soft-DTW is that it can be explicitly differentiated, a fact that was also noticed by ~\citet{saigo2006optimizing} in the related case of edit distances.
When $\gamma>0$, the gradient of Eq.~\eqref{eq:sdtw} is obtained via the chain rule,
\begin{equation}\label{eq:nablasdtw}\nabla_\bx \sdtw (\bx,\by)=  \left(\frac{\partial \Delta(\bx,\by)}{\partial \bx}\right)^T \E_\gamma[A],
\end{equation}
$$\text{where} \quad \E_\gamma[A] \coloneqq
\frac{1}{\kg(\bx,\by)}\sum_{A\in\Acal_{n,m}}e^{-\dotprod{A}{\Delta(\bx,\by)/\gamma}}
A,$$ is the average alignment matrix $A$ under the Gibbs distribution $p_\gamma\propto e^{-\dotprod{A}{\Delta(\bx,\by)}/\gamma}$ defined on all alignments in $\Acal_{n,m}$. The kernel $\kg(\bx,\by)$ can thus be interpreted as the normalization constant of
$p_\gamma$. Of course, since $\mathcal{A}_{n,m}$ has exponential size in $n$
and $m$, a naive summation is not tractable. Although a Bellman recursion to compute that
average alignment matrix $\E_\gamma[A]$ exists (see Appendix
\ref{appendix:forward_recursion}) that computation has \emph{quartic}
($n^2m^2$) complexity. Note that this stands in stark contrast to the quadratic complexity obtained by~\citet{saigo2006optimizing} for edit-distances, which is due to the fact the sequences they consider can only take values in a \emph{finite} alphabet.
To compute the gradient of soft-DTW, we propose instead an algorithm that manages to remain \emph{quadratic}
($nm$) in terms of complexity. The key to achieve this reduction is to apply
the chain rule in \emph{reverse} order of Bellman's recursion given in
Algorithm~\ref{algo:gendtw}, namely backpropagate.  A similar idea was recently used to compute the
gradient of ANOVA kernels in \citep{hofm}.

\subsection{Algorithmic differentiation} 

\begin{figure*}[t]
    \centering
	\includegraphics[width=0.9\textwidth]{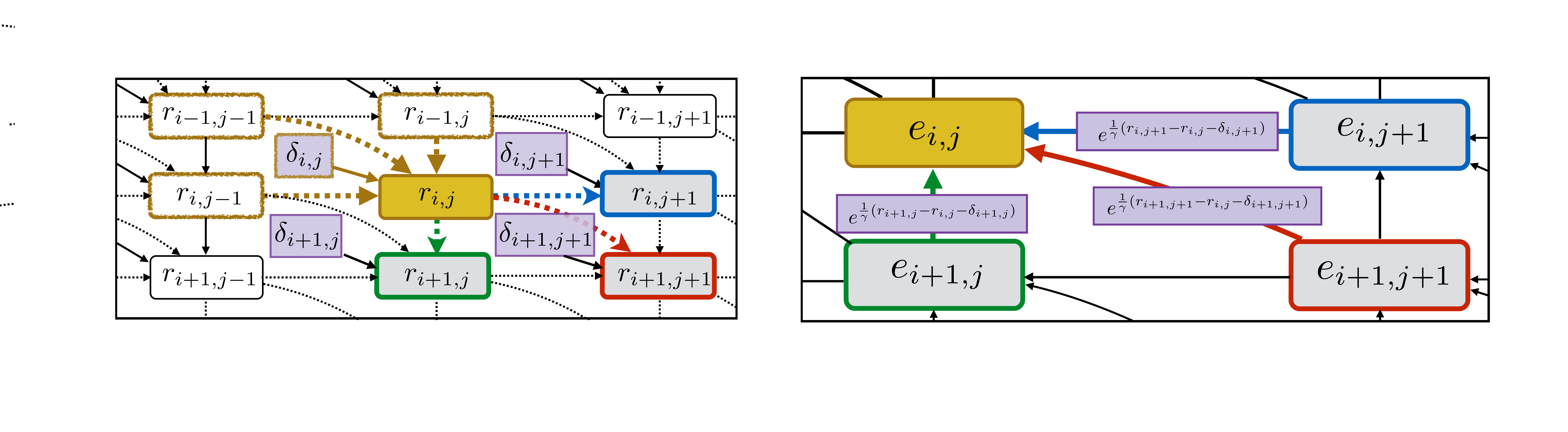}
    \caption{Sketch of the computational graph for soft-DTW, in the forward pass
    used to compute $\sdtw$ (left) and backward pass used to compute its
gradient $\nabla_\bx \sdtw$ (right). In both diagrams, purple shaded cells stand
for data values available before the recursion starts, namely cost values (left)
    and multipliers computed using forward pass results (right). In the left
diagram, the forward computation of $r_{i,j}$ as a function of its predecessors
and $\delta_{i,j}$ is summarized with arrows. Dotted lines indicate a $\mming$
operation, solid lines an addition. From the perspective of the final term
$r_{n,m}$, which stores $\sdtw(\bx,\by)$ at the lower right corner (not shown)
of the computational graph, a change in $r_{i,j}$ only impacts $r_{n,m}$ through
changes that $r_{i,j}$ causes to $r_{i+1,j}$, $r_{i,j+1}$ and $r_{i+1,j+1}$.
These changes can be tracked using Eq.~(\ref{eq:trick},\ref{eq:trick2}) and
appear in lines \ref{lst:a}-\ref{lst:c} in Algorithm~\ref{algo:gendtw2} as
variables $a,b,c$, as well as in the purple shaded boxes in the backward pass
(right) which represents the recursion of line~\ref{lst:rec} in
Algorithm~\ref{algo:gendtw2}.  }
\end{figure*}

Differentiating algorithmically $\sdtw(\bx,\by)$ requires doing first a forward pass of
Bellman's equation to store all intermediary computations and recover $R=[r_{i,j}]$ when running Algorithm~\ref{algo:gendtw}. The value of $\sdtw(\bx,\by)$---stored
in $r_{n,m}$ at the end of the forward recursion---is then impacted by a change in
$r_{i,j}$ exclusively through the terms in which $r_{i,j}$ plays a role, namely
the triplet of terms $r_{i+1,j},r_{i,j+1},r_{i+1,j+1}$. A straightforward application of the chain rule then gives
\begin{equation}\label{}
\!\!\underbrace{\partialfrac{r_{n,m}}{r_{i,j}}}_{e_{i,j}} \!\!=
\underbrace{\partialfrac{r_{n,m}}{r_{i+1,j}}}_{e_{i+1,j}} \!\EGGR{\partialfrac{r_{i+1,j}}{r_{i,j}}}
+\underbrace{\partialfrac{r_{n,m}}{r_{i,j+1}}}_{e_{i,j+1}}\! \EBLU{\partialfrac{r_{i,j+1}}{r_{i,j}}}
+\underbrace{\partialfrac{r_{n,m}}{r_{i+1,j+1}}}_{e_{i+1,j+1}}\! \ERED{\partialfrac{r_{i+1,j+1}}{r_{i,j}}},
\end{equation}
in which we have defined the notation of the main object of interest of the backward recursion: $e_{i,j} \coloneqq \partialfrac{r_{n,m}}{r_{i,j}}$. The Bellman recursion evaluated at $(i+1,j)$ as shown in line~\ref{lst:line} of
Algorithm~\ref{algo:gendtw} (here $\delta_{i+1,j}$ is $\delta(x_{i+1},y_j)$) yields :
\begin{equation}
r_{i+1,j}= \delta_{i+1,j} +\mming\{r_{i,j-1},r_{i,j},r_{i+1,j-1}\},
\end{equation}
which, when differentiated w.r.t $r_{i,j}$ yields the ratio:
$$
\EGGR{\partialfrac{r_{i+1,j}}{r_{i,j}}} = e^{-r_{i,j}/\gamma} / \left(e^{-r_{i,
j-1}/\gamma} +e^{-r_{i,j}/\gamma} + e^{-r_{i+1,j-1}/\gamma}\right).
$$
The logarithm of that derivative can be conveniently cast using evaluations of $\mming$ computed in the forward loop:
\begin{equation}\label{eq:trick}\begin{aligned}
\gamma \log \EGGR{\partialfrac{r_{i+1,j}}{r_{i,j}}}&= \mming\{r_{i,j-1},r_{i,j},r_{i+1,j-1}\}-r_{i,j}\\
&= r_{i+1,j}-\delta_{i+1,j}-r_{i,j}.\end{aligned}\end{equation}
Similarly, the following relationships can also be obtained:
\begin{equation}\label{eq:trick2}\begin{aligned}
\gamma \log \EBLU{\partialfrac{r_{i,j+1}}{r_{i,j}}}&=r_{i,j+1}-r_{i,j}-\delta_{i,j+1},\\
\gamma \log \ERED{\partialfrac{r_{i+1,j+1}}{r_{i,j}}}&= r_{i+1,j+1}-r_{i,j}-\delta_{i+1,j+1}.	
\end{aligned}
\end{equation}
We have therefore obtained a \textit{backward} recursion to compute the entire matrix
$E=[e_{i,j}]$, starting from $e_{n,m} = \partialfrac{r_{n,m}}{r_{n,m}}=1$ down
to $e_{1,1}$. To obtain $\nabla_\bx \sdtw(\bx,\by)$, notice that the derivatives w.r.t. the
entries of the cost matrix $\Delta$ can be computed by
$
\partialfrac{r_{n,m}}{\delta_{i,j}} =
\partialfrac{r_{n,m}}{r_{i,j}}
\partialfrac{r_{i,j}}{\delta_{i,j}} = e_{i,j} \cdot 1 = e_{i,j},
$
and therefore we have that
$$
\nabla_\bx \sdtw(\bx,\by) = \left(\frac{\partial \Delta(\bx,\by)}{\partial
\bx}\right)^T E,
$$
where $E$ is exactly the average alignment $\E_\gamma[A]$ in
Eq.~\eqref{eq:nablasdtw}. These computations are summarized in
Algorithm~\ref{algo:gendtw2}, which, once $\Delta$ has been computed, has complexity
$nm$ in time and space. Because $\mming$ has a $1/\gamma$-Lipschitz continuous gradient, the gradient of $\sdtw$ is $2/\gamma$-Lipschitz continuous when $\delta$ is the
squared Euclidean distance.

\begin{algorithm}[h]
	\caption{\small Backward recursion to compute $\nabla_\bx \sdtw(\bx,\by)$\label{algo:gendtw2}} 
	\begin{algorithmic}[1]		
		\State \textbf{Inputs}: $\bx,\by$, smoothing $\gamma \ge 0$, distance
        function $\delta$
		\State $(\cdot,R)=\sdtw(\bx,\by)$, $\Delta= [\delta(x_i,y_j)]_{i,j}$ 
		\State $\delta_{i,m+1}=\delta_{n+1,j}=0, i\in\In, j\in\Im$
        \State $e_{i,m+1}=e_{n+1,j}=0, i\in\In, j\in\Im$
		\State $r_{i,m+1}=r_{n+1,j}=-\infty, i\in\In, j\in\Im$
		\State $\delta_{n+1,m+1}=0, e_{n+1,m+1}=1, r_{n+1,m+1}=r_{n,m}$
        \For{$j=m,\dots,1$} 
				\For{$i=n,\dots,1$}
				\State $\!\!\!\!\!\!a= \exp \frac{1}{\gamma}(r_{i+1,j}-r_{i,j}-\delta_{i+1,j})$	\label{lst:a}			
				\State $\!\!\!\!\!\!b= \exp \frac{1}{\gamma}(r_{i,j+1}-r_{i,j}-\delta_{i,j+1})$\label{lst:b}
				\State $\!\!\!\!\!\!c= \exp \frac{1}{\gamma}(r_{i+1,j+1}-r_{i,j}-\delta_{i+1,j+1})$\label{lst:c}
				\State $\!\!\!\!\!\!e_{i,j}= e_{i+1,j}\cdot a+ e_{i,j+1} \cdot b+ e_{i+1,j+1}\cdot c$\label{lst:rec}
				\EndFor
		\EndFor
						
		\State \textbf{Output:}
		$\nabla_\bx \sdtw(\bx,\by)= \left(\frac{\partial \Delta(\bx,\by)}{\partial \bx}\right)^T E$
	\end{algorithmic}
\end{algorithm}

% 
% \For{$j=m,\dots,1$} \COMMENT{Backward pass}.
% 		\For{$i=n,\dots,1$}
% 		\State $d_{i,j}=e^{-D(x_i,y_j)} \left(d_{i-1,j-1}+ d_{i-1,j}+d_{i,j-1}\right)$.
% 		\EndFor
% \EndFor

\section{Learning with the soft-DTW loss}
\label{sec:learning}

\subsection{Averaging with the soft-DTW geometry}
\label{subsec:averaging}

\begin{figure}[t]
   \centering 
\subfigure[Euclidean loss]{
   \includegraphics[scale=0.28]{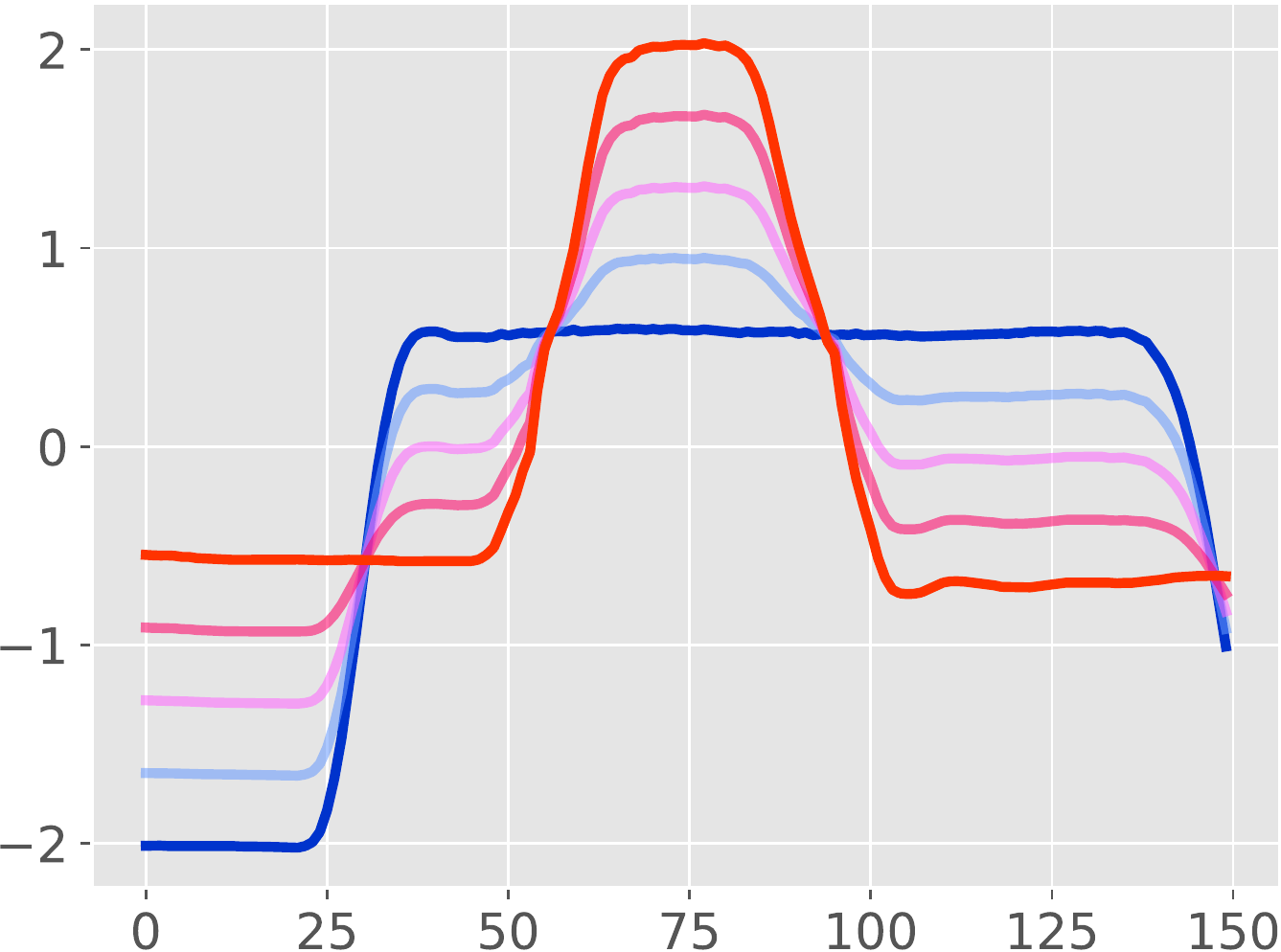}
}
\subfigure[Soft-DTW loss ($\gamma=1$)]{
   \includegraphics[scale=0.28]{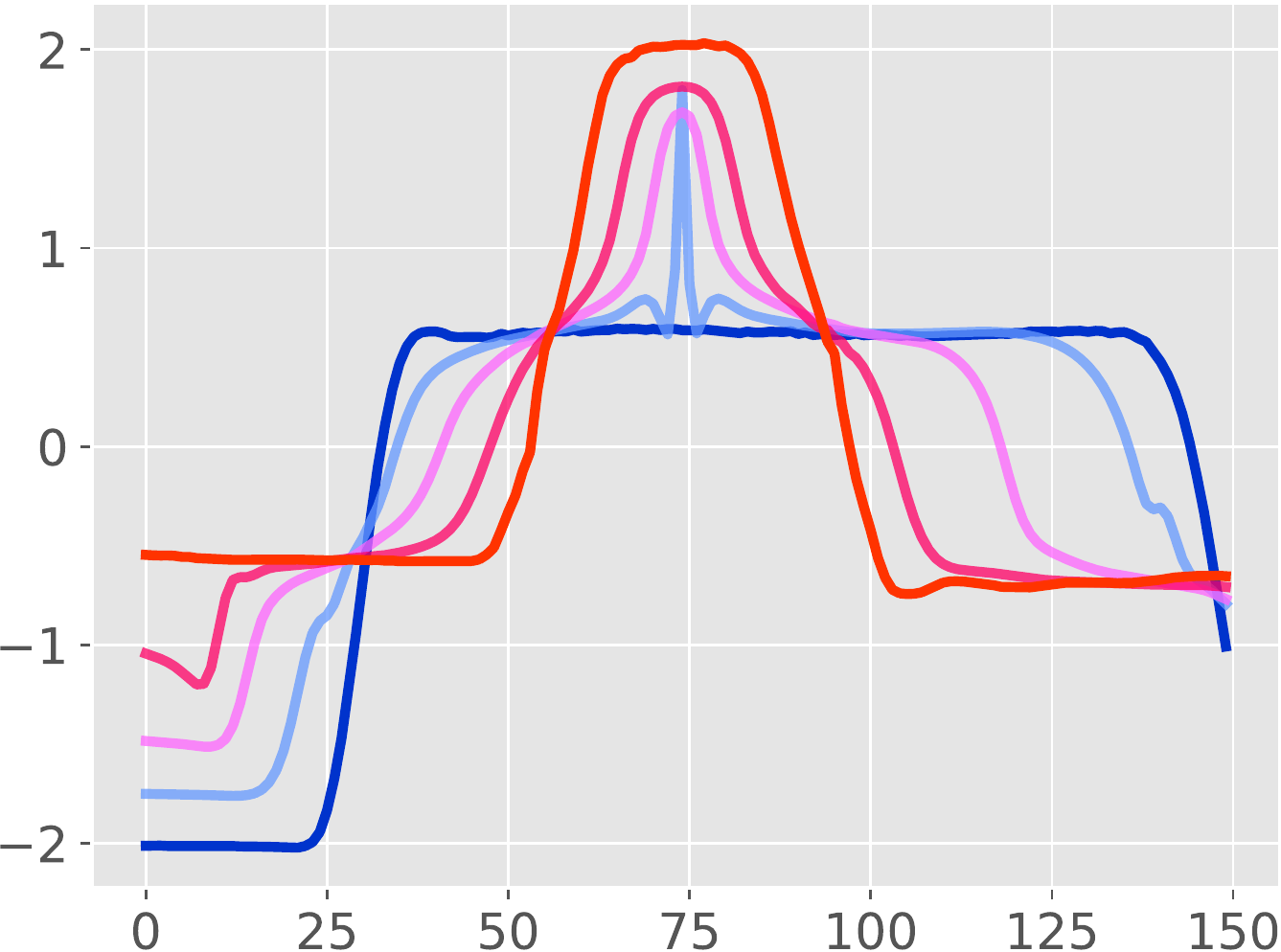}
}
   \caption{Interpolation between two time series (red and blue) on the Gun
       Point dataset. We computed the barycenter by solving
       Eq. \eqref{eq:barycenter_obj} with $(\lambda_1,\lambda_2)$ set to
       (0.25, 0.75), (0.5, 0.5) and (0.75, 0.25).  
       The soft-DTW geometry 
   leads to visibly different interpolations.
   }
   \label{fig:interpolation}
\end{figure}

We study in this section a direct application of Algorithm~\ref{algo:gendtw2}
to the problem of computing \citeauthor{frechet1948elements} means
\citeyearpar{frechet1948elements} of time series with respect to the $\sdtw$
discrepancy.
Given a family of $N$ times series $\by_1,\dots,\by_N$, namely $N$ matrices of
$p$ lines and varying number of columns, $m_1,\dots,m_N$, we are interested in
defining a single barycenter time series $\bx$ for that family under a set
of normalized weights $\lambda_1,\dots,\lambda_N\in\RR_+$ such that
$\sum_{i=1}^N\lambda_i=1$. Our goal is thus to solve approximately the following
problem, in which we have assumed that $\bx$ has fixed length $n$:
\begin{equation}
\min_{\bx\in\RR^{p\times n}} \sum_{i=1}^N \frac{\lambda_i}{m_i}\sdtw(\bx,\by_i).
\label{eq:barycenter_obj}
\end{equation}
Note that each $\sdtw(\bx,\by_i)$ term is divided by $m_i$, the length of
$\by_i$. Indeed, since $\dtw$ is an increasing (roughly linearly) function of
each of the input lengths $n$ and $m_i$, we follow the convention of normalizing
in practice each discrepancy by $n\times m_i$. Since the length $n$ of $\bx$ is
here fixed across all evaluations, we do not need to divide the objective of
Eq.~\eqref{eq:barycenter_obj} by $n$. Averaging under the soft-DTW geometry
results in substantially different results than those that can be obtained with
the Euclidean geometry (which can only be used in the case where all lengths
$n=m_1=\dots=m_N$ are equal), as can be seen in the intuitive interpolations we
obtain between two time series shown in Figure \ref{fig:interpolation}.

\textbf{Non-convexity of $\sdtw$.} A natural question that
arises from Eq.~\eqref{eq:barycenter_obj} is whether that objective is convex or
not. The answer is negative, in a way that echoes the non-convexity of the
$k$-means objective as a function of cluster centroids locations. Indeed, for
any alignment matrix $A$ of suitable size, each map $\bx\mapsto
\dotprod{A}{\Delta(\bx,\by)}$ shares the same convexity/concavity property that
$\delta$ may have. However, both $\min$ and $\mming$ can only preserve the
\emph{concavity} of elementary functions~\citep[pp.72-74]{Boyd:1072}. Therefore
$\sdtw$ will only be concave if $\delta$ is concave, or become instead a
(non-convex) (soft) minimum of convex functions if $\delta$ is convex. When
$\delta$ is a squared-Euclidean distance, $\dtw$ is a piecewise quadratic
function of $\bx$, as is also the case with the $k$-means energy (see for
instance Figure 2 in~\citealt{schultz2017nonsmooth}). Since this is the
setting we consider here, all of the computations involving barycenters should
be taken with a grain of salt, since we have no way of ensuring optimality when
approximating Eq.~\eqref{eq:barycenter_obj}. 

\textbf{Smoothing helps optimizing $\sdtw$.} Smoothing can be regarded, however,
as a way to ``convexify'' $\sdtw$. Indeed, notice that $\sdtw$ converges to the
sum of all costs as $\gamma\rightarrow \infty$. Therefore, if
$\delta$ is convex, $\sdtw$ will gradually become convex as $\gamma$ grows. For
smaller values of $\gamma$, one can intuitively foresee that using $\mming$
instead of a minimum will smooth out local minima and therefore provide a better
(although slightly different from $\dtw$) optimization landscape. We believe
this is why our approach recovers better results, \textbf{even when measured in
the original $\dtw$ discrepancy}, than subgradient or alternating minimization
approaches such as DBA~\citep{petitjean2011global}, which can, on the contrary, get more easily stuck in local minima. Evidence
for this statement is presented in the experimental section.

\subsection{Clustering with the soft-DTW geometry}

The (approximate) computation of $\sdtw$ barycenters can be seen as a first step
towards the task of clustering time series under the $\sdtw$ discrepancy.
Indeed, one can naturally formulate that problem as that of finding centroids
$\bx_1,\dots,\bx_k$ that minimize the following energy:
\begin{equation}
    \min_{\bx_1,\dots,\bx_k\in\RR^{p\times n}} \sum_{i=1}^N \frac{1}{m_i} \min_{j \in
    [[k]]} \sdtw(\bx_j,\by_i).
\label{eq:kmeans_obj}
\end{equation}
To solve that problem one can resort to a direct generalization of
\citeauthor{lloyd1982least}'s algorithm \citeyearpar{lloyd1982least} in which
each centering step and each clustering allocation step is done according to the
$\sdtw$ discrepancy.

\subsection{Learning prototypes for time series classification}

One of the de-facto baselines for learning to classify time series is the $k$
nearest neighbors ($k$-NN) algorithm, combined with DTW as discrepancy measure
between time series. However, $k$-NN has two main drawbacks. First, the
time series used for training must be stored, leading to potentially high
storage cost.  Second, in order to compute predictions on new time series, the
DTW discrepancy must be computed with all training time series, leading to high
computational cost. Both of these drawbacks can be addressed by the nearest centroid classifier
\citep[p.670]{ESLII}, \citep{tibshirani2002diagnosis}. This method chooses the
class whose barycenter (centroid) is closest to the time series to classify.
Although very simple, this method was shown to be competitive with $k$-NN, while
requiring much lower computational cost at prediction time
\citep{petitjean_icdm}. Soft-DTW can naturally be used in a nearest centroid
classifier, in order to compute the barycenter of each class at train time, and
to compute the discrepancy between barycenters and time series, at prediction
time.

\subsection{Multistep-ahead prediction}\label{subsec:predicting}

Soft-DTW is ideally suited as a loss function for any task that requires time series outputs. 
As an example of such a task, we consider the problem of, given the
first $1,\dots,t$ observations of a time series, predicting the remaining
$(t+1),\dots,n$ observations. Let $\bx^{t,t'} \in \mathbb{R}^{p \times (t'-t+1)}$
be the submatrix of $\bx \in \mathbb{R}^{p \times n}$ of all columns with
indices between $t$ and $t'$, where $1 \le t < t' < n$. Learning to predict the
segment of a time series can be cast as the
problem
\begin{equation}
    \min_{\theta\in\Theta} \sum_{i=1}^N \sdtw\left(f_\theta(\bx_i^{1,t}), \bx_i^{t+1, n}\right),
\end{equation}\vskip-.3cm
where $\{f_\theta\}$ is a set of parameterized function that take as input a time series and outputs a time series. Natural choices would be multi-layer perceptrons or
recurrent neural networks (RNN), which have been historically trained with a Euclidean loss~\citep[Eq.5]{parlos2000multi}.

 %We do not impose any restriction on these functions, other than they should be differentiable end-to-end (between input and output) to take advantage of the differentiability of $\s$

\section{Experimental results}
\label{sec:experiments}

% \subsection{Datasets} pas besoin.

Throughout this section, we use the UCR (University of California, Riverside)
time series classification archive \citep{UCRArchive}. We use a subset
containing 79 datasets encompassing a wide variety of fields (astronomy,
geology, medical imaging) and lengths. Datasets include class information (up to
60 classes) for each time series and are split into train and test sets.  Due to
the large number of datasets in the UCR archive, we choose to report only a
summary of our results in the main manuscript. Detailed results are included in
the appendices for interested readers.

\subsection{Averaging experiments}\label{subsec:barycs}

In this section, we compare the soft-DTW barycenter approach presented in
\S\ref{subsec:averaging} to
DBA \citep{petitjean2011global} and a simple batch subgradient
method.

\textbf{Experimental setup.}
For each dataset, we choose a class at random, pick 10 time series in that class
and compute their barycenter. For quantitative results below, we repeat this
procedure 10 times and report the averaged results. For each method, we set the
maximum number of iterations to 100. To minimize the proposed soft-DTW
barycenter objective, Eq. \eqref{eq:barycenter_obj}, we use L-BFGS.  

\begin{table}[t]
\centering
\mytablefontsize
\caption{Percentage of the datasets on which the proposed soft-DTW barycenter
    is achieving lower DTW loss (Equation \eqref{eq:barycenter_obj} with
$\gamma=0$) than competing methods.}
\begin{tabular}{l C{2cm} C{3cm}}
\toprule
& Random initialization & Euclidean mean initialization \\
\cmidrule[1pt](rl){2-3}
\multicolumn{3}{l}{{\bfseries Comparison with DBA}} \\
\addlinespace[0.3em]
$\gamma=1$ & 40.51\% & 3.80\% \\
$\gamma=0.1$ & 93.67\% & 46.83\% \\
$\gamma=0.01$ & 100\% & 79.75\% \\
$\gamma=0.001$ & 97.47\% & 89.87\% \\
\midrule
\multicolumn{3}{l}{{\bfseries Comparison with subgradient method}} \\
$\gamma=1$ & 96.20\% & 35.44\% \\
$\gamma=0.1$ & 97.47\% & 72.15\% \\
$\gamma=0.01$ & 97.47\% & 92.41\% \\
$\gamma=0.001$ & 97.47\% & 97.47\% \\
\bottomrule
\end{tabular}
\label{table:barycenter_res}
\end{table}

\begin{figure}[t]
\centering
\includegraphics[scale=0.32]{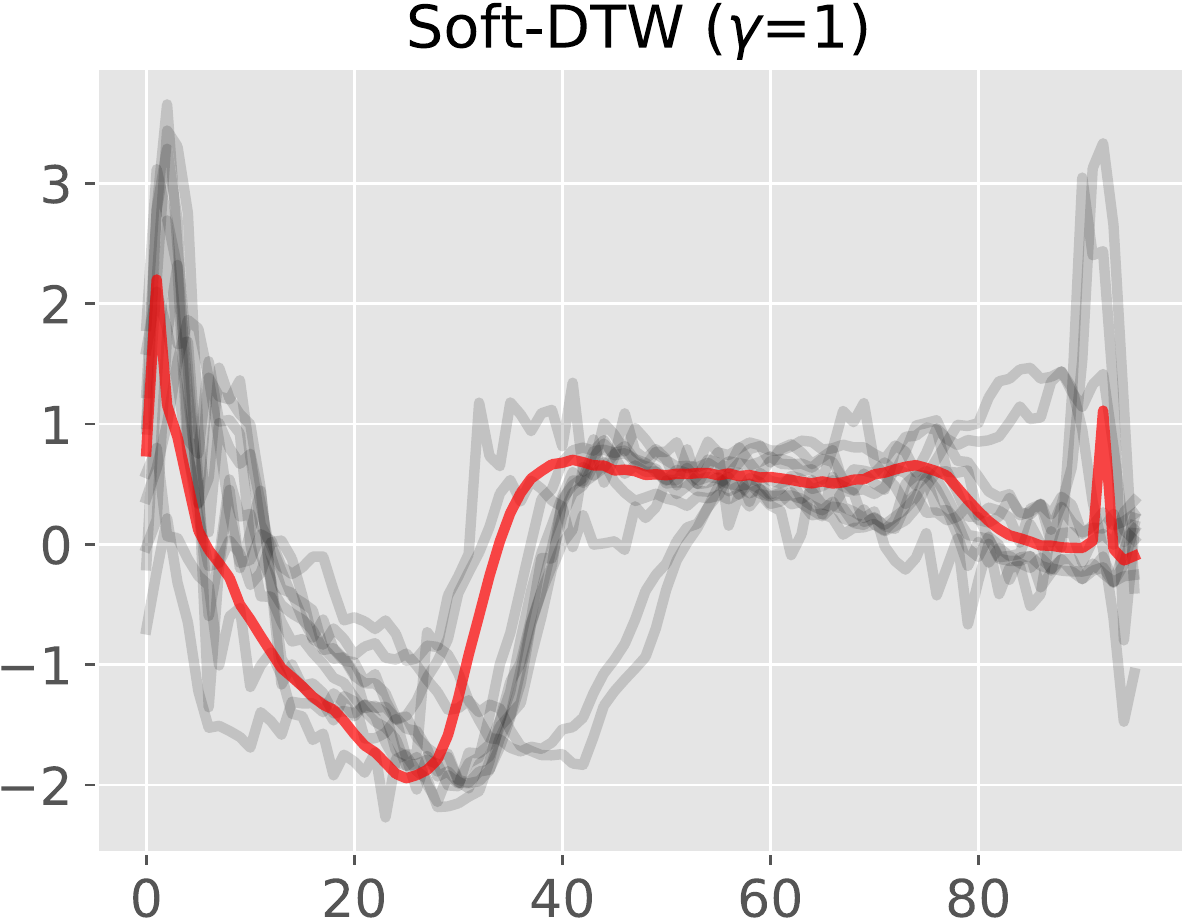}
\includegraphics[scale=0.32]{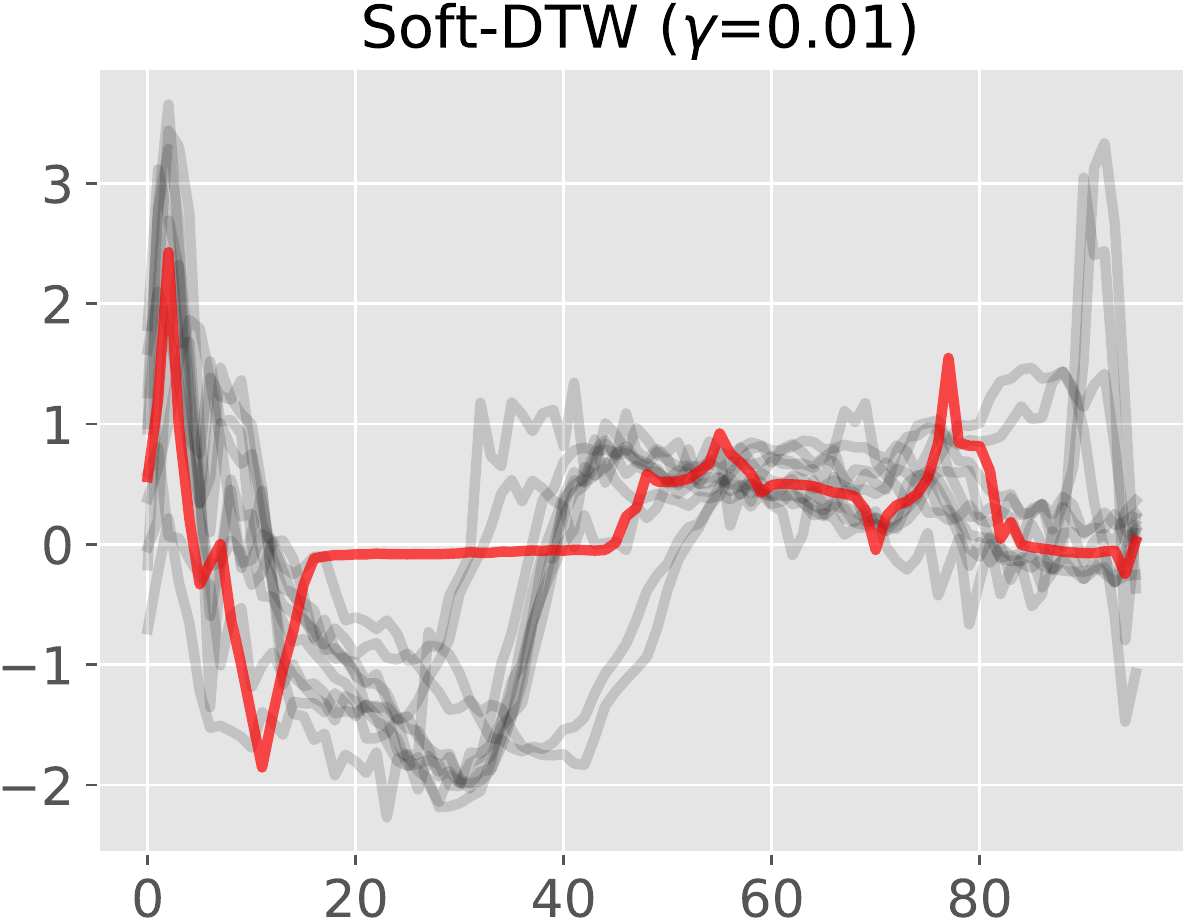}
\includegraphics[scale=0.32]{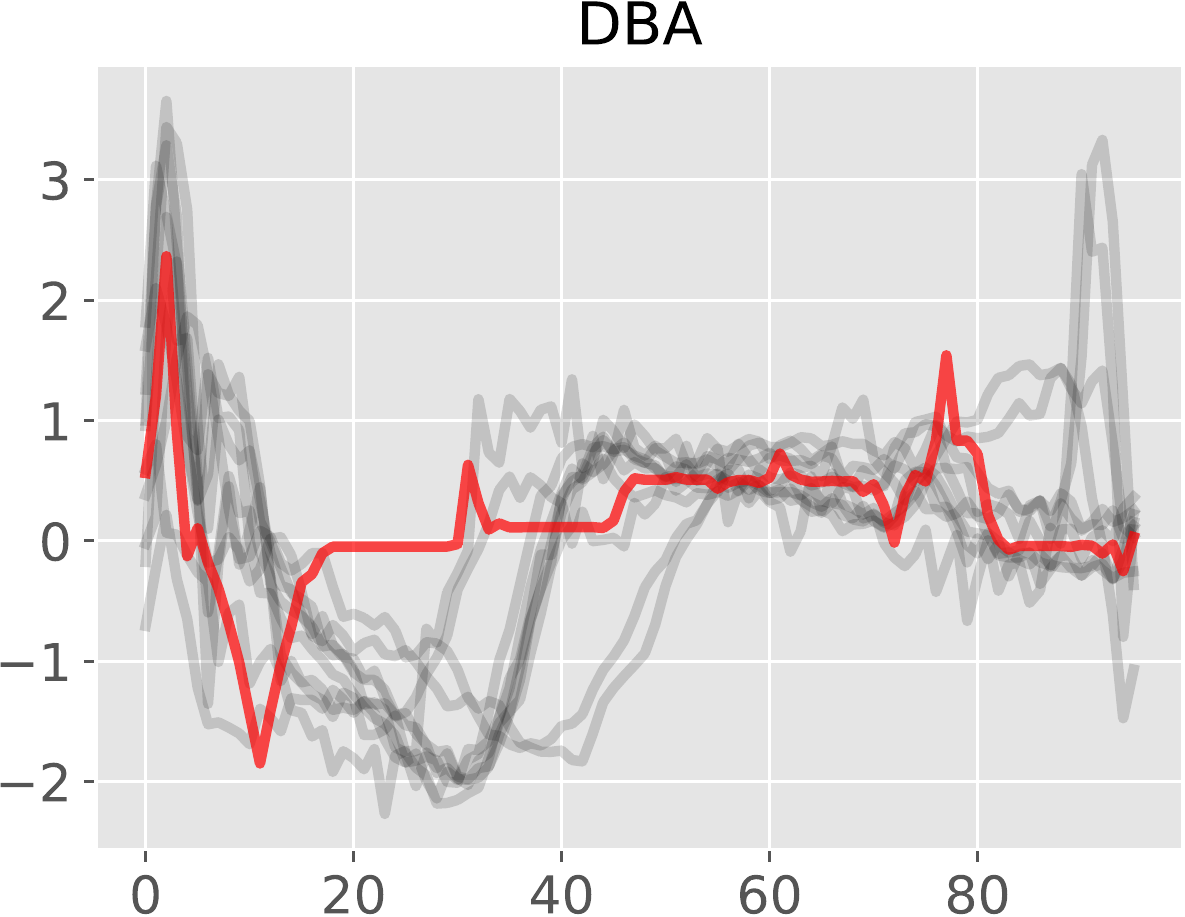}
\includegraphics[scale=0.32]{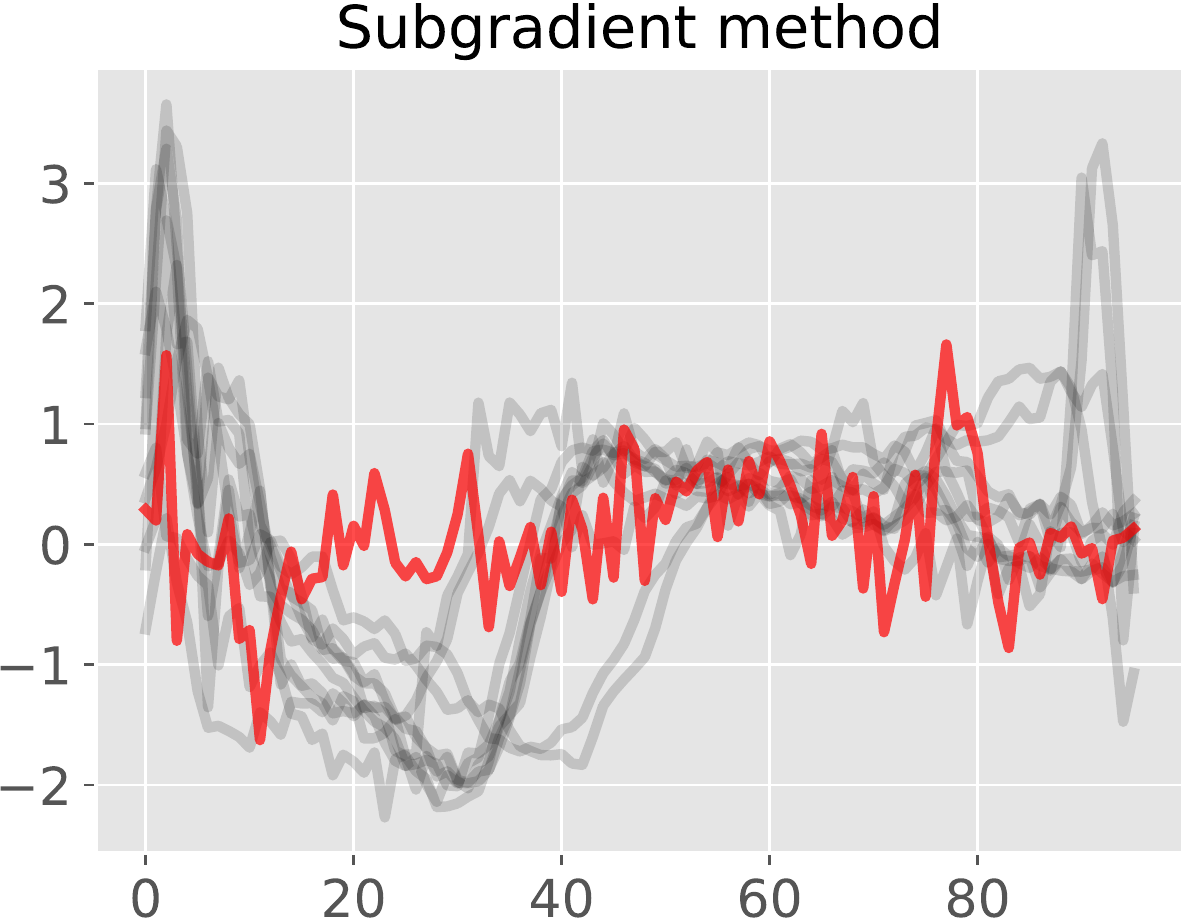}
\caption{Comparison between our proposed soft barycenter and the barycenter
obtained by DBA and the subgradient method, on the ECG200 dataset. When DTW is
insufficiently smoothed, barycenters often get stuck in a bad local minimum that
does not correctly match the time series.} 
\label{fig:barycenter_vis}
\end{figure}

\textbf{Qualitative results.}  We first visualize the barycenters obtained by
soft-DTW when $\gamma=1$ and $\gamma=0.01$, by DBA and by the subgradient
method.  Figure \ref{fig:barycenter_vis} shows barycenters obtained using random
initialization on the ECG200 dataset. More results with both random and Euclidean mean initialization are given in Appendix
\ref{appendix:barycenter_vis_random} and \ref{appendix:barycenter_vis_euc}.

We observe that both DBA or soft-DTW with low smoothing parameter $\gamma$ yield barycenters that are spurious. On the other hand, a descent on the soft-DTW loss with sufficiently high $\gamma$
converges to a reasonable solution. For example, as indicated in Figure
\ref{fig:barycenter_vis}
with DTW or soft-DTW ($\gamma=0.01$), the small kink
around $x=15$ is not representative of any of the time series in the dataset. However,
with soft-DTW ($\gamma=1$), the barycenter closely matches the time series. This
suggests that DTW or soft-DTW with too low $\gamma$ can get stuck in bad local
minima.

When using Euclidean mean initialization (only possible if time series have the
same length), DTW or soft-DTW with low $\gamma$ often yield barycenters that
better match the shape of the time series. However, they tend to overfit: they
absorb the idiosyncrasies of the data. In contrast, soft-DTW is able to learn
barycenters that are much smoother.

\textbf{Quantitative results.} Table \ref{table:barycenter_res} summarizes
the percentage of datasets on which the proposed soft-DTW barycenter achieves
lower DTW loss when varying the smoothing parameter $\gamma$. The actual loss
values achieved by different methods are indicated in Appendix
\ref{appendix:barycenter_random} and Appendix \ref{appendix:barycenter_euc}.

As $\gamma$ decreases, soft-DTW achieves a lower DTW loss than
other methods on almost all datasets. This confirms our claim
that the smoothness of soft-DTW leads to an objective that is better behaved and
more amenable to optimization by gradient-descent methods. %Indeed, soft-DTW yields solutions that compare favorably to other methods, even when measured in terms of the original DTW loss, despite the fact that it does not play a rolw  considered to estimate smoothed barycenters.  %On the practical slide, smoothness allows the use of line search; therefore we do not need to care about tuning a step size hyper-parameter, which is an important drawback of subgradient methods.

\subsection{$k$-means clustering experiments}
We consider in this section the same computational tools used in \S\ref{subsec:barycs} above, but use them to cluster time series.

\textbf{Experimental setup.} For all datasets, the number of clusters $k$ is equal to the number of classes available in the dataset. \citeauthor{lloyd1982least}'s algorithm alternates between a centering step (barycenter computation) and an assignment step. We set the maximum number of outer iterations to $30$ and the maximum number of inner
(barycenter) iterations to 100, as before. Again, for soft-DTW, we use L-BFGS.

\begin{figure}[t]
\centering
\subfigure[Soft-DTW ($\gamma=1$)]{
\includegraphics[scale=0.32]{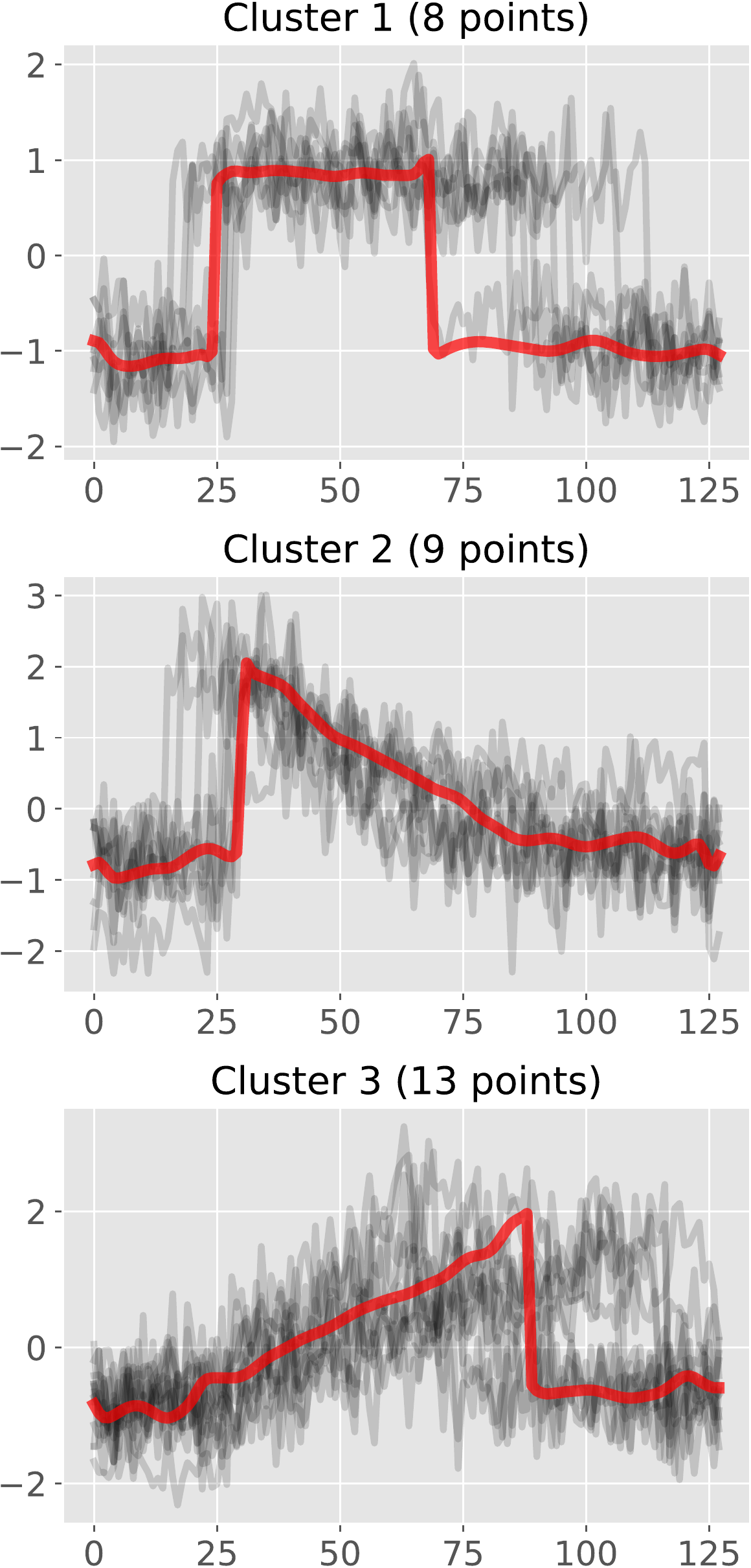}
}
\subfigure[DBA]{
\includegraphics[scale=0.32]{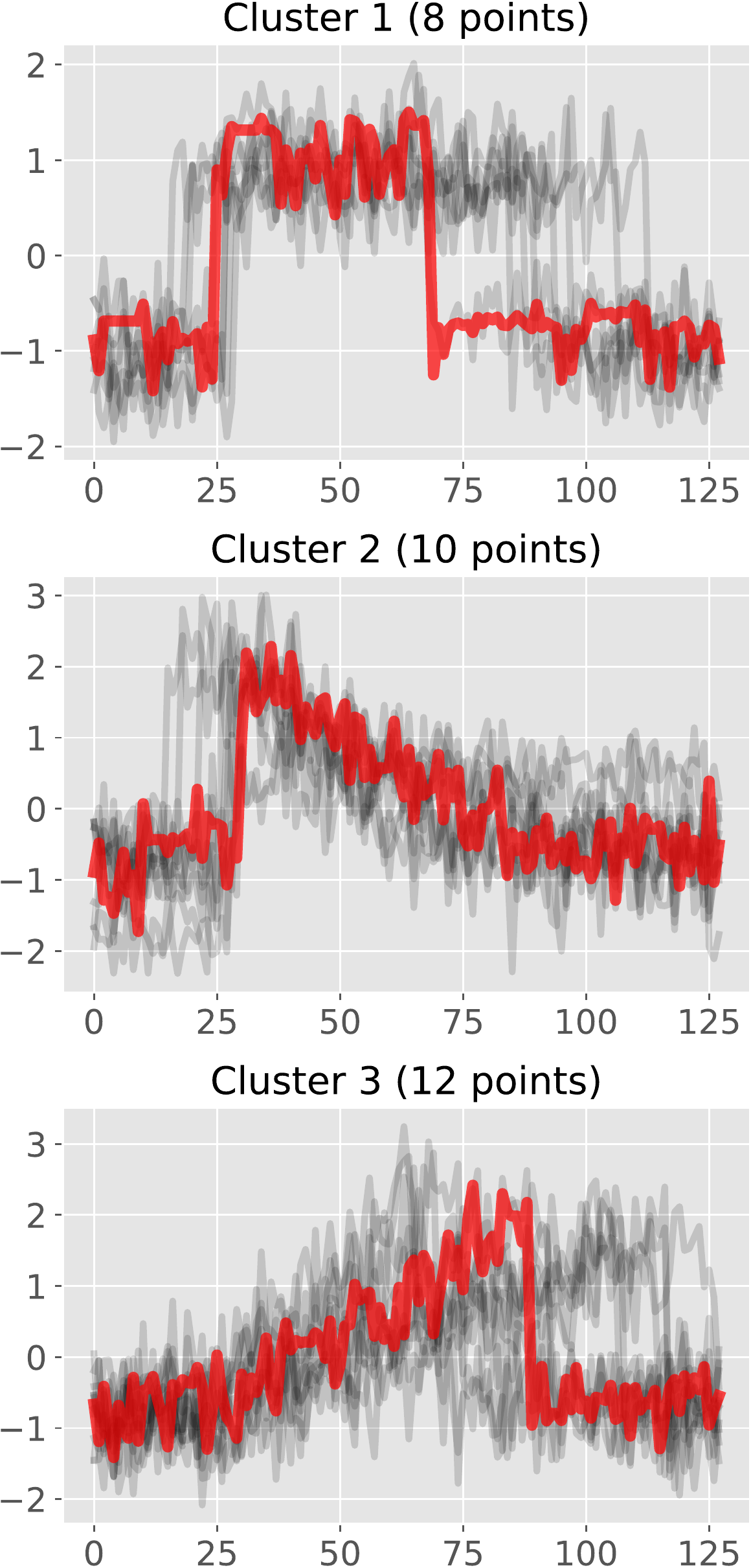}
}
\caption{Clusters obtained on the CBF dataset when plugging our proposed
    soft barycenter and that of DBA in Lloyd's algorithm. DBA
    absorbs the
idiosyncrasies of the data, while soft-DTW can learn much smoother barycenters.}
\label{fig:clustering_vis}
\end{figure}

\textbf{Qualitative results.} Figure \ref{fig:clustering_vis} shows the
clusters obtained when runing Lloyd's algorithm on the CBF dataset with soft-DTW
($\gamma=1$) and DBA, in the case of random
initialization. More results are included in
Appendix \ref{appendix:clustering_vis}. Clearly, DTW absorbs the tiny details in
the data, while soft-DTW is able to learn much smoother barycenters.

\textbf{Quantitative results.} Table \ref{table:kmeans_res} summarizes the
percentage of datasets on which soft-DTW barycenter achieves lower $k$-means
loss under DTW, i.e. Eq. \eqref{eq:kmeans_obj} with $\gamma=0$.
The actual loss values achieved by all
methods are indicated in Appendix \ref{appendix:k_means_random} and Appendix
\ref{appendix:k_means_euc}. The results confirm the same trend as for the barycenter experiments. Namely, as $\gamma$ decreases, soft-DTW is able to achieve lower loss than
other methods on a large proportion of the datasets. Note that we have not run
experiments with smaller values of $\gamma$ than 0.001, since
$\mathbf{dtw}_{0.001}$ is very close to $\dtw$ in practice.

\begin{table}[t]
\centering
\mytablefontsize
\caption{Percentage of the datasets on which the proposed soft-DTW based
    {\bf $k$-means} is achieving lower DTW loss (Equation \eqref{eq:kmeans_obj}
    with $\gamma=0$) than competing methods.}
\begin{tabular}{l C{2cm} C{3cm}}
\toprule
& Random initialization & Euclidean mean initialization \\
\cmidrule[1pt](rl){2-3}
\multicolumn{3}{l}{{\bfseries Comparison with DBA}} \\
\addlinespace[0.3em]
$\gamma=1$ & 15.78\% & 29.31\% \\
$\gamma=0.1$ & 24.56\% & 24.13\% \\
$\gamma=0.01$ & 59.64\% & 55.17\% \\
$\gamma=0.001$ & 77.19\% & 68.97\% \\
\midrule
\multicolumn{3}{l}{{\bfseries Comparison with subgradient method}} \\
$\gamma=1$ & 42.10\% & 46.44\% \\
$\gamma=0.1$ & 57.89\% & 50\% \\
$\gamma=0.01$ & 76.43\% & 65.52\% \\
$\gamma=0.001$ & 96.49\% & 84.48\% \\
\bottomrule
\end{tabular}
\label{table:kmeans_res}
\end{table}

\subsection{Time-series classification experiments}

In this section, we investigate whether the smoothing in soft-DTW can act as a
useful regularization and improve classification accuracy in the
nearest centroid classifier. 

\textbf{Experimental setup.} We use 50\% of the data for training, 25\% for
validation and 25\% for testing.  We choose $\gamma$ from 15 log-spaced values
between $10^{-3}$ and $10$.

\begin{figure}[t]
\centering
\subfigure{\includegraphics[width=0.30\textwidth]{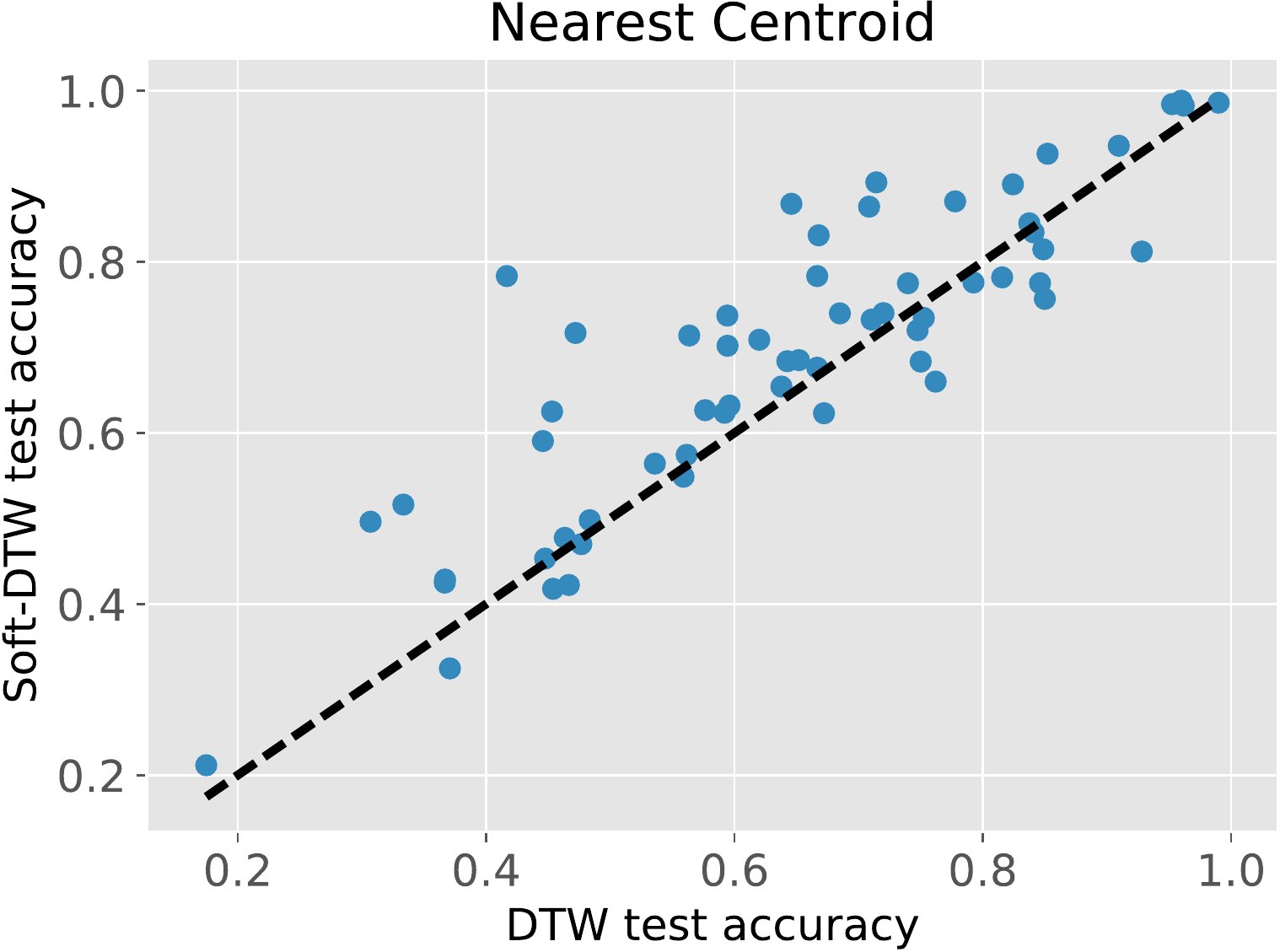}}
\caption{Each point above the diagonal represents a dataset
where using our soft-DTW barycenter rather than that of
DBA improves the accuracy of the nearest nearest centroid classifier.
This is the case for
{\bf 75\%} of the datasets in the UCR archive.}
\label{fig:classif_res}
\vskip-.3cm\end{figure}

\textbf{Quantitative results.} Each point in Figure \ref{fig:classif_res}
above the diagonal line represents a dataset for which using soft-DTW for
barycenter computation rather than DBA improves the accuracy of the nearest
centroid classifier.  To summarize, we found that soft-DTW is working better or at least as well as DBA in 75\% of the datasets.

\subsection{Multistep-ahead prediction experiments}

In this section, we present preliminary experiments for the task of
multistep-ahead prediction, described in \S\ref{subsec:predicting}.

\textbf{Experimental setup.} We use the training and test sets pre-defined in
the UCR archive. In both the training and test sets, we use the first 60\% of
the time series as input and the remaining 40\% as output, ignoring class information. We then use the
training set to learn a model that predicts the outputs from inputs and the
test set to evaluate results with both Euclidean and DTW losses. In this
experiment, we focus on a simple multi-layer perceptron (MLP) with one hidden
layer and sigmoid activation. We also experimented with linear models and
recurrent neural networks (RNNs) but they did not improve over a simple MLP.

\textbf{Implementation details.} Deep learning frameworks such as Theano,
TensorFlow and Chainer allow the user to specify a custom backward pass for
their function. Implementing such a backward pass, rather than resorting to
automatic differentiation (autodiff), is particularly important in the case of
soft-DTW: First, the autodiff in these frameworks is
designed for vectorized operations, whereas the dynamic program used by the
forward pass of Algorithm \ref{algo:gendtw} is inherently element-wise; Second,
as we explained in \S\ref{subsec:differentiation}, our backward pass is able to
re-use log-sum-exp computations from the forward pass, leading to both lower
computational cost and better numerical stability. We implemented a custom
backward pass in Chainer, which can then be used to plug soft-DTW as a loss function
in any network architecture. To estimate the MLP's parameters, we used
Chainer's implementation of Adam~\cite{kingma2014adam}.

\textbf{Qualitative results.} Visualizations of the predictions obtained
under Euclidean and soft-DTW losses are given in Figure
\ref{fig:ts_pred_example}, as well as in Appendix \ref{appendix:more_ts_pred}.
We find that for simple one-dimensional time series, an MLP works very well,
showing its ability to capture patterns in the training set. Although the
predictions under Euclidean and soft-DTW losses often agree with each other,
they can sometimes be visibly different. Predictions under soft-DTW loss can confidently predict abrupt and sharp changes since those have a low DTW cost as long as such a sharp change is present, under a small time shift, in the ground truth.

\textbf{Quantitative results.} A comparison summary of our MLP under
Euclidean and soft-DTW losses over the UCR archive is given in Table
\ref{table:ts_predict_res}. Detailed results are given in the appendix. 
Unsurprisingly, we achieve lower DTW loss when training with the soft-DTW loss, and
lower Euclidean loss when training with the Euclidean loss. Because DTW is
robust to several useful invariances, a small error in the soft-DTW sense could be a more judicious choice than an error in an Euclidean sense for many applications.

\begin{table}[t]
\centering
\mytablefontsize
\caption{Averaged rank obtained by a multi-layer perceptron
(MLP) under Euclidean and soft-DTW losses.  Euclidean initialization means that
we initialize the MLP trained with soft-DTW loss by the solution of the MLP
trained with Euclidean loss.}
\begin{tabular}{l C{2cm} C{2cm}}
\toprule
Training loss & Random initialization & Euclidean initialization \\
\cmidrule[1pt](rl){1-3}
\multicolumn{3}{l}{{\bfseries When evaluating with DTW loss}} \\
\addlinespace[0.3em]
Euclidean & 3.46 & 4.21  \\
soft-DTW ($\gamma=1$) & 3.55 & 3.96 \\
soft-DTW ($\gamma=0.1$) & 3.33 & 3.42 \\
soft-DTW ($\gamma=0.01$) & 2.79 & 2.12\\
soft-DTW ($\gamma=0.001$) & {\bf 1.87} & {\bf 1.29}\\
\midrule
\multicolumn{3}{l}{{\bfseries When evaluating with Euclidean loss}} \\
\addlinespace[0.3em]
Euclidean & {\bf 1.05} & {\bf 1.70}\\
soft-DTW ($\gamma=1$) & 2.41 & 2.99\\
soft-DTW ($\gamma=0.1$) & 3.42 & 3.38 \\
soft-DTW ($\gamma=0.01$) & 4.13 & 3.64 \\
soft-DTW ($\gamma=0.001$) & 3.99 & 3.29\\
\bottomrule
\end{tabular}
\label{table:ts_predict_res}\vskip-.5cm
\end{table}

%\paragraph{Conclusion.} 
\section{Conclusion}
We propose in this paper to turn the popular DTW
discrepancy between time series into a full-fledged loss function between ground
truth time series and outputs from a learning machine. We have shown
experimentally that, on the existing problem of computing barycenters and
clusters for time series data, our computational approach is superior to
existing baselines. We have shown promising results on the problem of
multistep-ahead time series prediction, which could prove extremely useful in
settings where a user's actual loss function for time series is closer to the
robust perspective given by DTW, than to the local parsing of the Euclidean
distance.

\paragraph{Acknowledgements.} MC gratefully acknowledges the support of a \emph{chaire de l'IDEX Paris Saclay}.
%\paragraph{Conclusion}


\clearpage

%\bibliography{softDTW}
%\bibliographystyle{icml2017}

\clearpage
\onecolumn
\appendix

\begin{center}
    {\Huge \bf Appendix material}
\end{center}

\section{Recursive forward computation of the average path matrix} 
\label{appendix:forward_recursion}

The average alignment under Gibbs distribution $p_\gamma$ can be computed with
the following forward recurrence, which mimics closely Bellman's original
recursion. For each $i\in\In,j\in\Im$, define
		$$
		E_{i+1,j+1}= \begin{bmatrix}
		e^{-\delta_{i+1,j+1}/\gamma} E_{i,j}&\mathbf{0}_i \\ \mathbf{0}_j^T &e^{-r_{i+1,j+1}/\gamma} \end{bmatrix}+
		\begin{bmatrix}\multicolumn{2}{c}{e^{-\delta_{i+1,j+1}/\gamma} E_{i,j+1}}\\  \mathbf{0}_j^T &e^{-r_{i+1,j+1}/\gamma} \end{bmatrix}+
		\begin{bmatrix}\multirow{2}{*}{$e^{-\delta_{i+1,j+1}/\gamma}E_{i+1,j}$} & \mathbf{0}_i\\ & e^{-r_{ij}/\gamma}\end{bmatrix}
		$$		
Here terms $r_{ij}$ are computed following the recursion in Algorithm~\ref{algo:gendtw2}. Border matrices are initialized to $0$, except for $E_{1,1}$ which is initialized to $[1]$. Upon completion, the average alignment matrix is stored in $E_{n,m}$.

The operation above consists in summing three matrices of size $(i+1,j+1)$. There are exactly $(nm)$ such updates. A careful implementation of this algorithm, that would only store two arrays of matrices, as Algorithm~\ref{algo:gendtw} only store two arrays of values, can be carried out in $nm \min(n,m)$ space but it would still require $(nm)^2$ operations.
\newpage
	
\section{Barycenters obtained with random initialization}
\label{appendix:barycenter_vis_random}

\begin{figure}[H]
\centering
\subfigure[CBF]{
\includegraphics[scale=0.35]{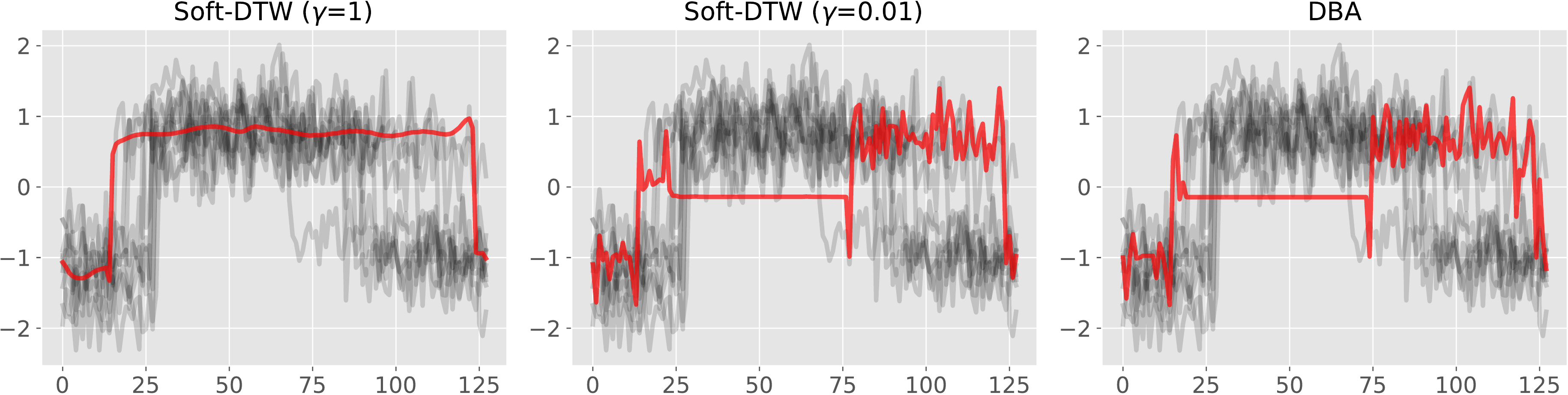}
}
\subfigure[Herring]{
\includegraphics[scale=0.35]{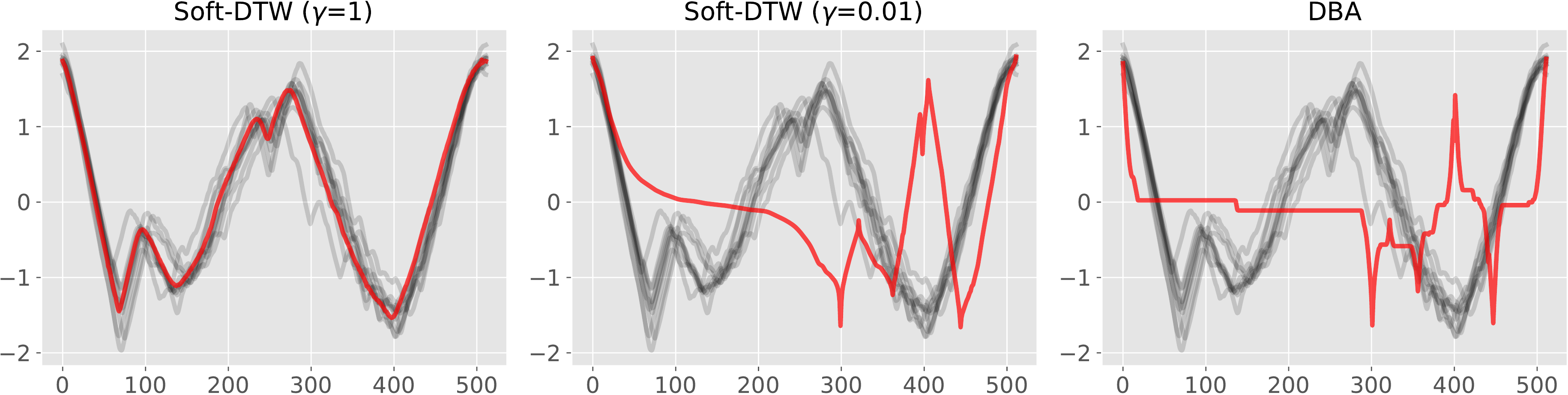}
}
\subfigure[Medical Images]{
\includegraphics[scale=0.35]{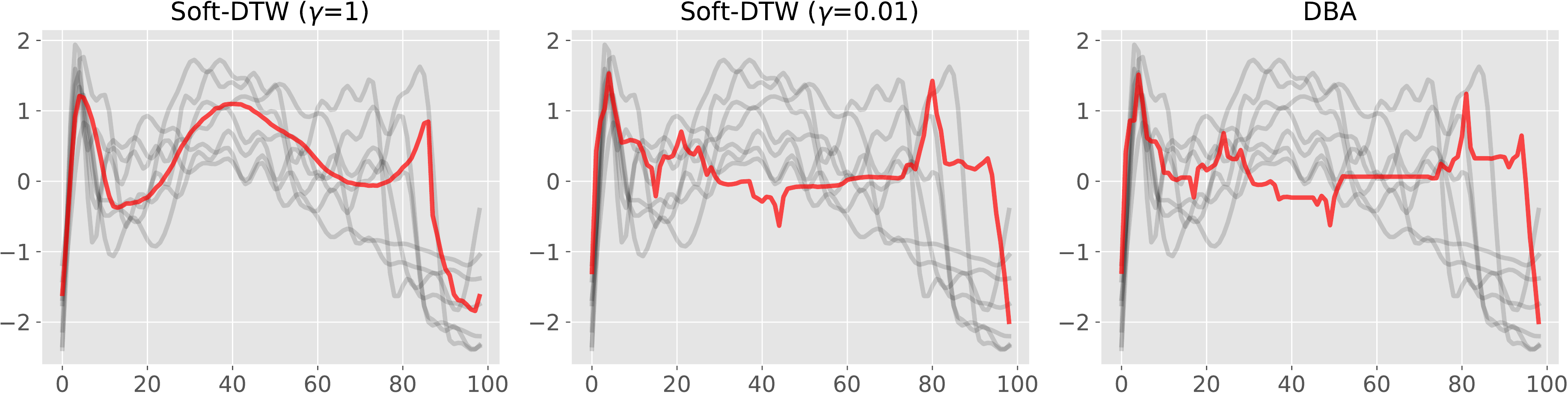}
}
\subfigure[Synthetic Control]{
\includegraphics[scale=0.35]{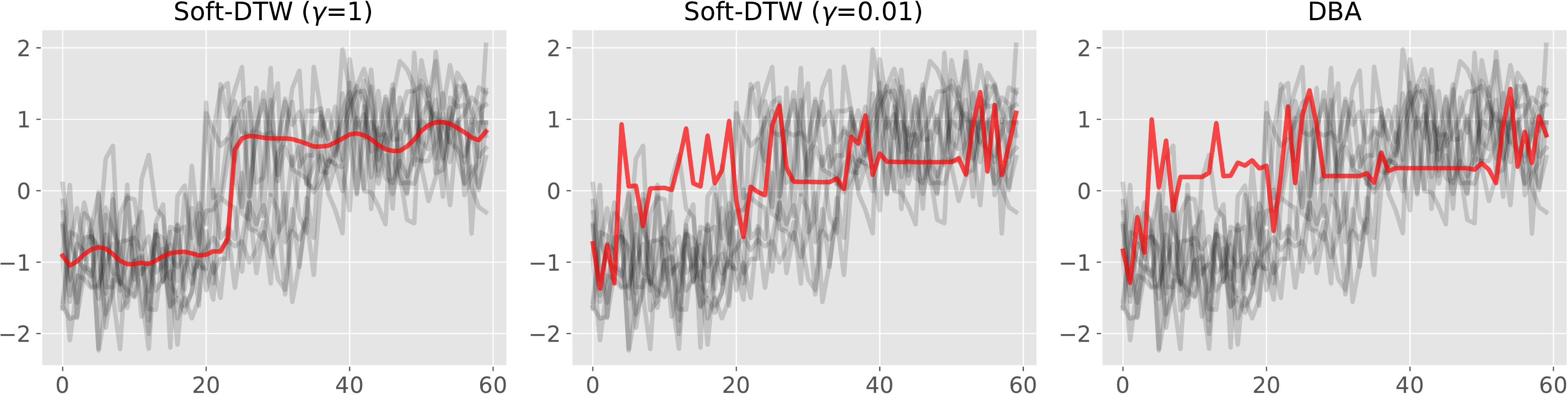}
}
\subfigure[Wave Gesture Library Y]{
\includegraphics[scale=0.35]{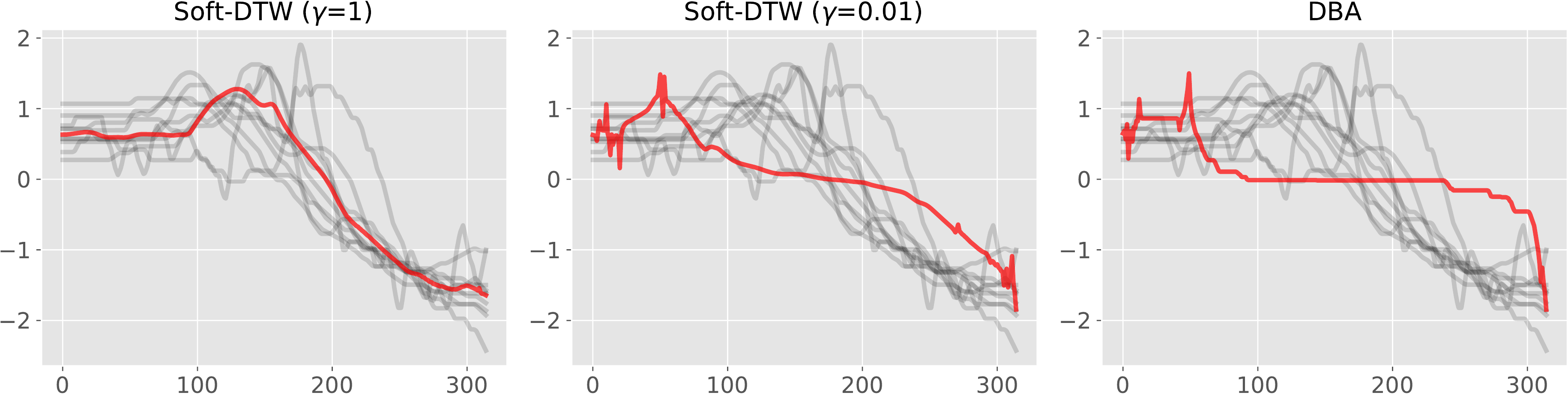}
}
%\caption{}
\end{figure}

\section{Barycenters obtained with Euclidean mean initialization}
\label{appendix:barycenter_vis_euc}

\setcounter{subfigure}{0}
\begin{figure}[H]
\centering
\subfigure[CBF]{
\includegraphics[scale=0.35]{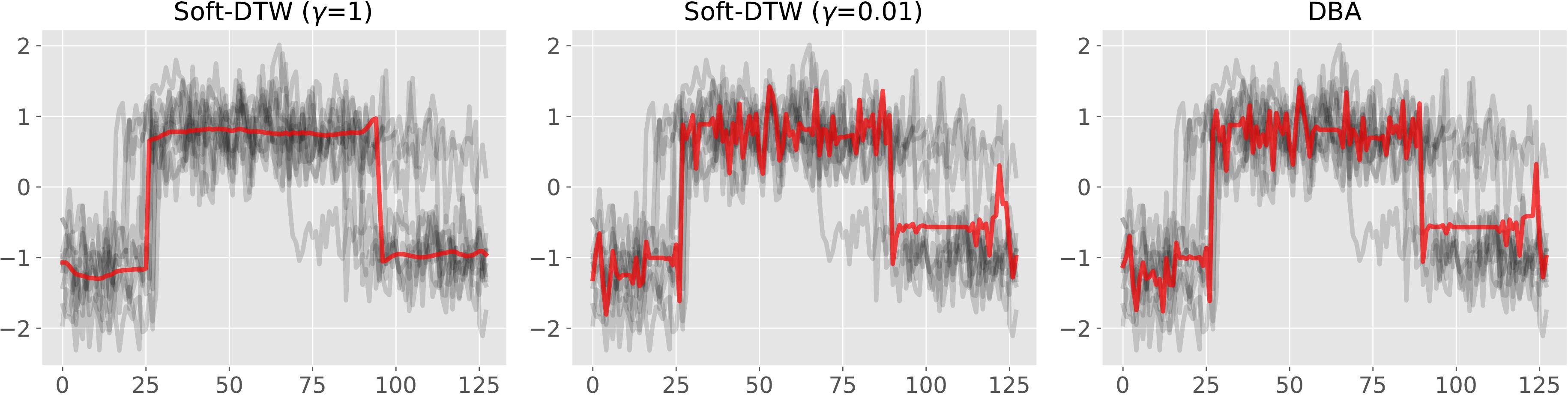}
}
\subfigure[Herring]{
\includegraphics[scale=0.35]{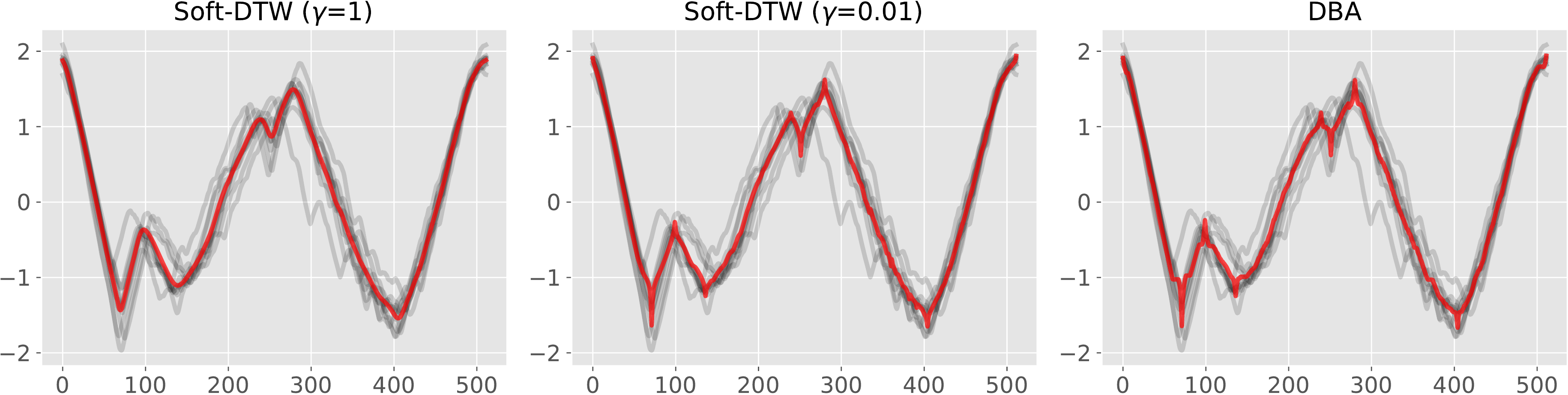}
}
\subfigure[Medical Images]{
\includegraphics[scale=0.35]{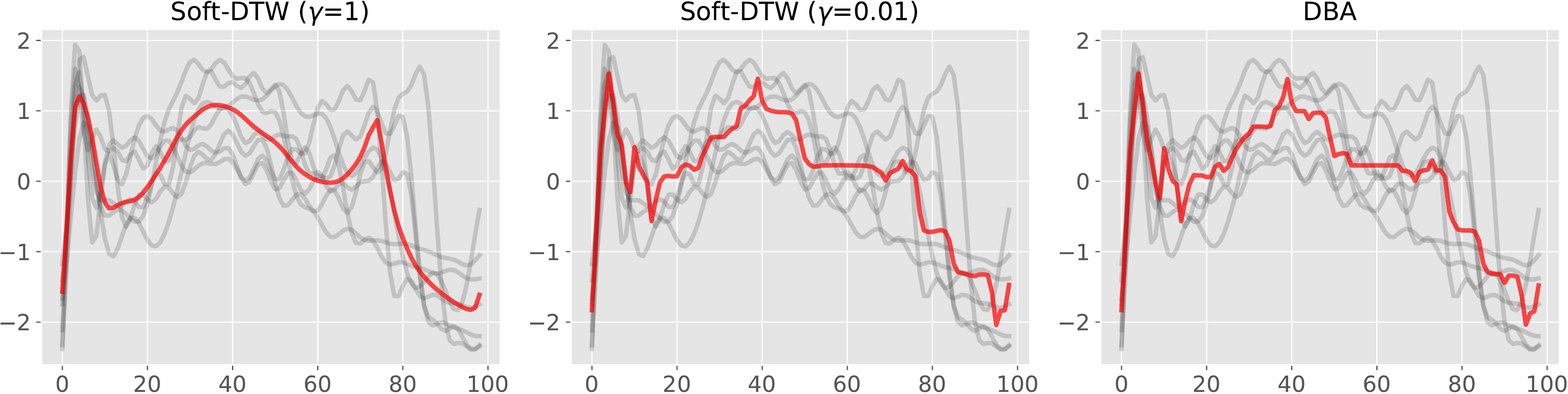}
}
\subfigure[Synthetic Control]{
\includegraphics[scale=0.35]{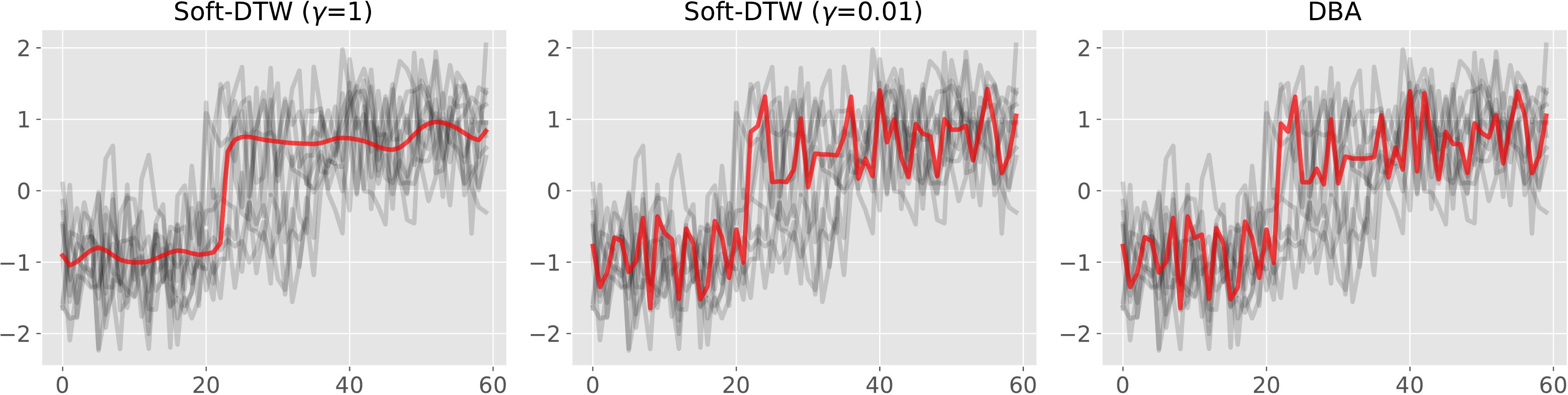}
}
\subfigure[Wave Gesture Library Y]{
\includegraphics[scale=0.35]{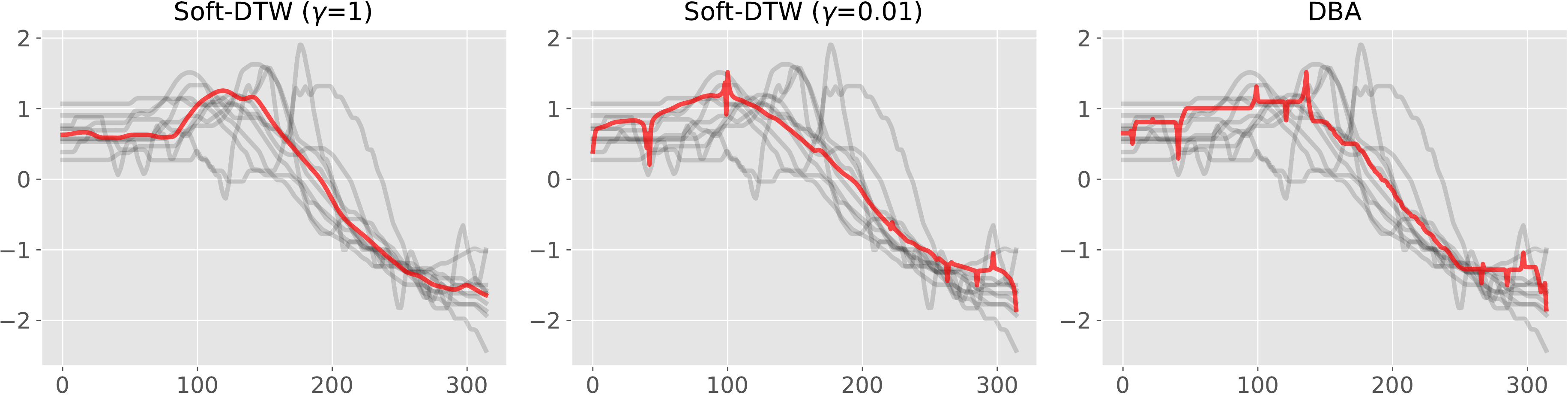}
}
%\caption{}
\end{figure}

\section{More interpolation results}

Left: results obtained under Euclidean loss. Right: results obtained under
soft-DTW ($\gamma=1$) loss.

\setcounter{subfigure}{0}
\begin{figure}[H]
\centering
%\subfigure[Adiac]{
%\includegraphics[scale=0.45]{tmp/Adiac_82_13_euc.pdf}
%\includegraphics[scale=0.45]{tmp/Adiac_82_13_sdtw.pdf}
%}
\subfigure[ArrowHead]{
\includegraphics[scale=0.43]{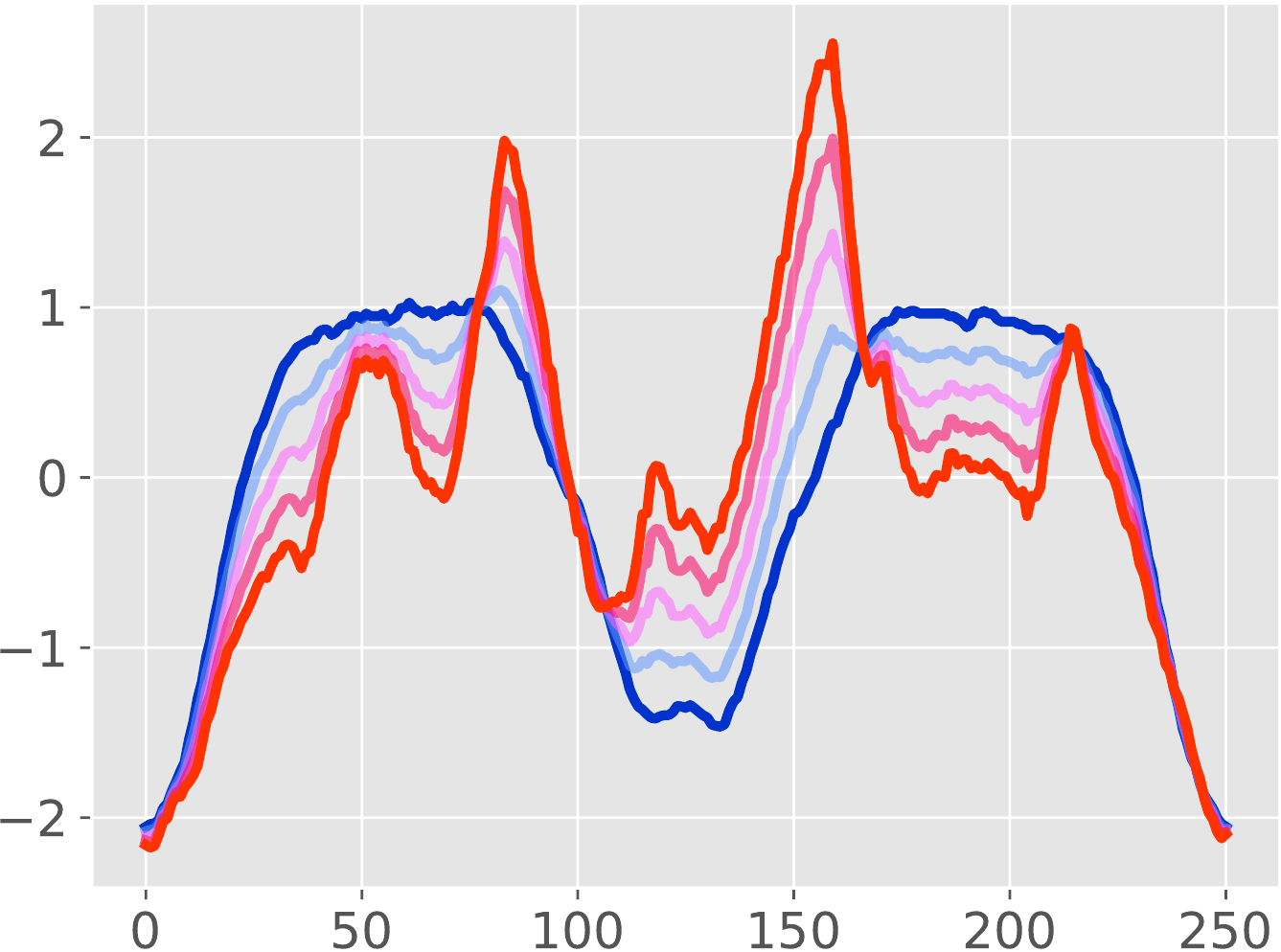}
\includegraphics[scale=0.43]{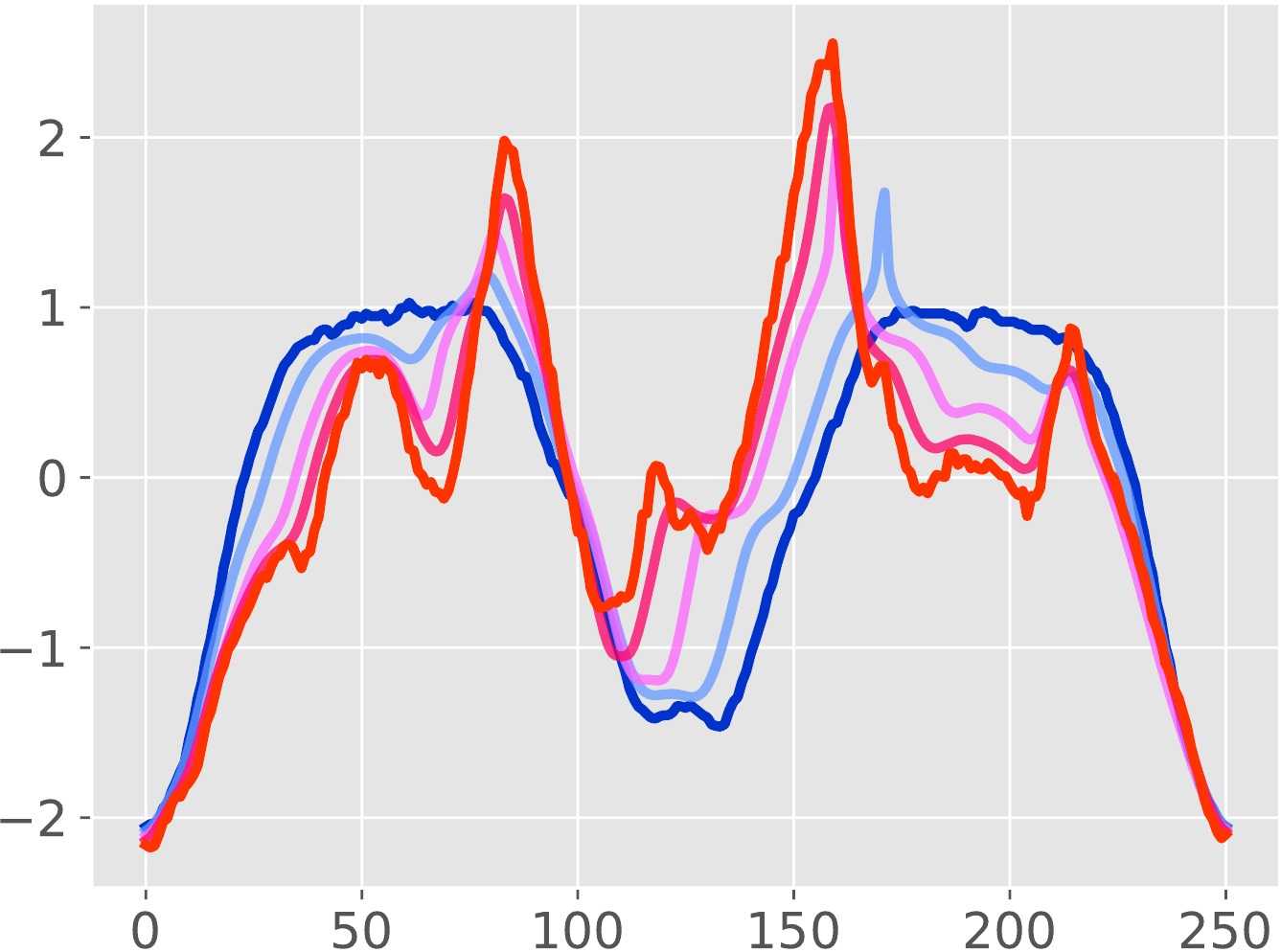}
}
\subfigure[ECG200]{
\includegraphics[scale=0.43]{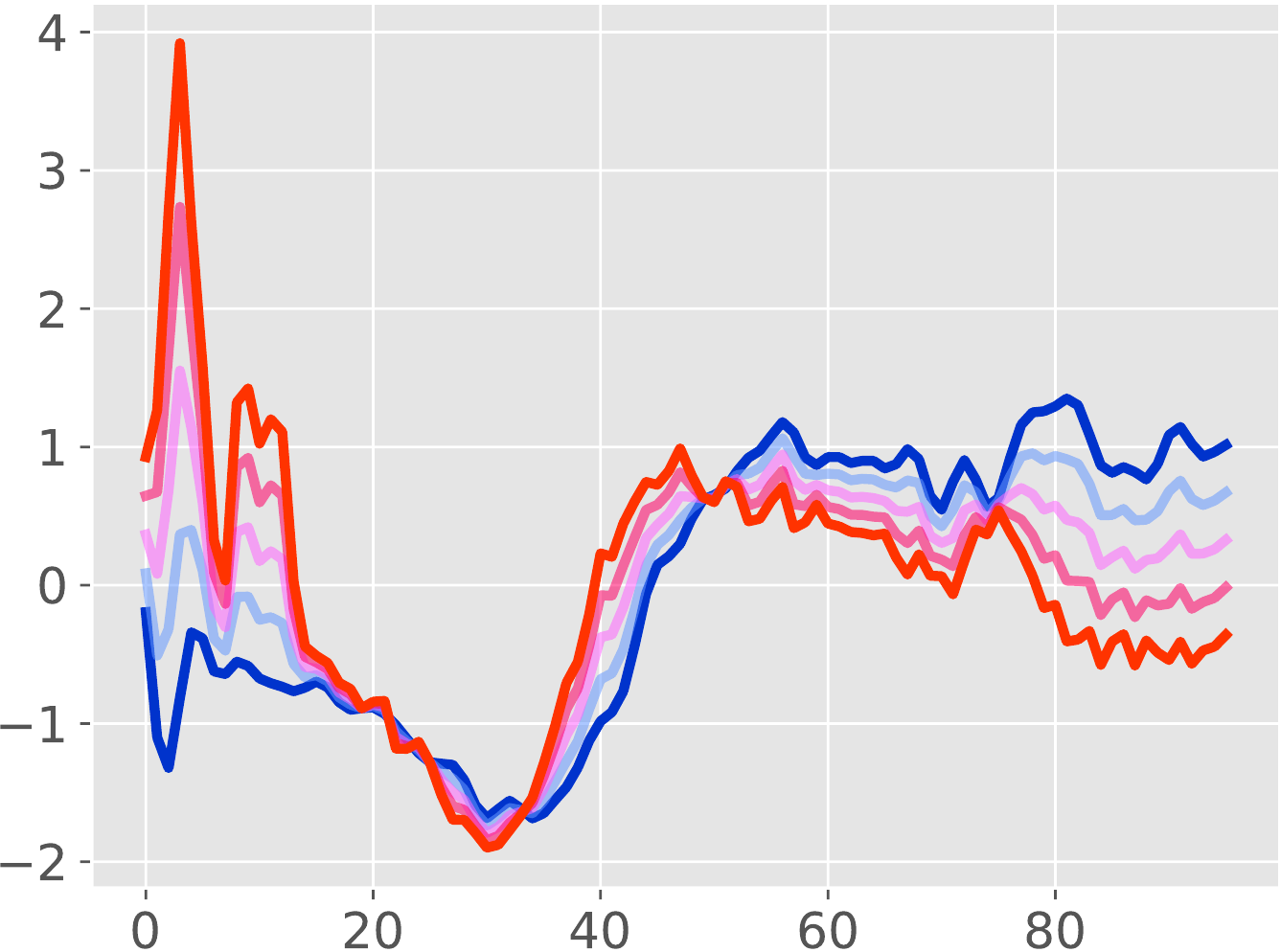}
\includegraphics[scale=0.43]{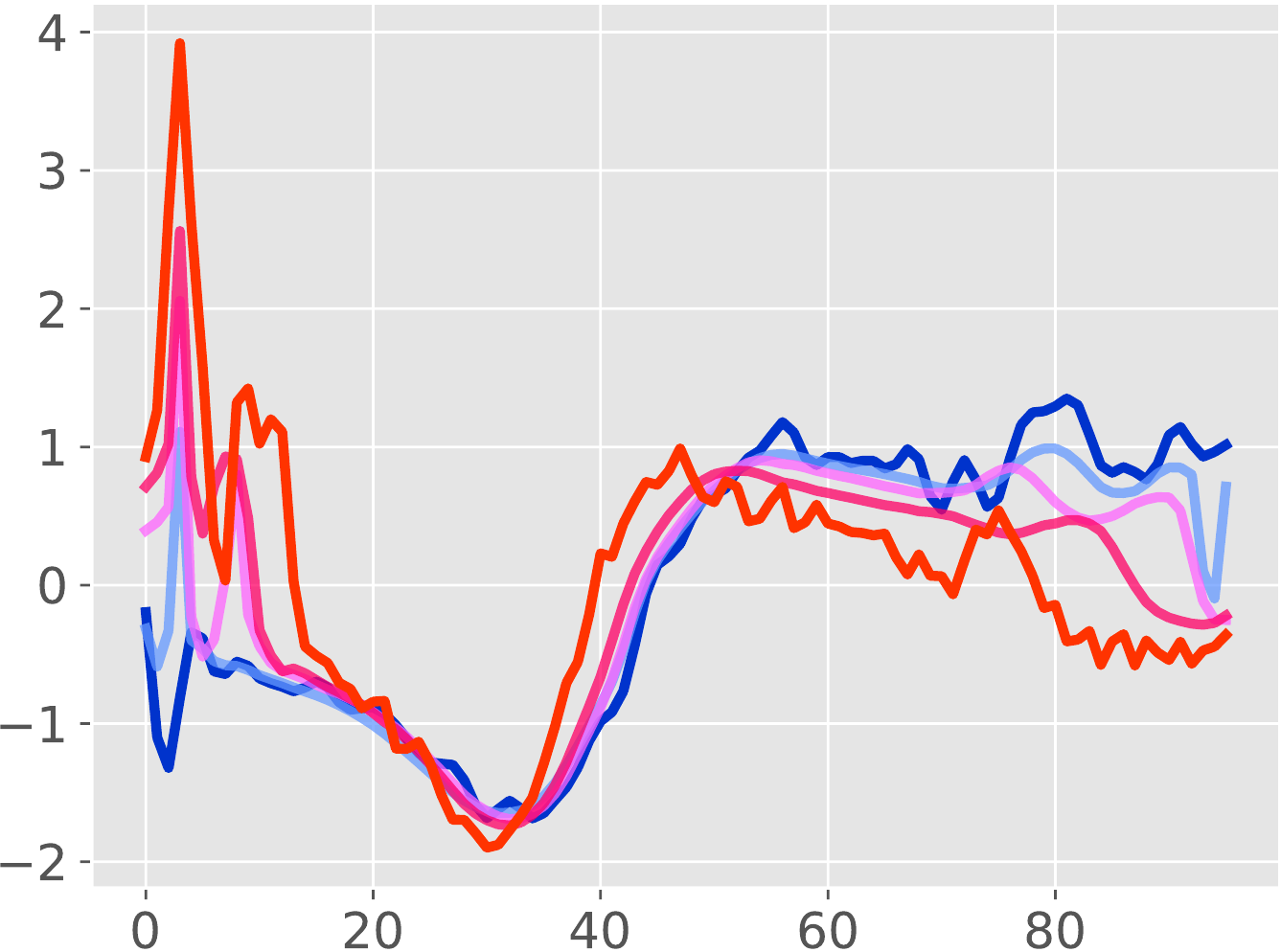}
}
%\subfigure[FISH]{
%\includegraphics[scale=0.45]{tmp/FISH_16_143_euc.pdf}
%\includegraphics[scale=0.45]{tmp/FISH_16_143_sdtw.pdf}
%}
\subfigure[ItalyPowerDemand]{
\includegraphics[scale=0.43]{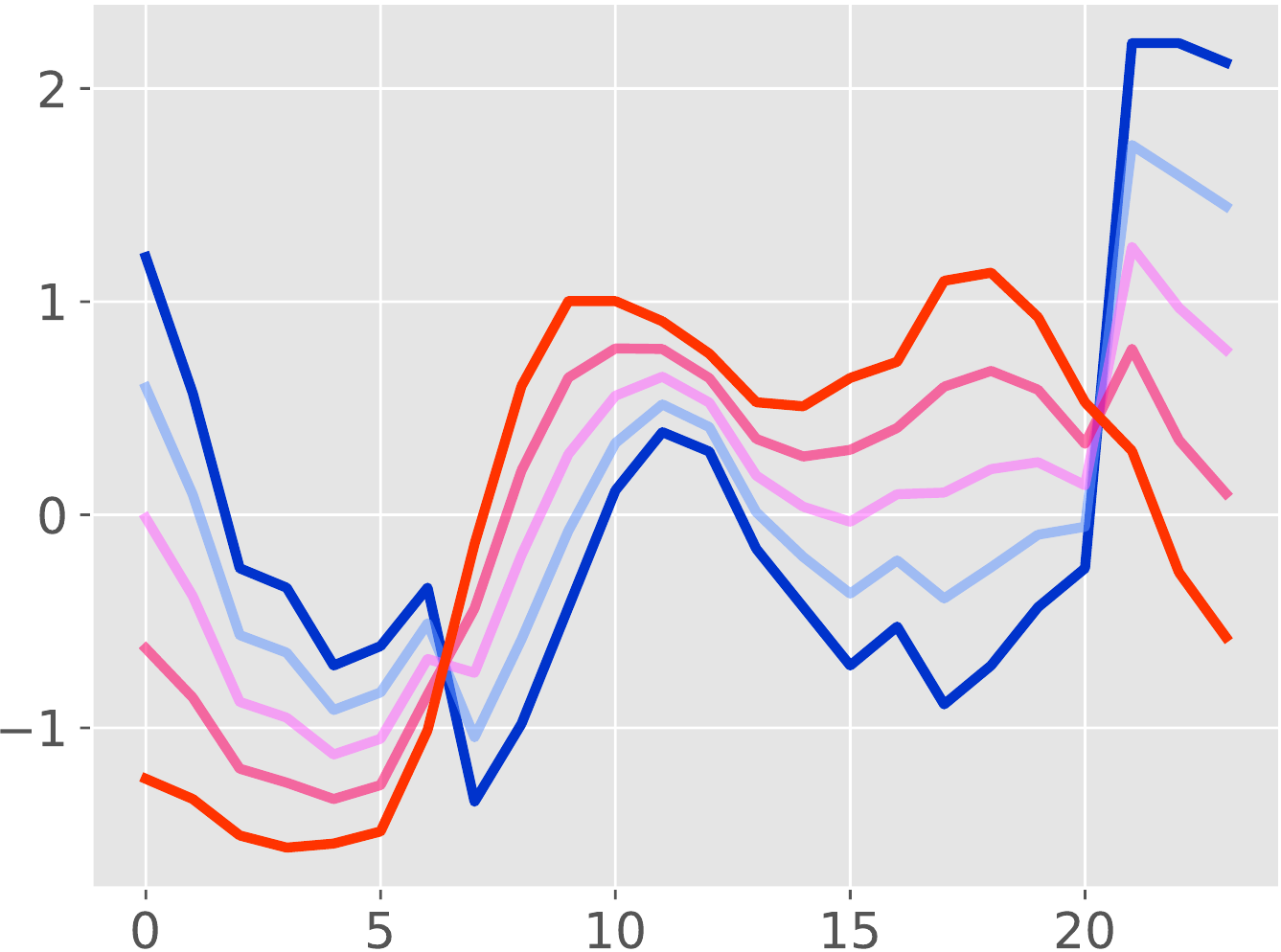}
\includegraphics[scale=0.43]{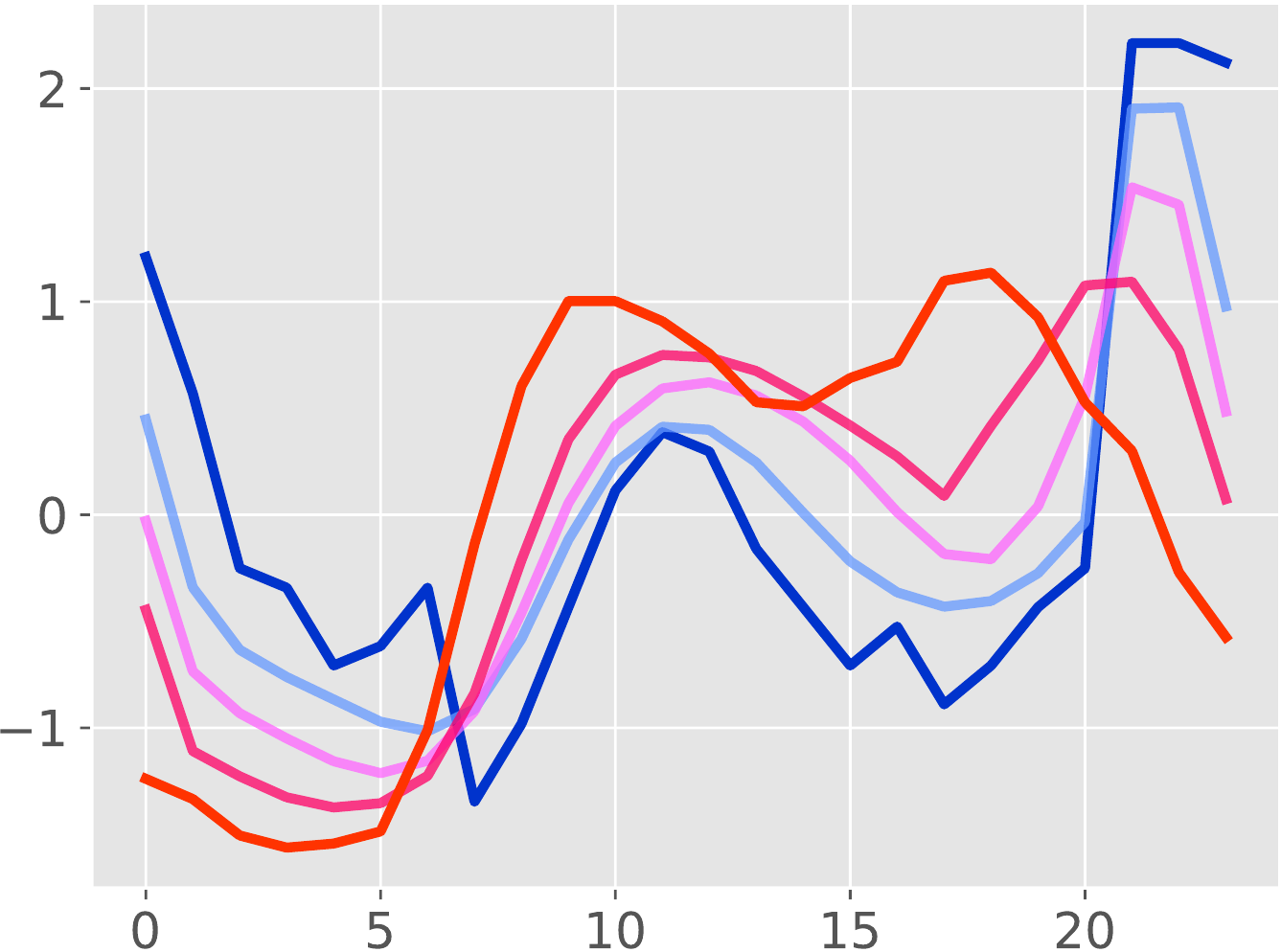}
}
\subfigure[TwoLeadECG]{
\includegraphics[scale=0.43]{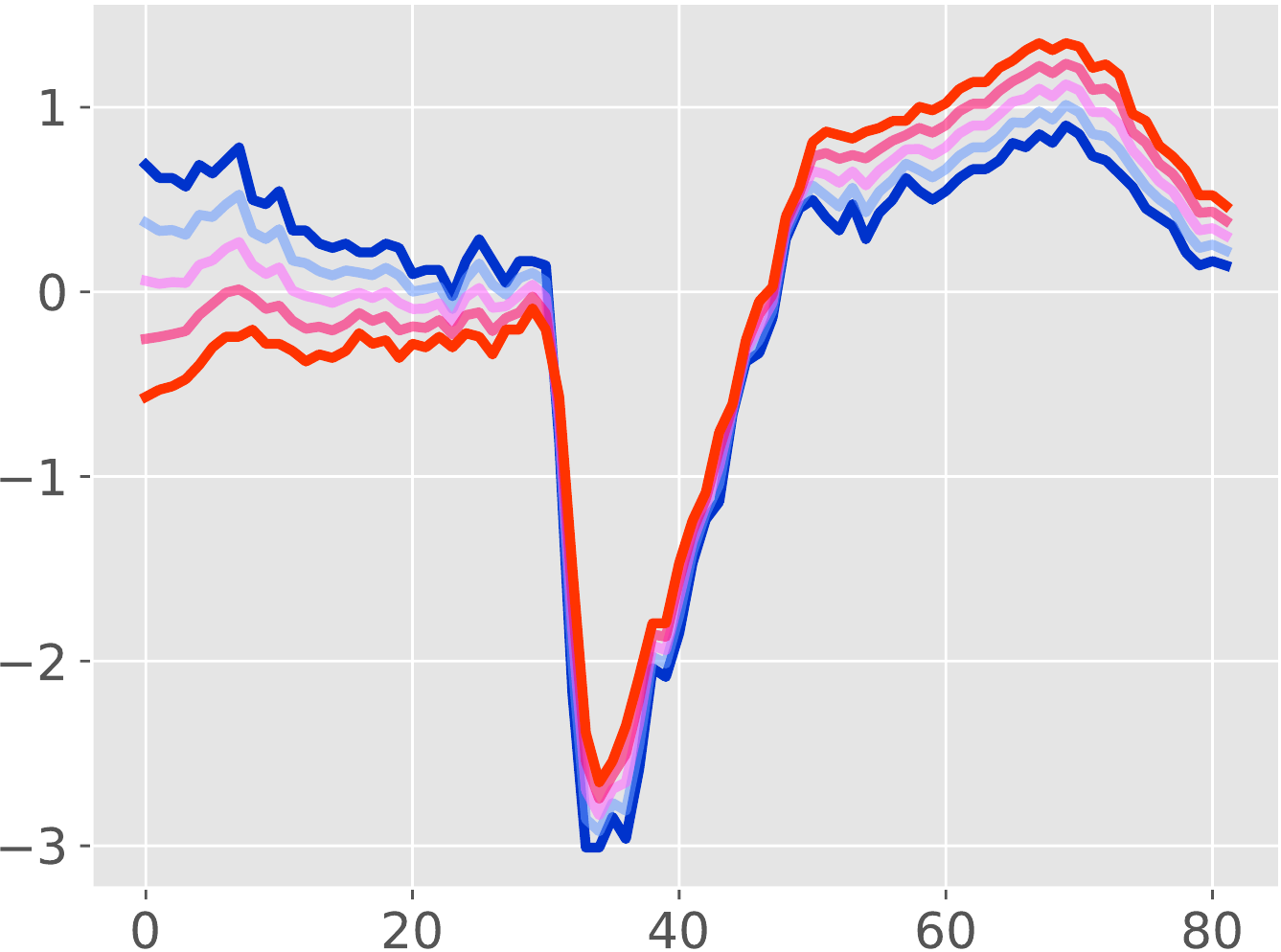}
\includegraphics[scale=0.43]{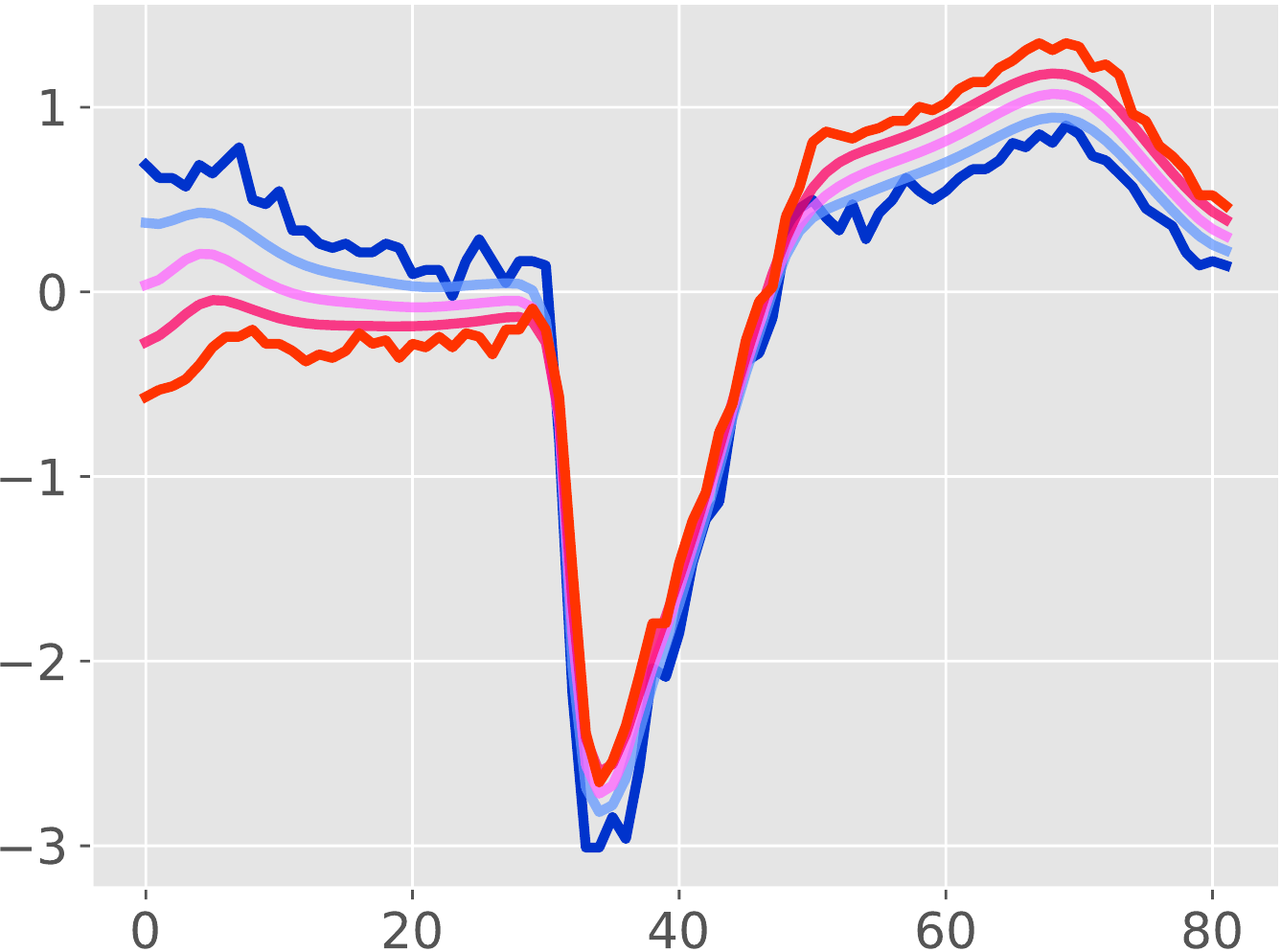}
}
%\caption{}
\end{figure}

\clearpage
\section{Clusters obtained by $k$-means under DTW or soft-DTW geometry}
\label{appendix:clustering_vis}

{\bf \large CBF dataset}

\setcounter{subfigure}{0}
\begin{figure}[H]
\centering
\subfigure[Soft-DTW ($\gamma=1$, random initialization)]{
\includegraphics[scale=0.40]{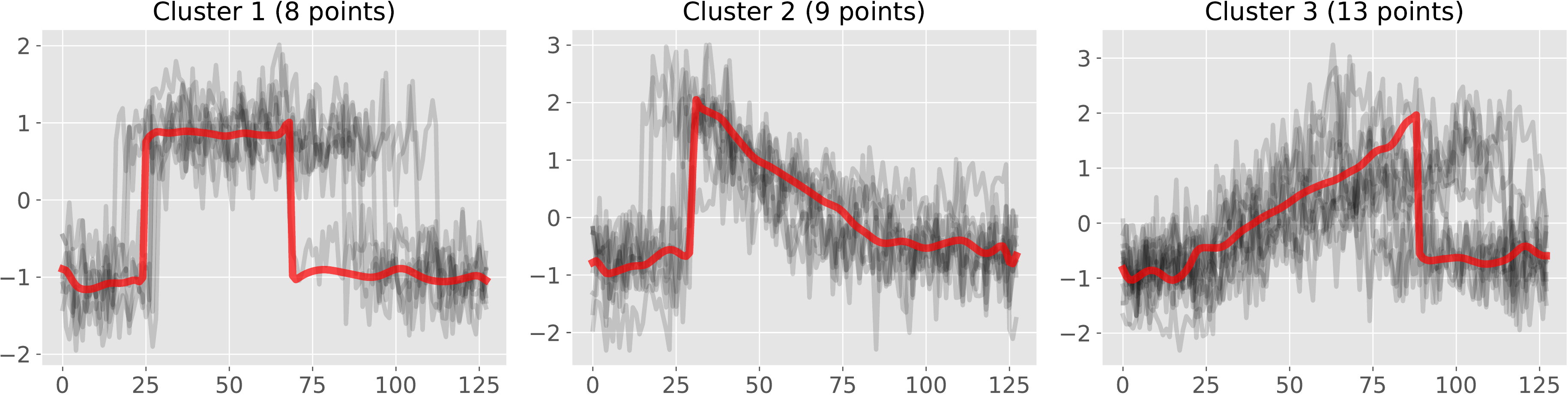} 
}
\subfigure[Soft-DTW ($\gamma=1$, Euclidean mean initialization)]{
\includegraphics[scale=0.40]{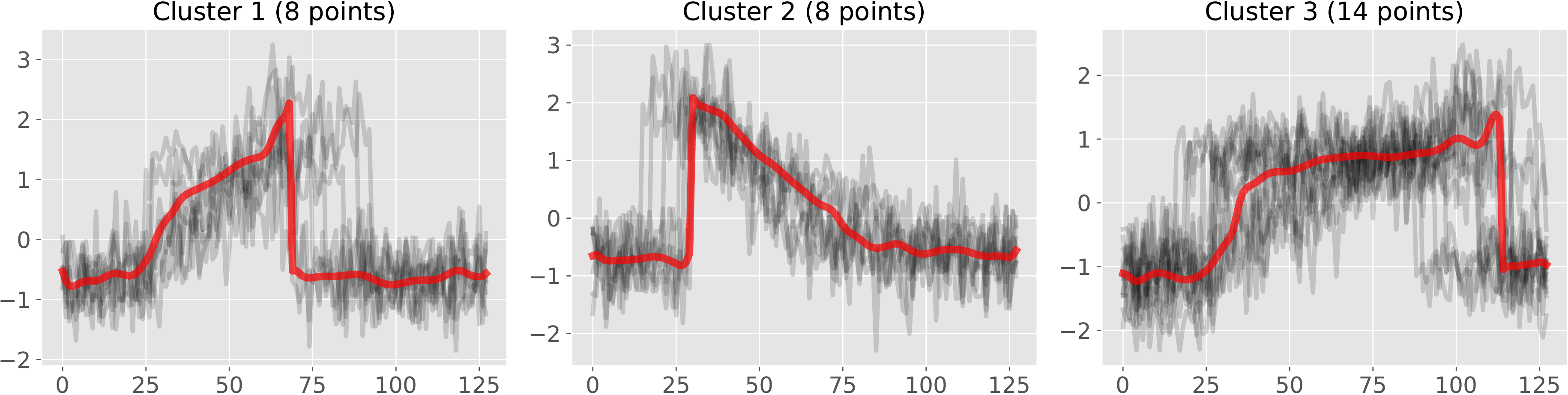} 
}

\vspace{1cm}

\subfigure[DBA (random initialization)]{
\includegraphics[scale=0.40]{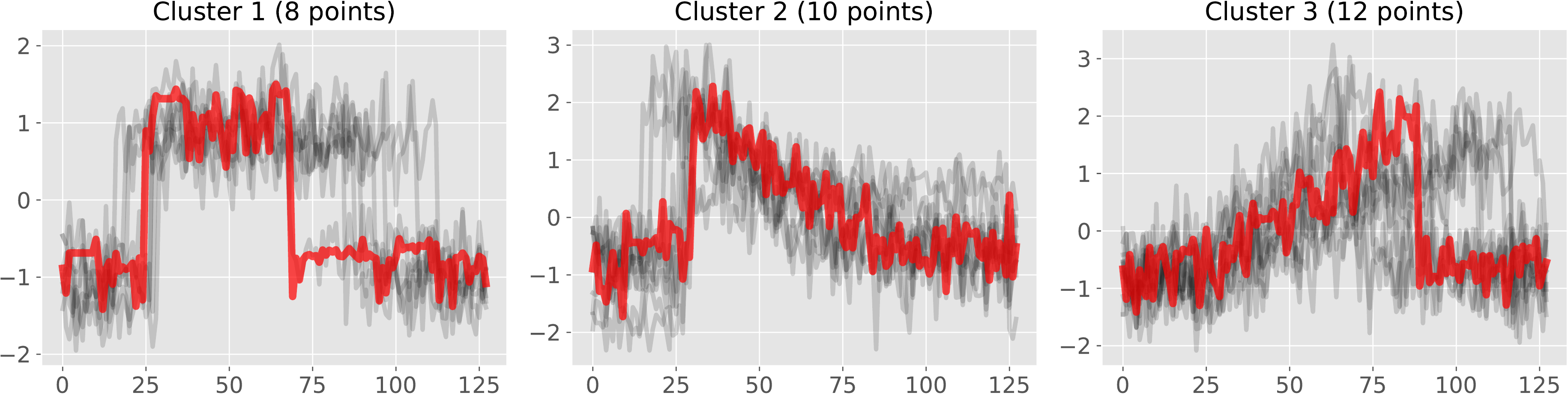}
}
\subfigure[DBA (Euclidean mean initialization)]{
\includegraphics[scale=0.40]{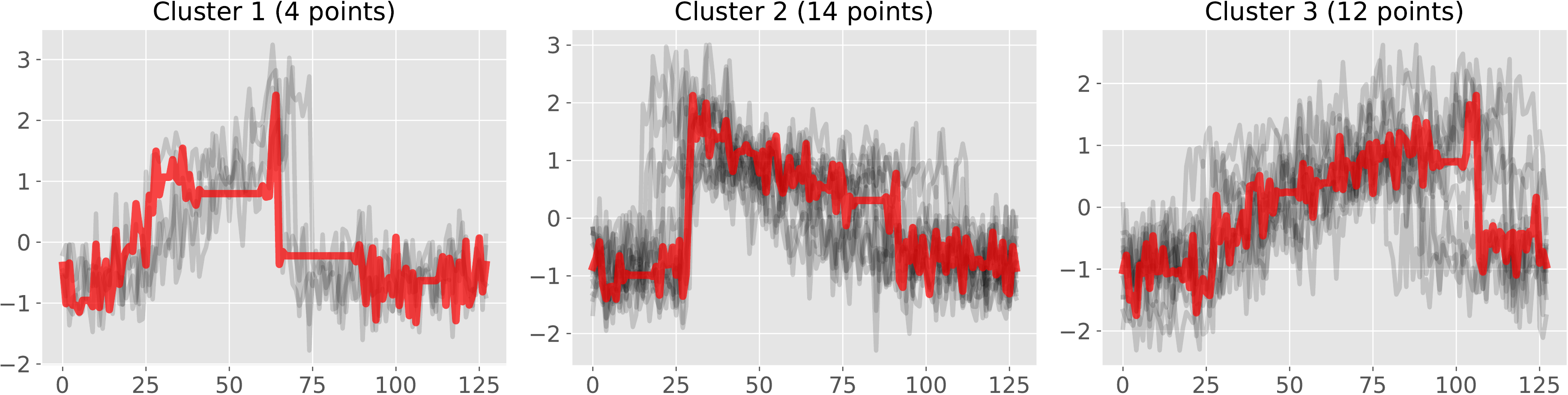}
}
%\caption{}
\end{figure}

\newpage
{\bf \large ECG200 dataset}

\setcounter{subfigure}{0}
\begin{figure}[H]
\centering
\subfigure[Soft-DTW ($\gamma=1$, random initialization)]{
\includegraphics[scale=0.45]{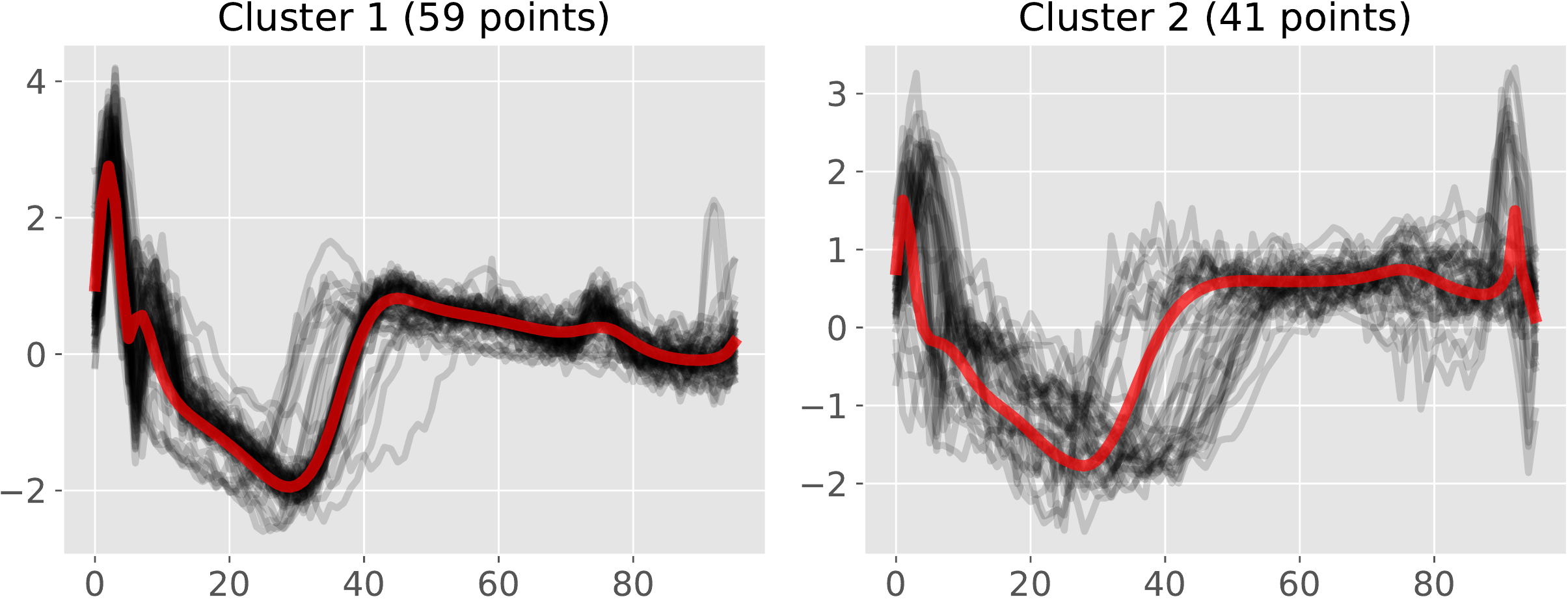} 
}
\subfigure[Soft-DTW ($\gamma=1$, Euclidean mean initialization)]{
\includegraphics[scale=0.45]{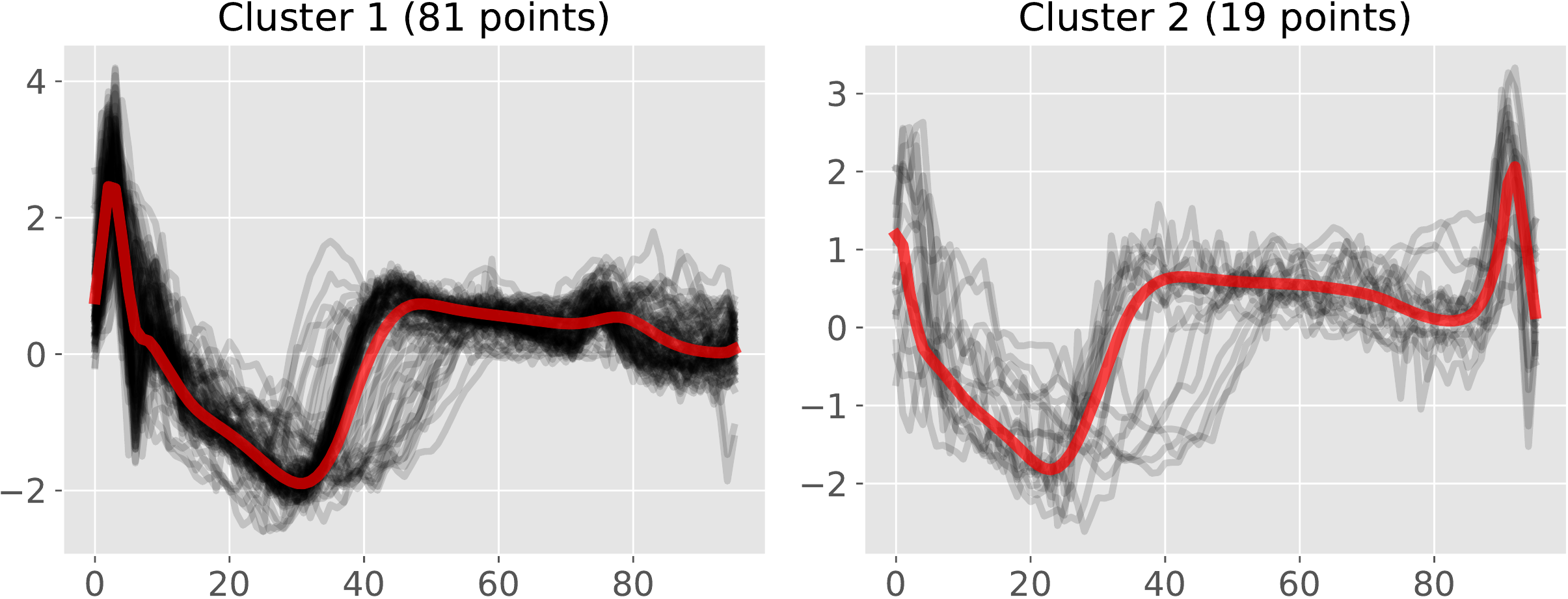} 
}

\vspace{1cm}

\subfigure[DBA (random initialization)]{
\includegraphics[scale=0.45]{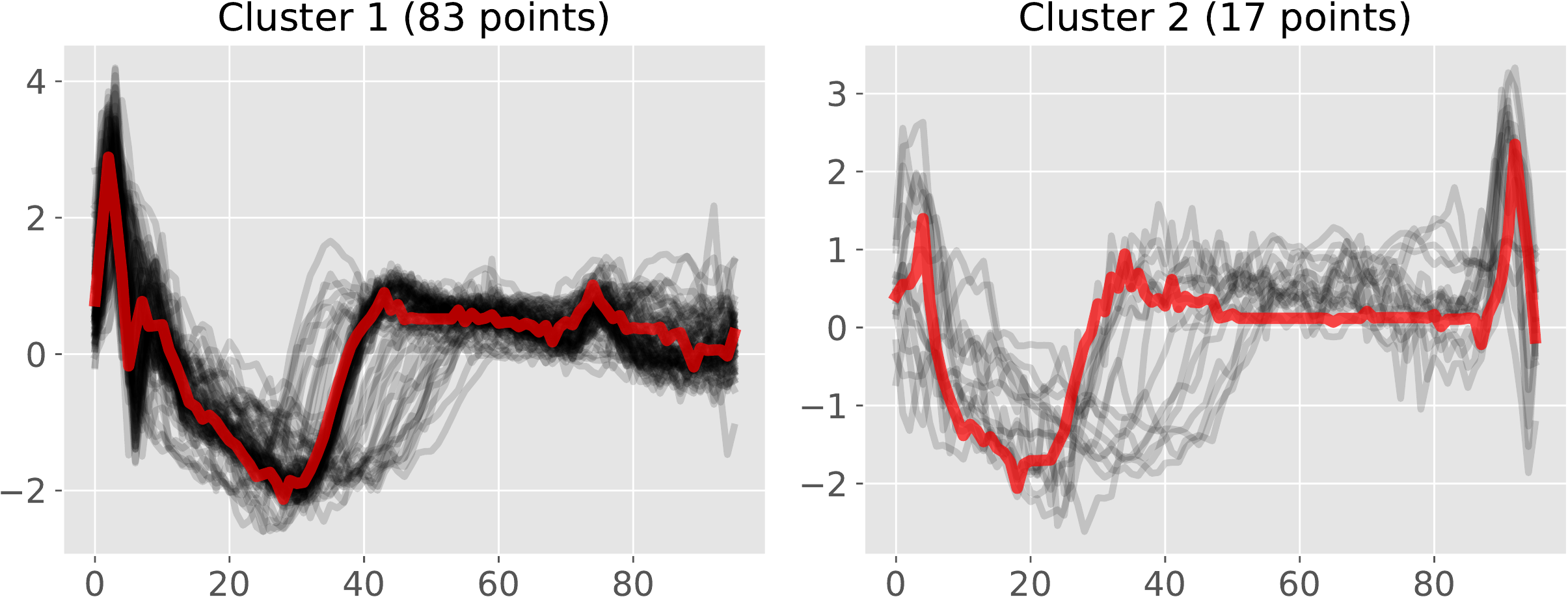}
}
\subfigure[DBA (Euclidean mean initialization)]{
\includegraphics[scale=0.45]{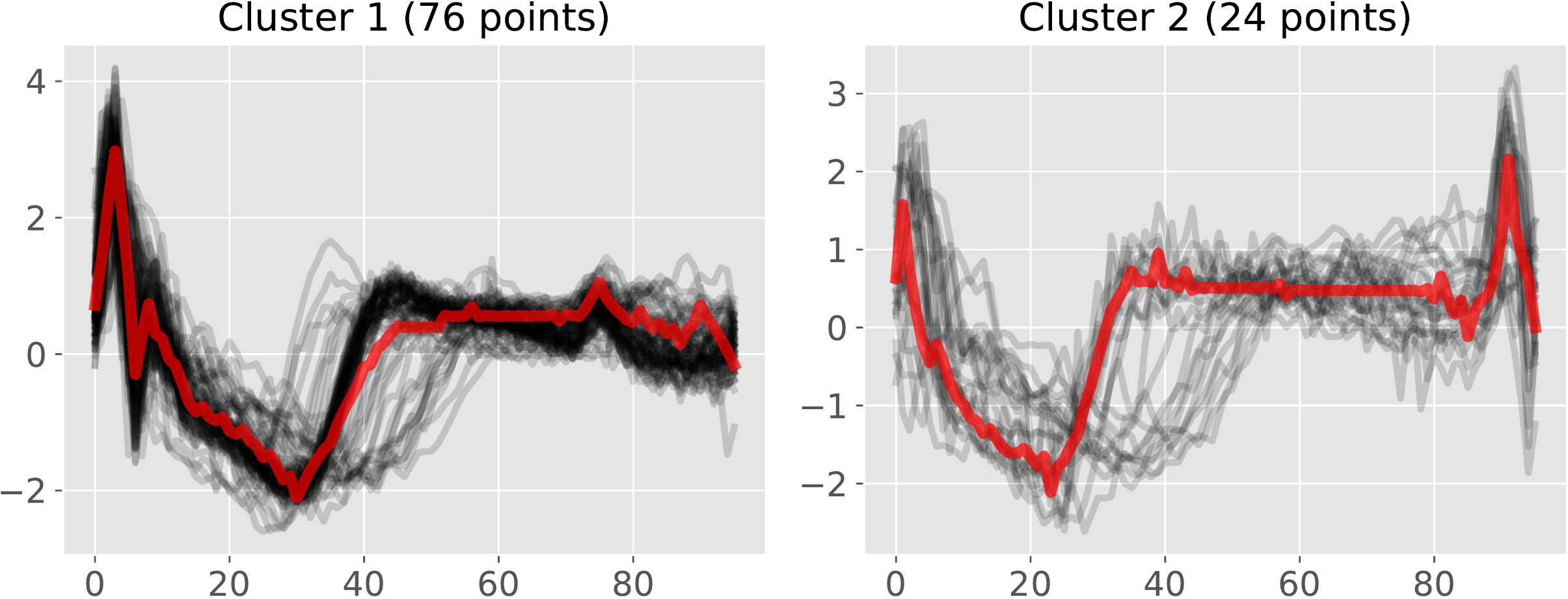}
}
%\caption{}
\end{figure}

\section{More visualizations of time-series prediction}
\label{appendix:more_ts_pred}

\setcounter{subfigure}{0}
\begin{figure}[H]
\centering
\subfigure[CBF]{
\includegraphics[scale=0.30]{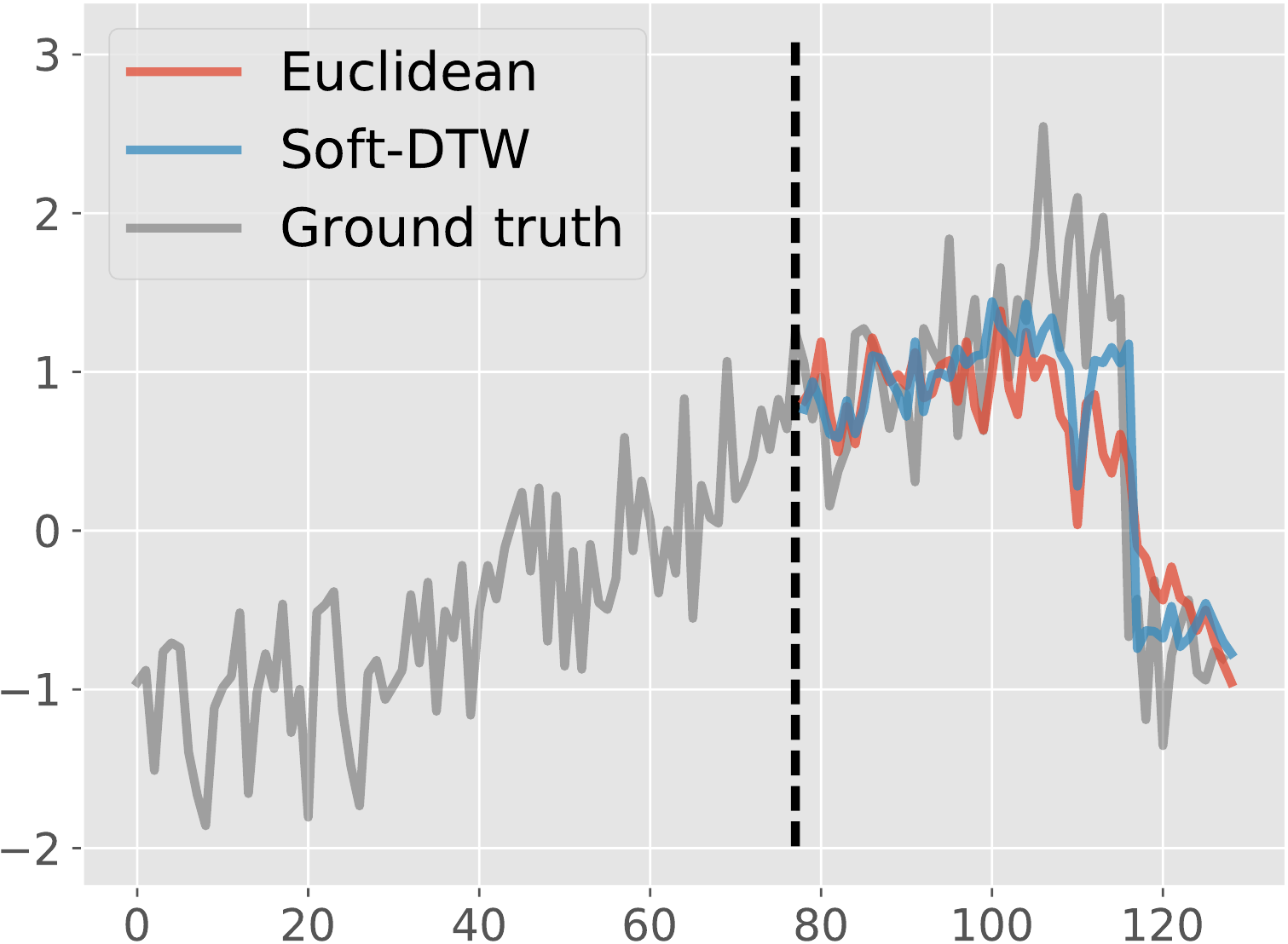} 
\includegraphics[scale=0.30]{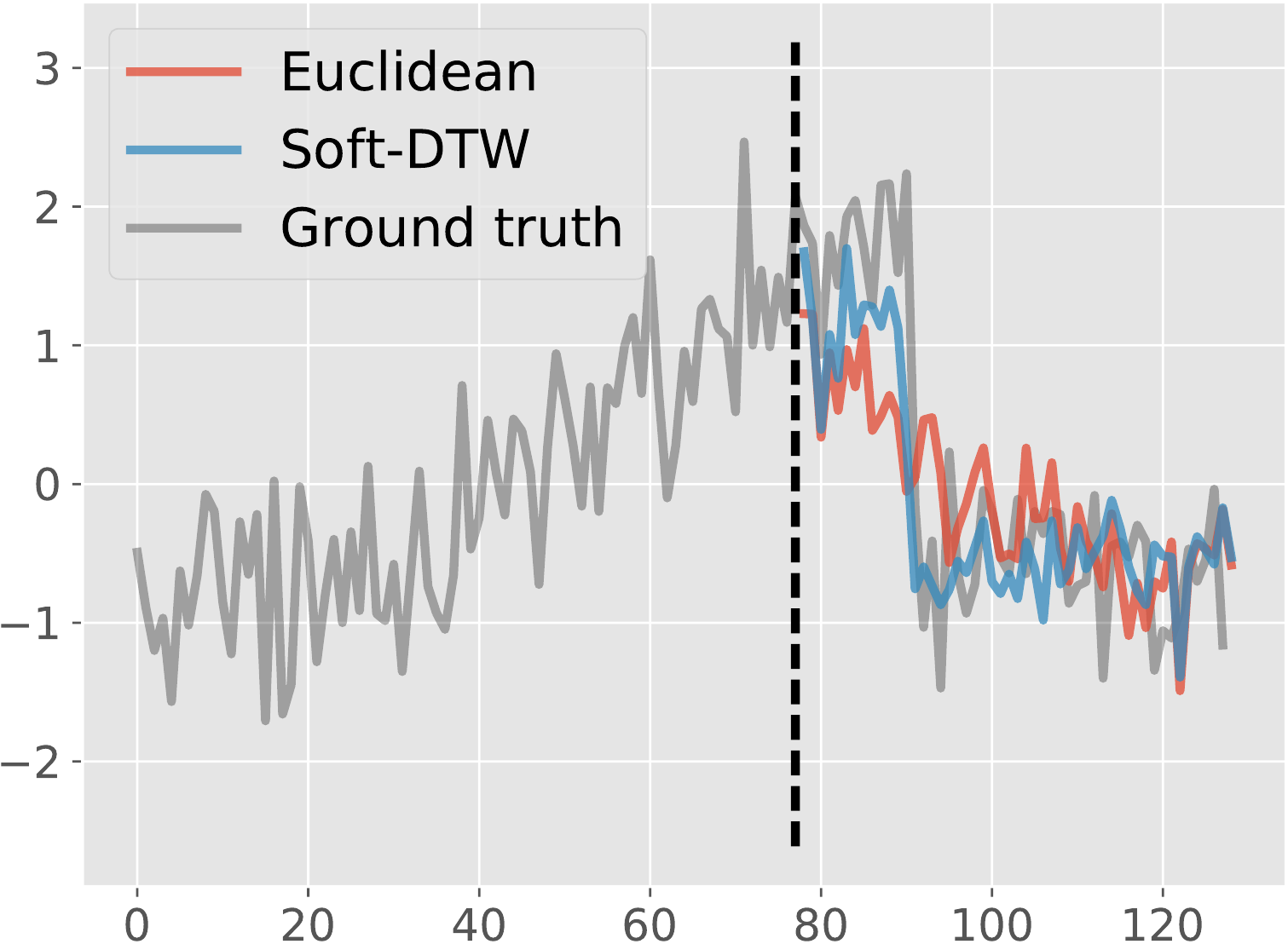} 
\includegraphics[scale=0.30]{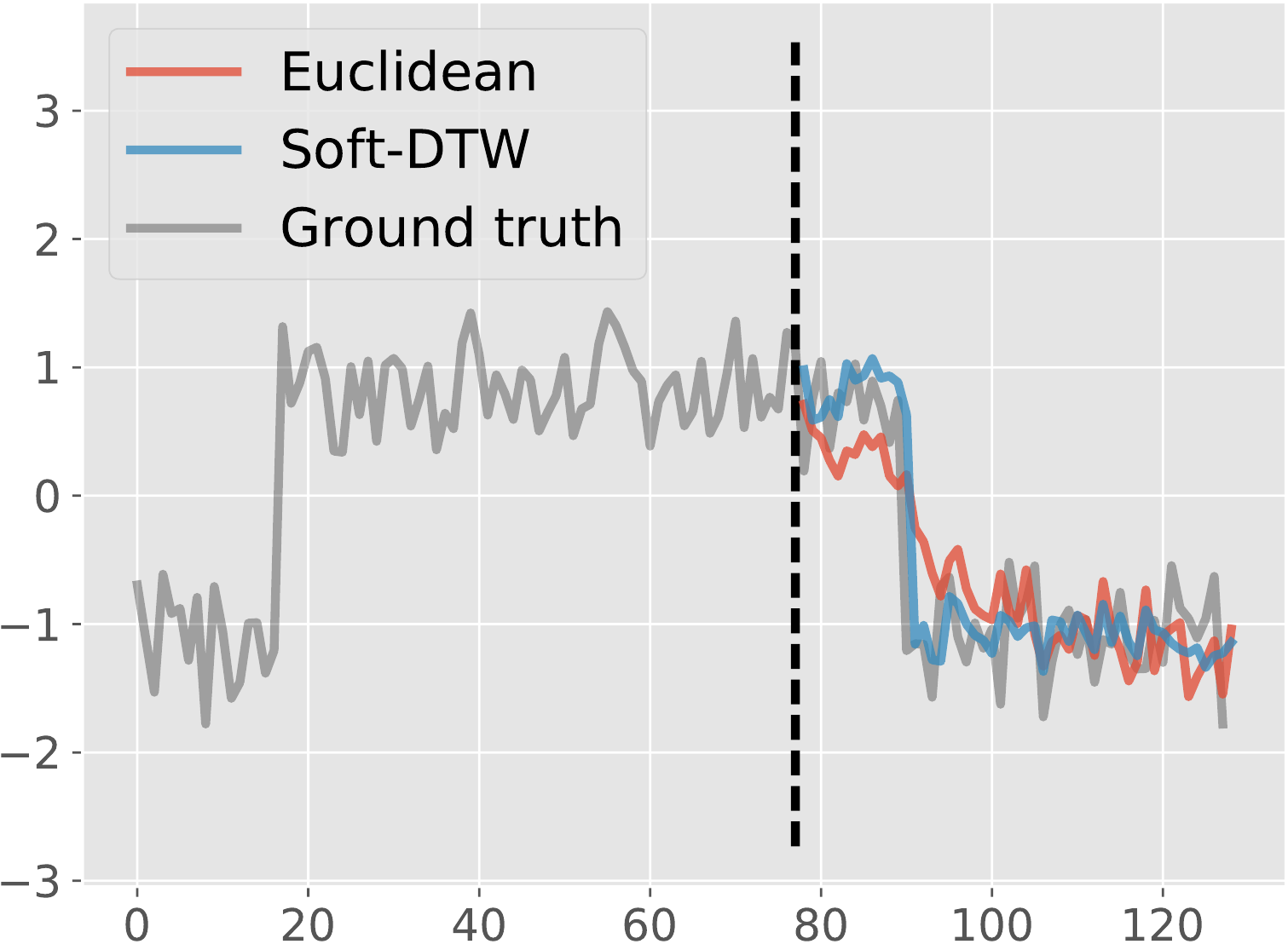} 
}
\subfigure[ECG200]{
\includegraphics[scale=0.30]{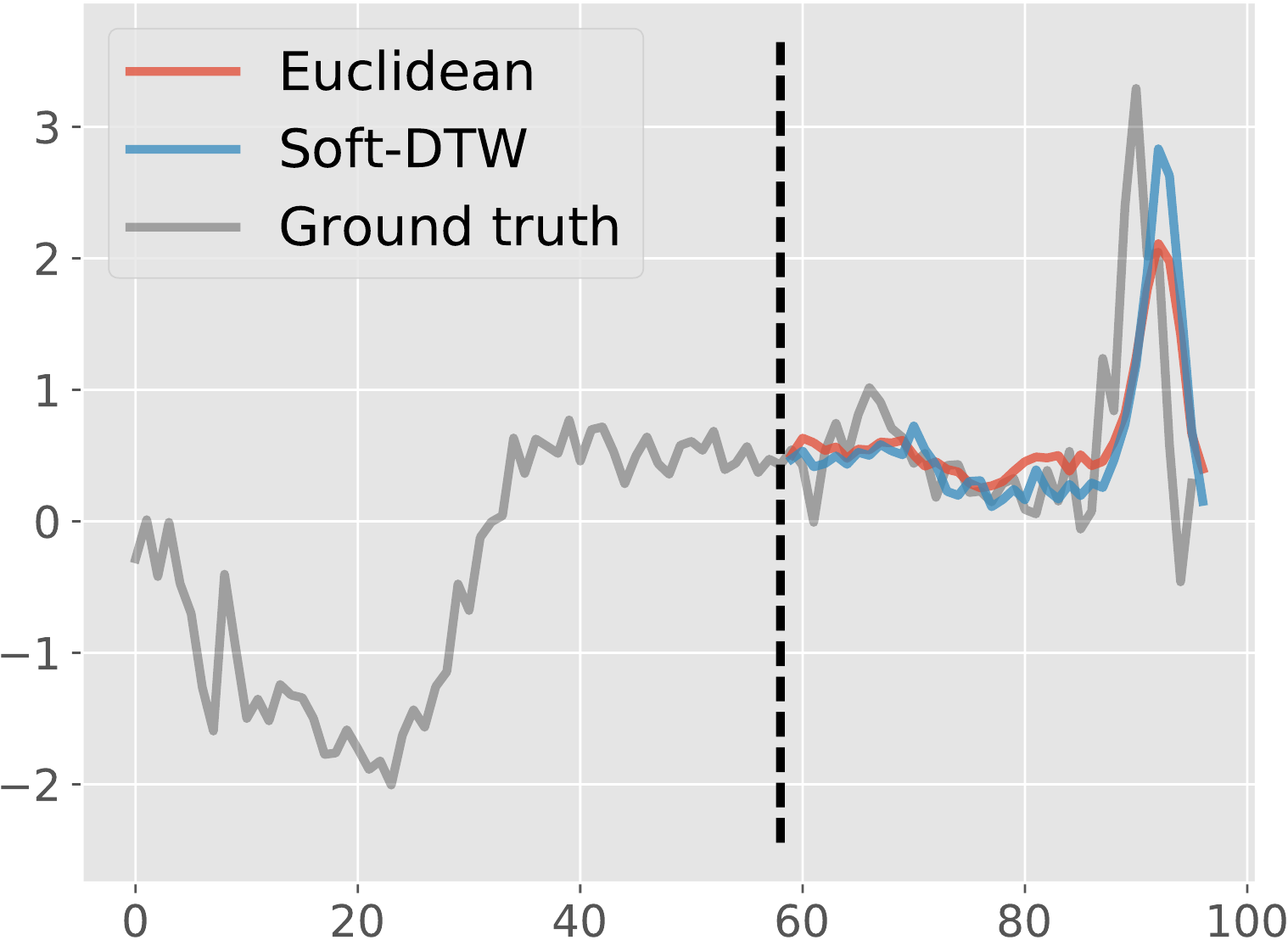} 
\includegraphics[scale=0.30]{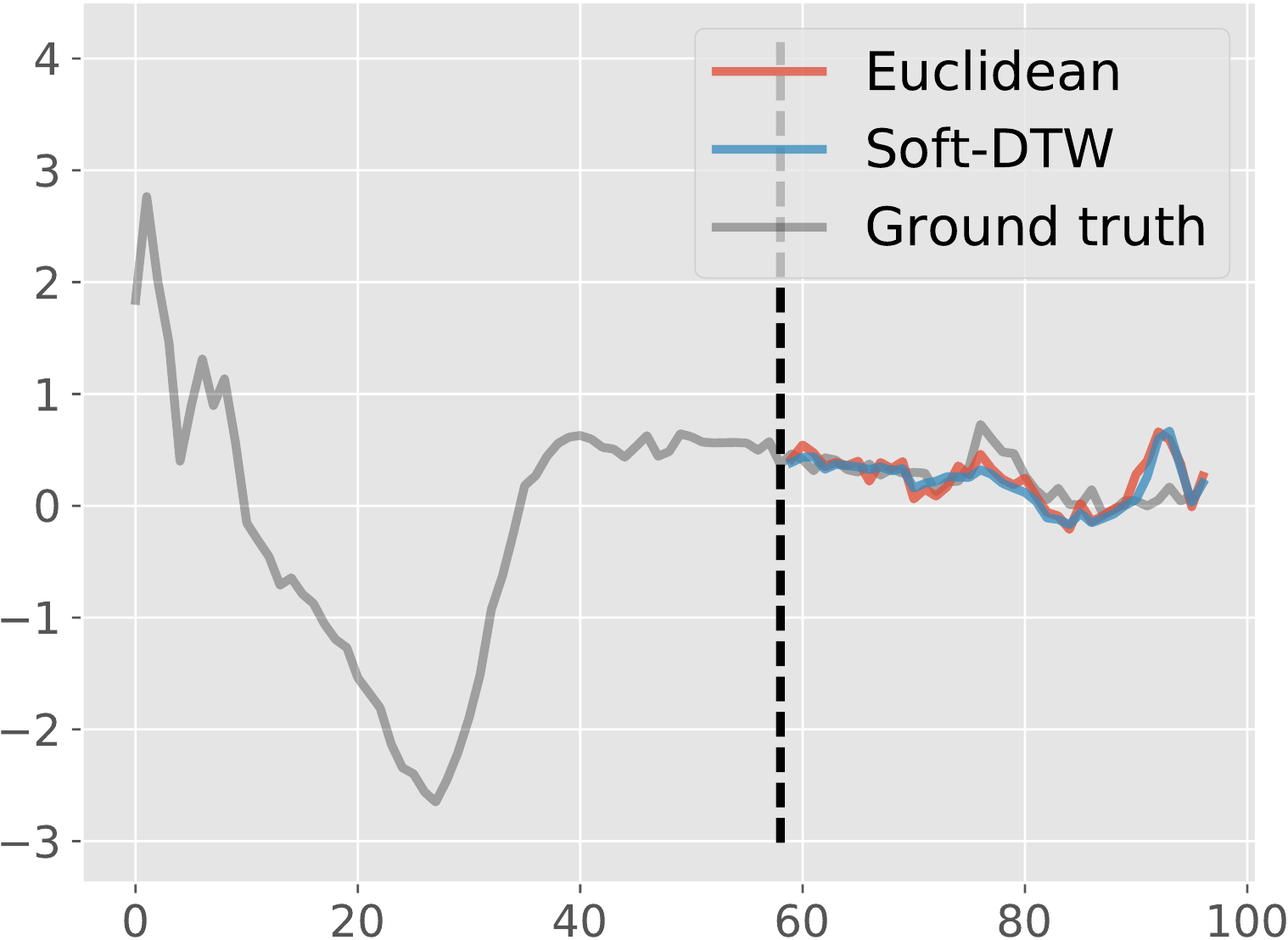} 
\includegraphics[scale=0.30]{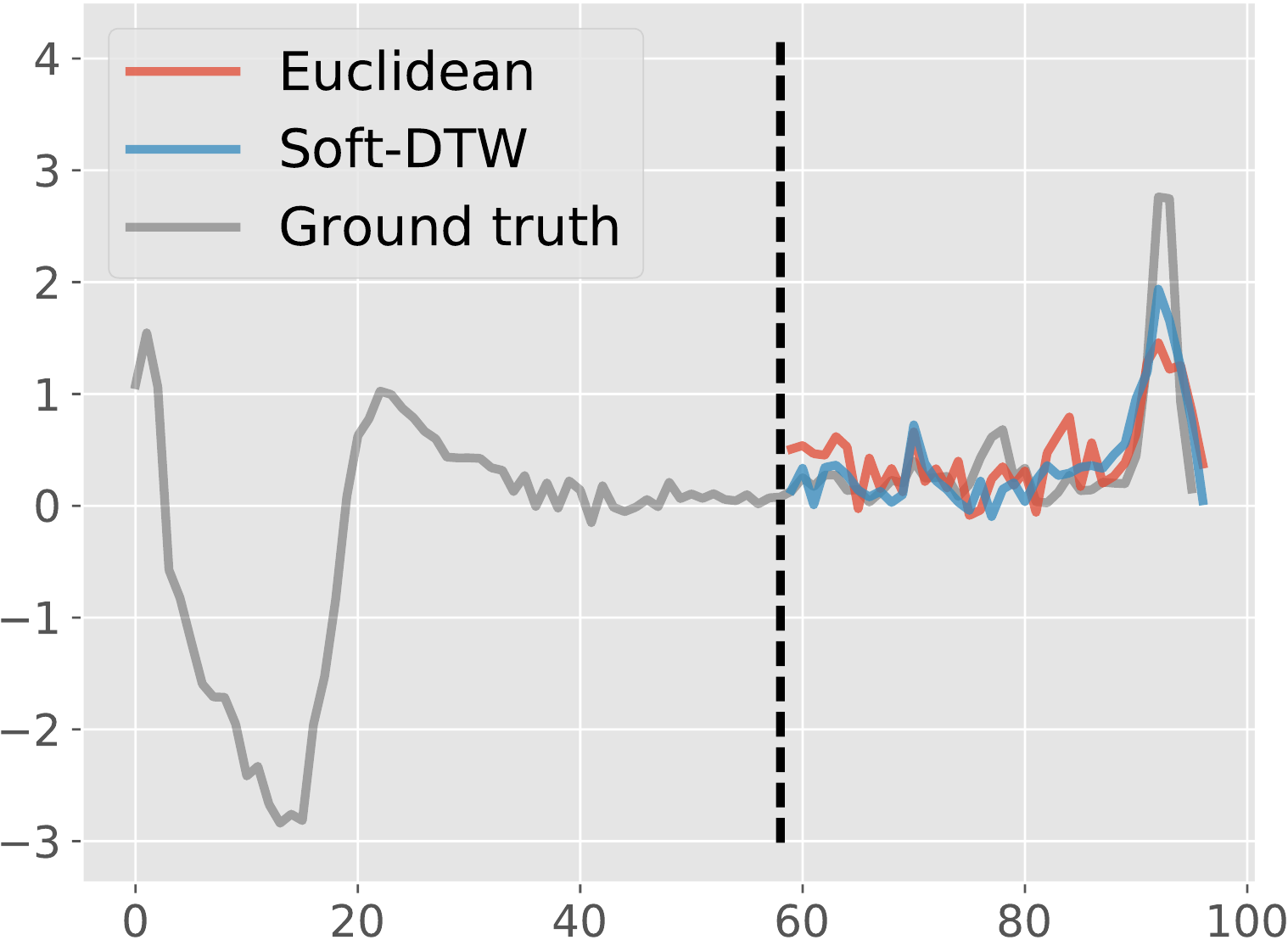} 
}
\subfigure[ECG5000]{
\includegraphics[scale=0.30]{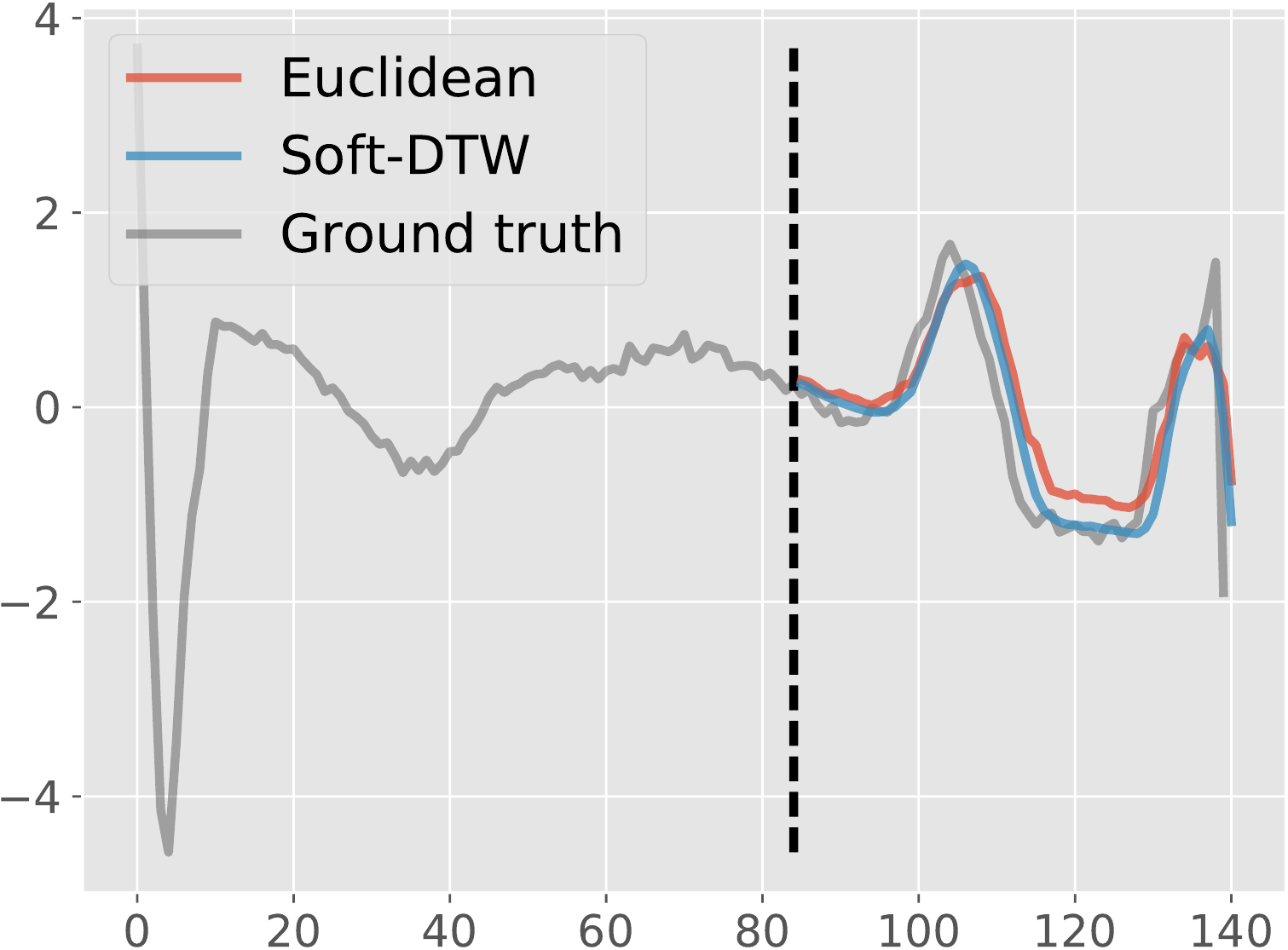} 
\includegraphics[scale=0.30]{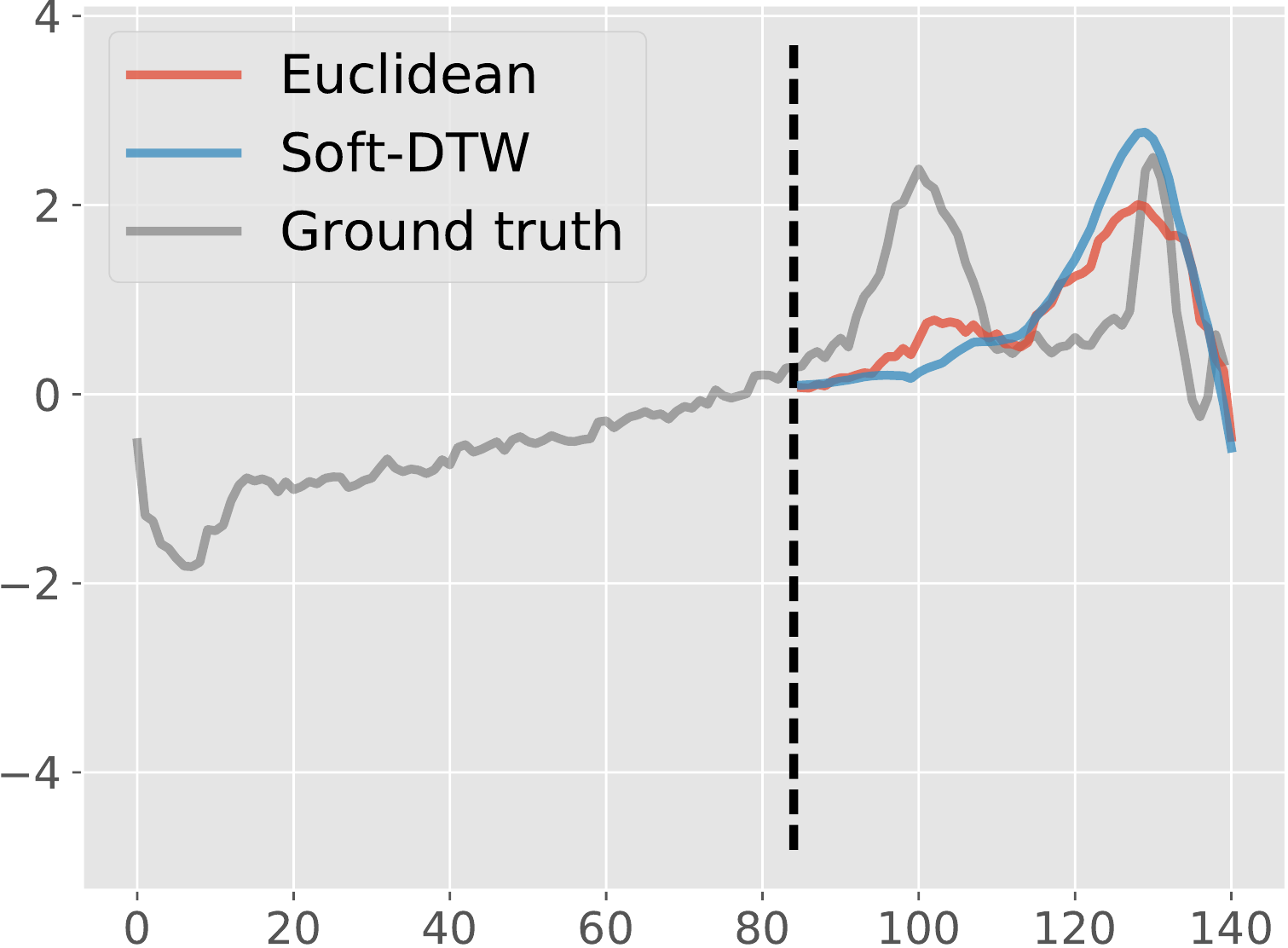} 
\includegraphics[scale=0.30]{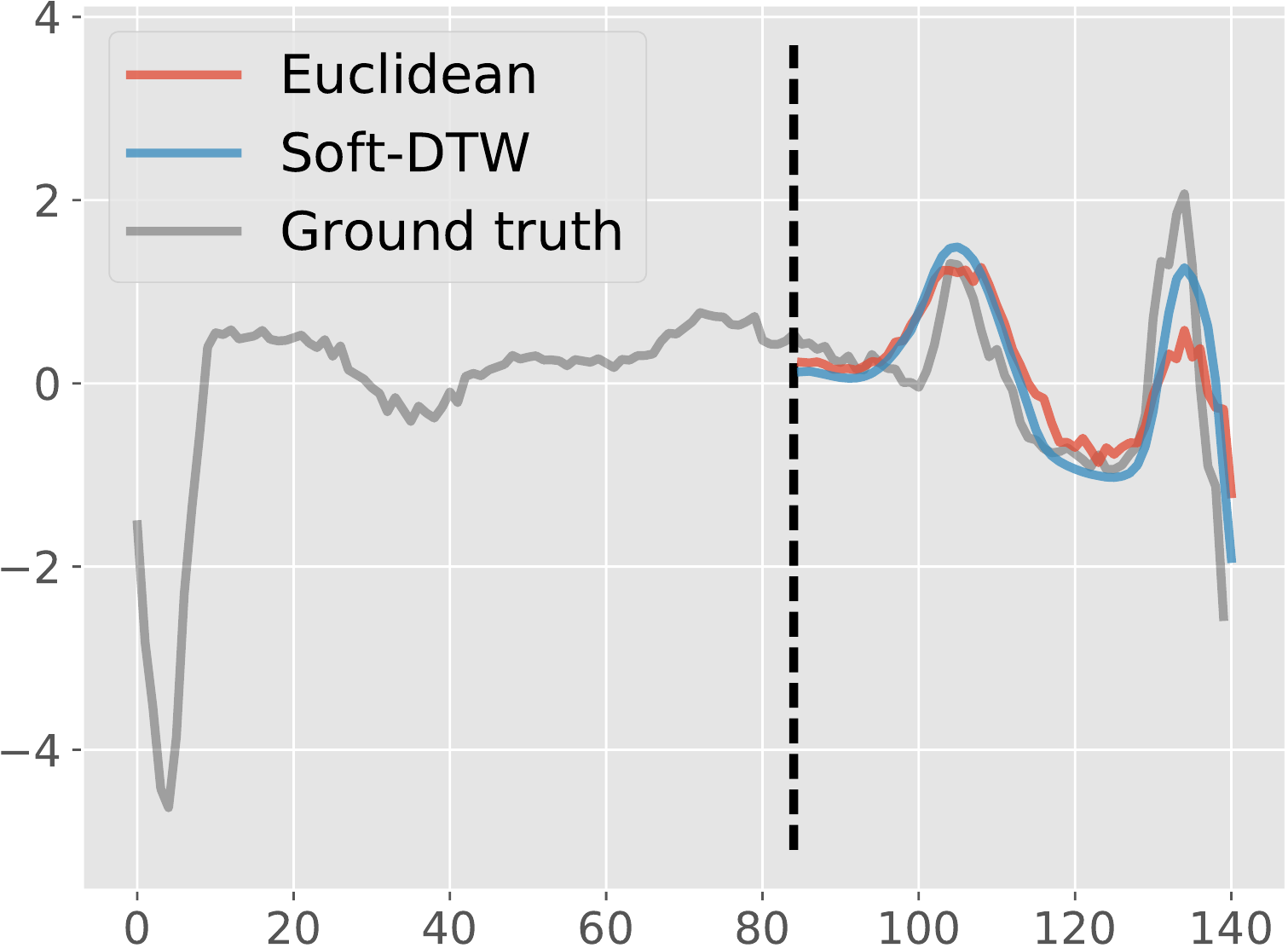} 
}
\subfigure[ShapesAll]{
\includegraphics[scale=0.30]{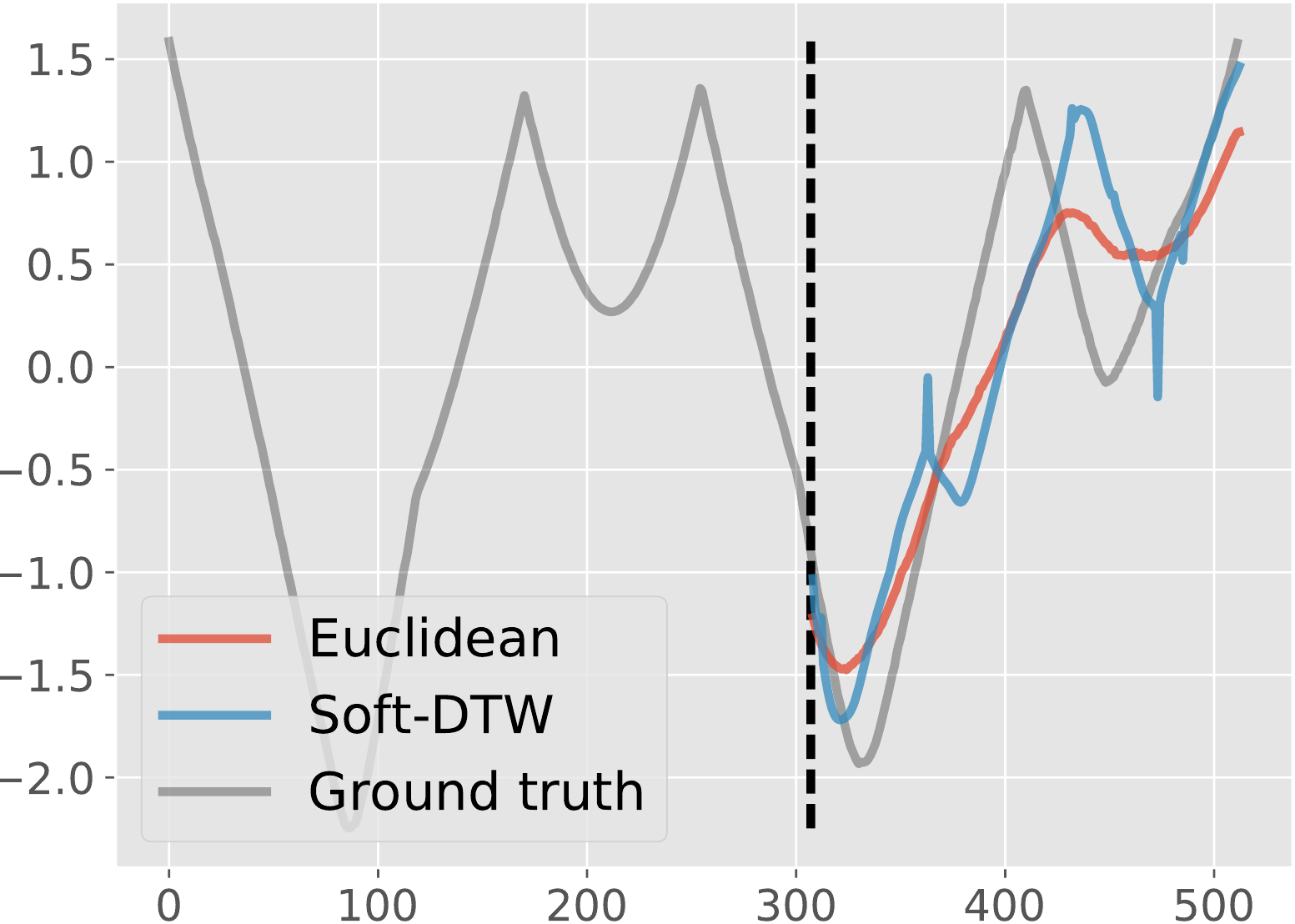} 
\includegraphics[scale=0.30]{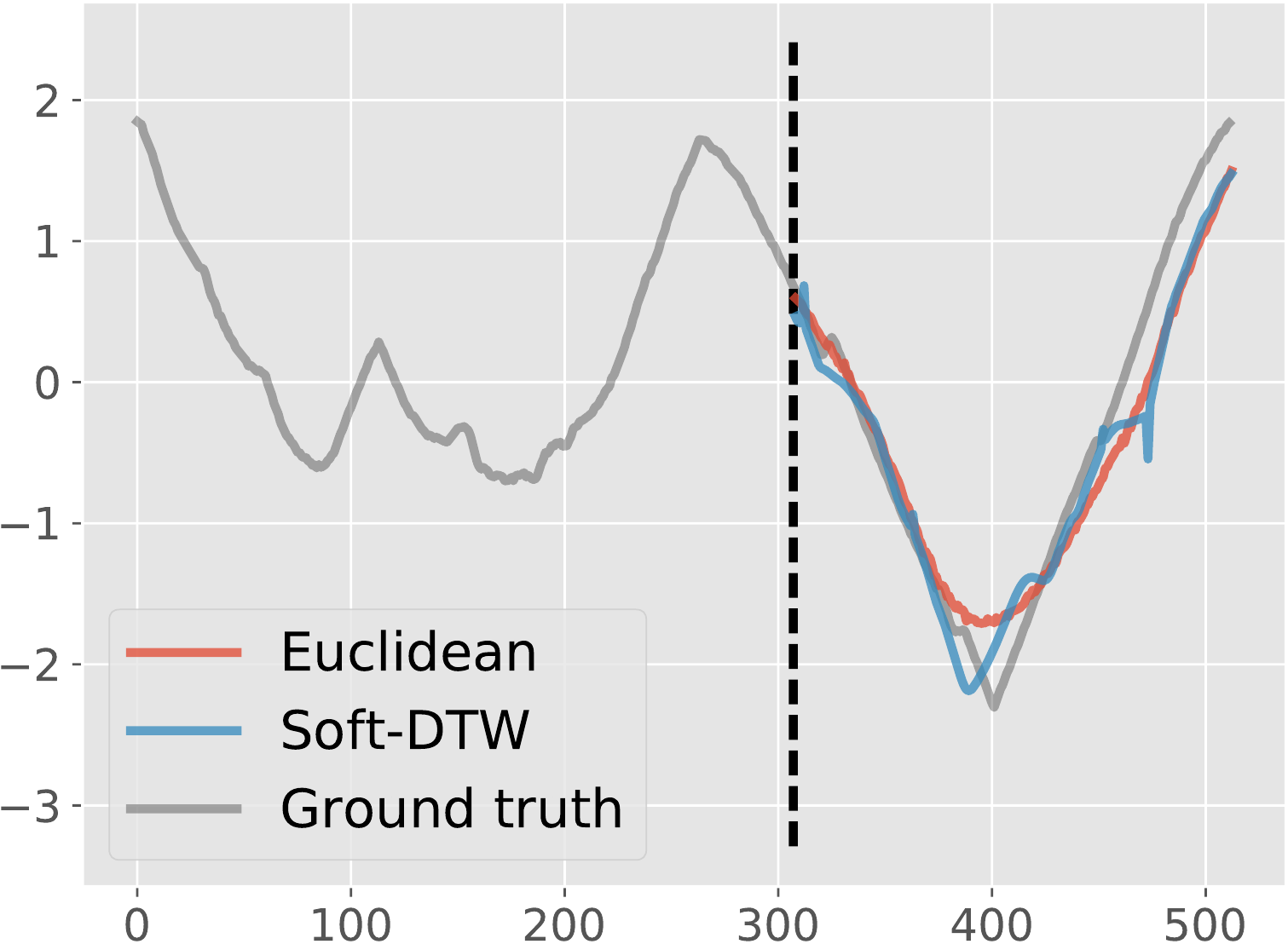} 
\includegraphics[scale=0.30]{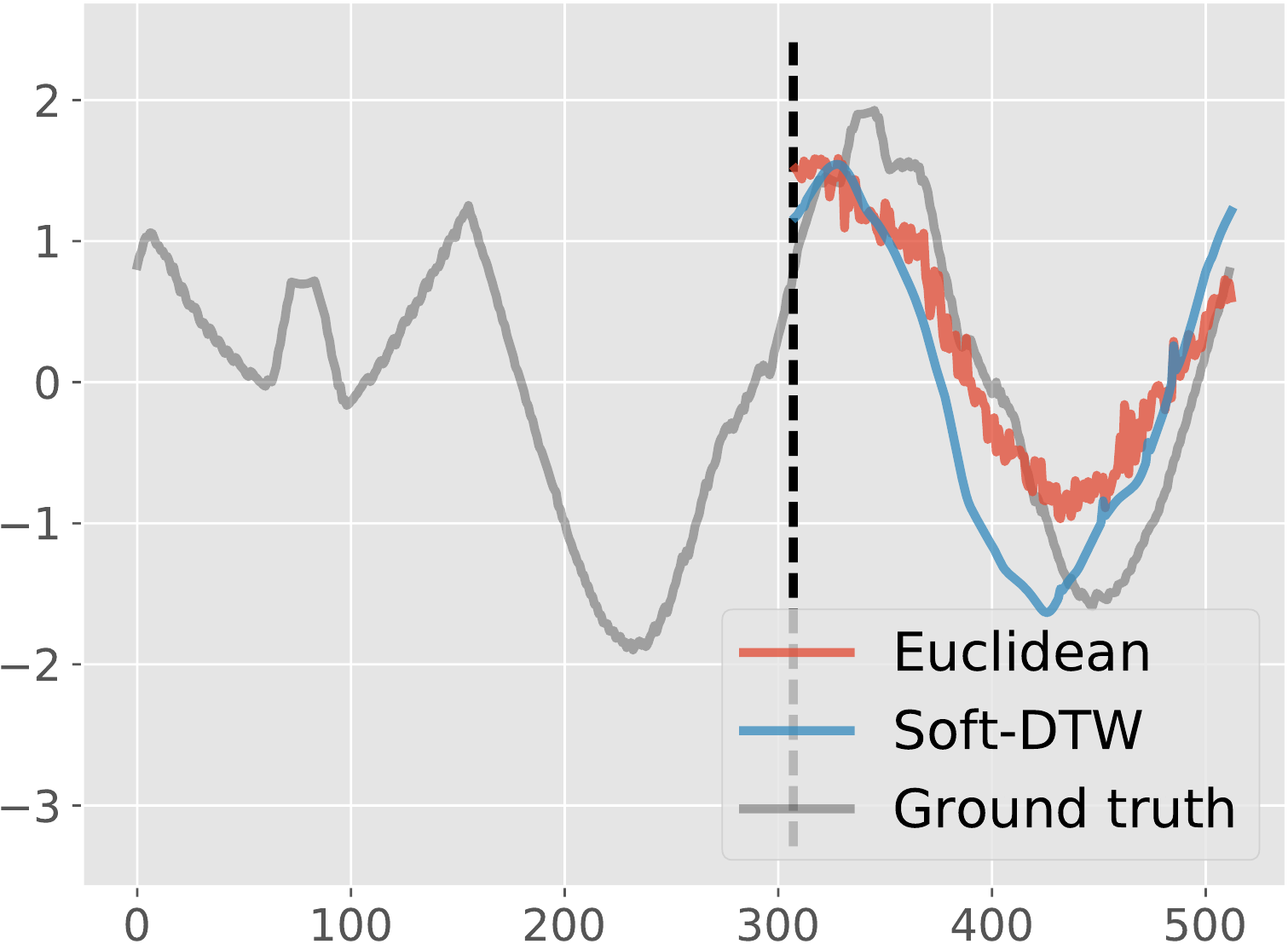} 
}
\subfigure[uWaveGestureLibrary\_Y]{
\includegraphics[scale=0.30]{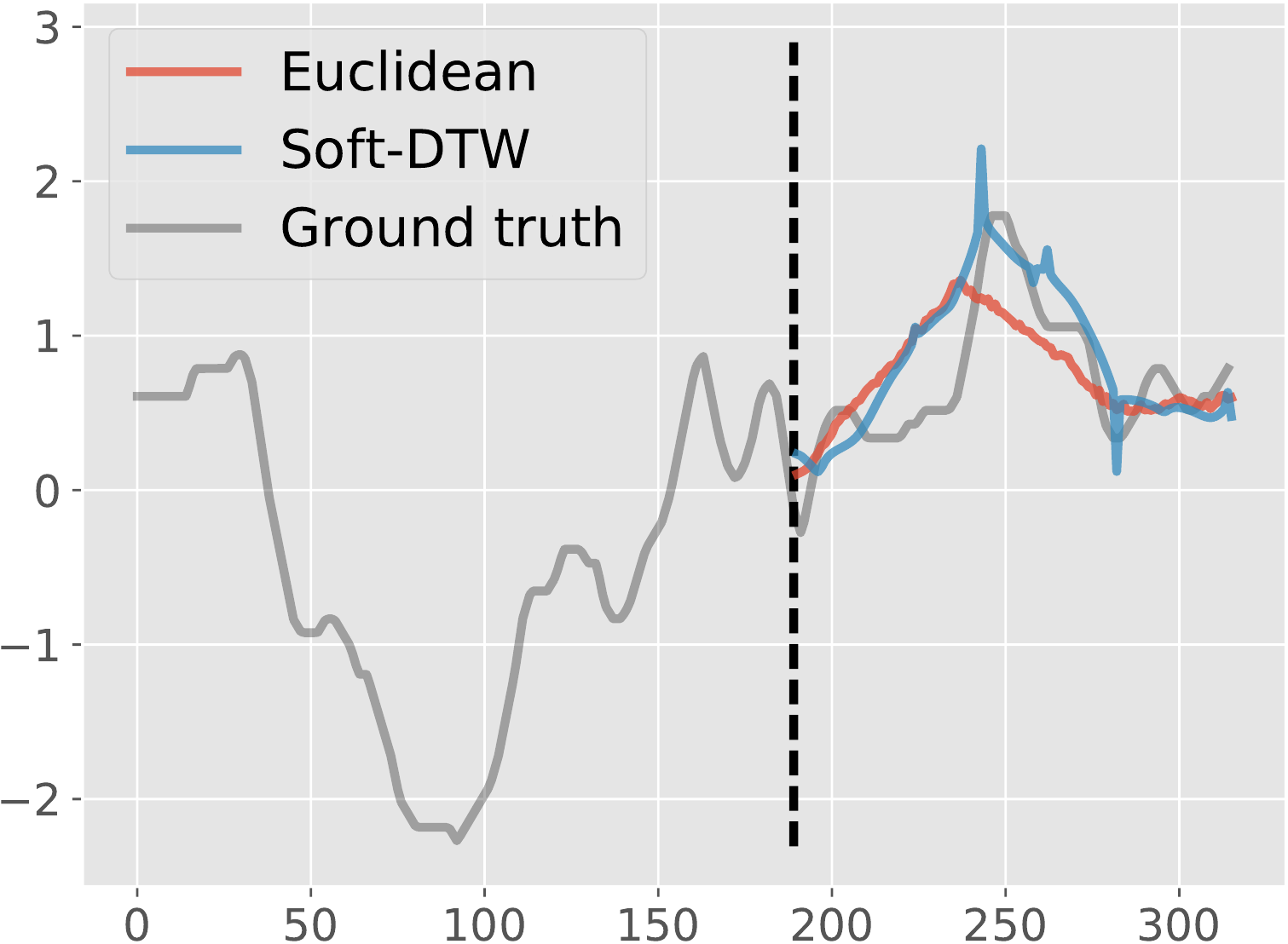} 
\includegraphics[scale=0.30]{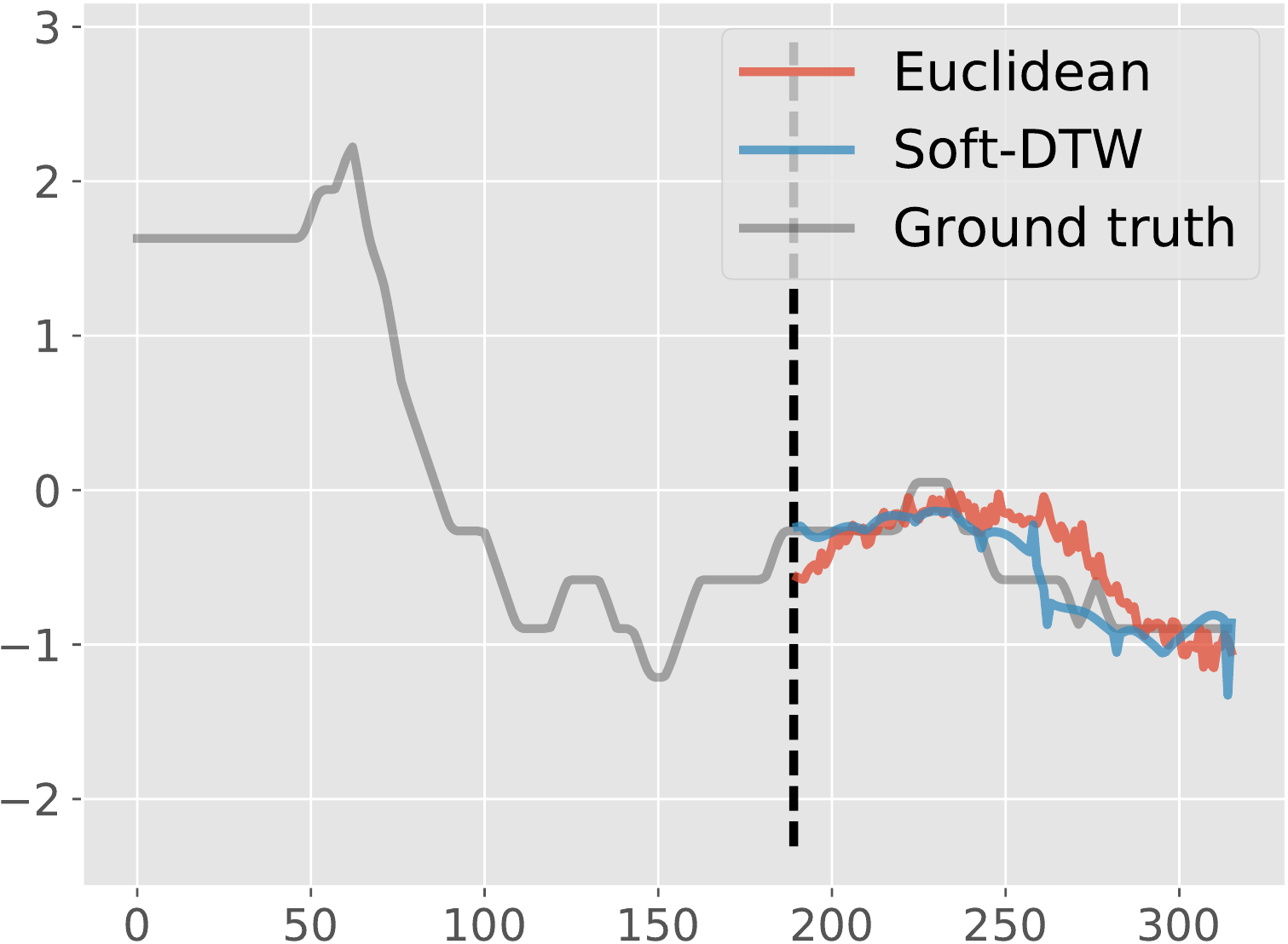} 
\includegraphics[scale=0.30]{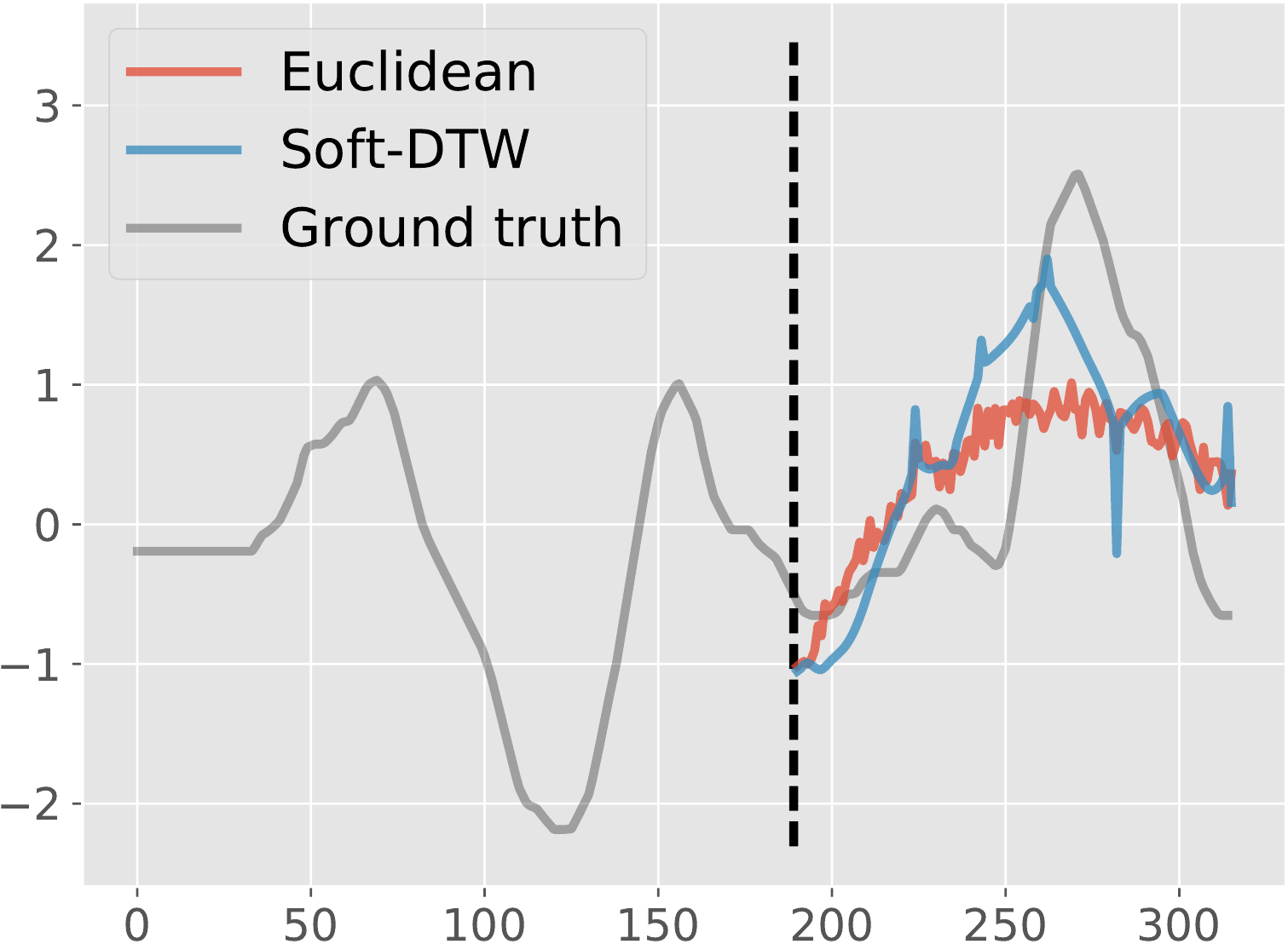} 
}
%\caption{}
\end{figure}

\clearpage

\section{Barycenters: DTW loss (Eq. \ref{eq:barycenter_obj} with $\gamma=0$)
achieved with random init}
\label{appendix:barycenter_random}

\begin{table}[H]
%\scriptsize
\fontsize{8}{7}\selectfont
\centering
\begin{tabular}{r c c c c c c c}
\toprule
Dataset & Soft-DTW $\gamma=1$ & $\gamma=0.1$ & $\gamma=0.01$ & $\gamma=0.001$ & Subgradient
method & DBA & Euclidean mean \\
\cmidrule[1pt](rl){2-8}
50words & 5.000 & 2.785 & {\bf 2.513} & 2.721 & 44.399 & 4.554 & 25.388 \\
Adiac & 0.235 & 0.207 & 0.257 & 0.428 & 25.533 & 0.754 & {\bf 0.177} \\
ArrowHead & 2.390 & 1.598 & {\bf 1.487} & 1.664 & 36.125 & 2.512 & 2.743 \\
Beef & 10.471 & 6.541 & {\bf 6.200} & 6.238 & 88.100 & 7.780 & 25.347 \\
BeetleFly & 35.790 & 23.655 & {\bf 22.559} & 23.105 & 77.993 & 25.122 & 191.574 \\
BirdChicken & 23.300 & 12.542 & {\bf 11.164} & 11.954 & 45.777 & 12.820 & 92.061 \\
CBF & 21.098 & {\bf 11.949} & 12.564 & 12.667 & 30.281 & 14.836 & 28.236 \\
Car & 2.639 & 1.750 & {\bf 1.611} & 1.914 & 80.437 & 2.609 & 5.106 \\
ChlorineConcentration & 22.260 & {\bf 13.932} & 14.818 & 15.044 & 32.134 & 16.168 & 15.411 \\
CinC\_ECG\_torso & 118.872 & 80.248 & {\bf 76.536} & 76.812 & 262.221 & 90.663 & 761.238 \\
Coffee & 1.036 & 0.871 & 1.262 & 1.630 & 41.741 & 2.380 & {\bf 0.591} \\
Computers & 231.421 & 182.380 & {\bf 178.184} & 179.886 & $\infty$ & 183.388 & 391.830 \\
Cricket\_X & 41.514 & 29.290 & {\bf 28.424} & 28.851 & 70.128 & 28.955 & 104.699 \\
Cricket\_Y & 51.858 & {\bf 30.321} & 30.337 & 31.041 & 70.989 & 33.098 & 107.712 \\
Cricket\_Z & 43.458 & 30.264 & {\bf 28.668} & 29.373 & 68.382 & 32.182 & 128.129 \\
DiatomSizeReduction & 0.055 & 0.054 & 0.064 & 0.132 & 49.308 & 0.418 & {\bf 0.033} \\
DistalPhalanxOutlineAgeGroup & 1.380 & 1.074 & 1.407 & 1.509 & 11.539 & 1.761 & {\bf 0.981} \\
DistalPhalanxOutlineCorrect & 2.501 & {\bf 1.968} & 2.267 & 2.634 & 13.169 & 2.991 & 2.374 \\
DistalPhalanxTW & 1.148 & 0.906 & 1.015 & 1.159 & 10.957 & 1.228 & {\bf 0.867} \\
ECG200 & 7.374 & {\bf 6.400} & 6.871 & 7.047 & 19.514 & 8.257 & 10.107 \\
ECG5000 & {\bf 8.951} & 8.961 & 10.601 & 10.265 & 34.558 & 12.098 & 14.517 \\
ECGFiveDays & 9.816 & {\bf 9.019} & 9.364 & 9.407 & 17.898 & 9.837 & 23.975 \\
Earthquakes & 148.959 & {\bf 85.219} & 85.470 & 85.515 & $\infty$ & 85.788 & 330.776 \\
ElectricDevices & 27.852 & {\bf 23.769} & 23.783 & 24.009 & $\infty$ & 24.869 & 56.470 \\
FISH & 0.978 & {\bf 0.641} & 0.662 & 0.957 & 63.566 & 1.680 & 1.543 \\
FaceAll & 15.068 & {\bf 12.373} & 13.248 & 13.494 & 24.582 & 15.980 & 29.982 \\
FaceFour & 15.500 & {\bf 14.519} & 15.002 & 14.849 & $\infty$ & 16.339 & 25.410 \\
FacesUCR & 15.033 & {\bf 13.077} & 13.405 & 14.174 & 30.621 & 14.796 & 27.026 \\
FordA & 56.936 & {\bf 45.492} & 46.170 & 46.038 & 96.087 & 49.723 & 218.482 \\
FordB & 59.117 & 47.812 & {\bf 47.058} & 47.642 & 102.279 & 50.262 & 250.595 \\
Gun\_Point & 7.204 & 2.507 & {\bf 2.037} & 2.211 & 22.590 & 2.374 & 7.286 \\
Ham & 24.833 & {\bf 19.101} & 20.397 & 20.713 & 55.769 & 22.807 & 30.685 \\
HandOutlines & 3.400 & {\bf 2.690} & 2.814 & 28.759 & 353.235 & 3.422 & 7.838 \\
Haptics & 16.424 & 14.351 & {\bf 14.129} & 14.320 & 172.988 & 16.464 & 39.559 \\
Herring & 1.212 & {\bf 0.946} & 1.022 & 1.349 & 71.388 & 2.097 & 1.884 \\
InlineSkate & 83.107 & 29.672 & {\bf 22.819} & N/A & N/A & N/A & N/A \\
ItalyPowerDemand & 2.442 & {\bf 2.124} & 2.316 & 2.372 & 5.434 & 2.355 & 2.329 \\
MedicalImages & 6.934 & {\bf 5.809} & 5.980 & 6.089 & 22.777 & 6.252 & 10.911 \\
MiddlePhalanxOutlineAgeGroup & 0.858 & 0.753 & 1.305 & 1.375 & 11.474 & 1.605 & {\bf 0.624} \\
MiddlePhalanxOutlineCorrect & 0.832 & 0.714 & 0.985 & 1.030 & 11.643 & 1.678 & {\bf 0.611} \\
MiddlePhalanxTW & 0.755 & 0.581 & 0.963 & 1.206 & 10.684 & 1.274 & {\bf 0.447} \\
MoteStrain & 24.177 & 21.639 & 21.616 & {\bf 21.554} & 32.007 & 22.437 & 26.646 \\
NonInvasiveFatalECG\_Thorax1 & 6.671 & {\bf 3.324} & 4.738 & 5.031 & 59.162 & 6.378 & 3.568 \\
NonInvasiveFatalECG\_Thorax2 & {\bf 2.559} & 3.159 & 3.587 & 4.097 & 40339.200 & 6.494 & 2.864 \\
OSULeaf & 30.041 & 20.692 & 20.034 & {\bf 19.950} & 76.057 & 23.915 & 136.512 \\
OliveOil & 0.657 & 0.959 & 1.494 & 1.804 & 95.499 & 3.420 & {\bf 0.008} \\
PhalangesOutlinesCorrect & 1.383 & {\bf 1.114} & 1.405 & 1.516 & 11.070 & 1.743 & 1.210 \\
Phoneme & 99.205 & {\bf 72.412} & 73.666 & 73.767 & 157.124 & 78.664 & 138.157 \\
Plane & 1.079 & {\bf 0.849} & 1.220 & 1.629 & 20.328 & 2.111 & 1.209 \\
ProximalPhalanxOutlineAgeGroup & 0.618 & 0.511 & 0.691 & 0.878 & 10.437 & 1.177 & {\bf 0.322} \\
ProximalPhalanxOutlineCorrect & 0.749 & 0.654 & 0.833 & 0.882 & 10.767 & 1.111 & {\bf 0.615} \\
ProximalPhalanxTW & 0.653 & 0.536 & 0.672 & 0.778 & 10.377 & 1.133 & {\bf 0.462} \\
RefrigerationDevices & 159.745 & 146.601 & {\bf 140.634} & 141.200 & $\infty$ & 148.931 & 363.732 \\
ScreenType & 156.442 & 123.746 & {\bf 121.432} & 122.247 & $\infty$ & 132.748 & 334.379 \\
ShapeletSim & 236.039 & {\bf 123.605} & 125.657 & 126.062 & 154.480 & 130.428 & 283.458 \\
ShapesAll & 15.267 & 7.108 & {\bf 6.466} & 6.818 & 58.734 & 8.236 & 67.790 \\
SmallKitchenAppliances & 176.073 & 164.419 & {\bf 162.009} & 162.585 & $\infty$ & 168.170 & 524.542 \\
SonyAIBORobotSurface & 5.735 & {\bf 4.916} & 5.425 & 5.337 & $\infty$ & 5.813 & 5.430 \\
SonyAIBORobotSurfaceII & 11.994 & {\bf 11.048} & 11.278 & 11.405 & 6637415.793 & 11.481 & 13.783 \\
StarLightCurves & 22.342 & 11.757 & 7.934 & {\bf 7.654} & 102.255 & 11.600 & 47.353 \\
Strawberry & 1.392 & {\bf 1.191} & 1.477 & 1.627 & 28.696 & 2.347 & 1.300 \\
SwedishLeaf & 2.409 & {\bf 1.968} & 2.476 & 2.904 & 19.638 & 3.283 & 6.274 \\
Symbols & 0.845 & 0.488 & {\bf 0.439} & 0.460 & $\infty$ & 0.828 & 4.953 \\
ToeSegmentation1 & 35.904 & 27.461 & {\bf 26.554} & 26.988 & 63.040 & 29.838 & 129.858 \\
ToeSegmentation2 & 34.177 & 24.476 & {\bf 23.003} & 23.194 & $\infty$ & 24.944 & 170.222 \\
Trace & 2.686 & 1.453 & {\bf 0.870} & 1.031 & 43.017 & 2.233 & 26.037 \\
TwoLeadECG & 1.811 & {\bf 1.514} & 1.641 & 1.701 & 7.961 & 1.802 & 2.216 \\
Two\_Patterns & 12.048 & 9.294 & {\bf 7.764} & 8.143 & 22.489 & 8.937 & 60.963 \\
UWaveGestureLibraryAll & 68.276 & 42.692 & {\bf 38.327} & 40.320 & $\infty$ & 49.486 & 181.901 \\
Wine & 0.728 & 0.500 & 0.746 & 1.147 & 32.463 & 1.812 & {\bf 0.094} \\
WordsSynonyms & 9.305 & 4.917 & {\bf 4.491} & 4.740 & 48.605 & 7.209 & 29.713 \\
Worms & 100.683 & 64.029 & 61.527 & 61.296 & {\bf 35.906} & 68.282 & 421.381 \\
WormsTwoClass & 110.292 & 68.932 & 66.258 & 65.964 & {\bf 37.047} & 72.387 & 430.774 \\
synthetic\_control & 14.366 & {\bf 7.115} & 7.506 & 7.516 & 15.931 & 8.123 & 12.187 \\
uWaveGestureLibrary\_X & 27.610 & 16.618 & 14.902 & {\bf 14.442} & $\infty$ & 18.269 & 75.119 \\
uWaveGestureLibrary\_Y & 29.964 & 16.106 & 14.556 & {\bf 14.450} & $\infty$ & 15.961 & 74.405 \\
uWaveGestureLibrary\_Z & 40.154 & 24.001 & {\bf 22.462} & 22.656 & $\infty$ & 25.040 & 107.540 \\
wafer & 25.831 & {\bf 23.595} & 25.828 & 25.195 & $\infty$ & 27.323 & 65.100 \\
yoga & 27.418 & 13.524 & {\bf 11.828} & 12.051 & 40.171 & 15.319 & 111.236 \\
\bottomrule
\end{tabular}
\end{table}

\section{Barycenters: DTW loss (Eq. (\protect\ref{eq:barycenter_obj}) with $\gamma=0$)
achieved with Euclidean init}
\label{appendix:barycenter_euc}

\begin{table}[H]
%\scriptsize
\fontsize{8}{7}\selectfont
\centering
\begin{tabular}{r c c c c c c c}
\toprule
Dataset & Soft-DTW $\gamma=1$ & $\gamma=0.1$ & $\gamma=0.01$ & $\gamma=0.001$ & Subgradient
method & DBA & Euclidean mean \\
\cmidrule[1pt](rl){2-8}
50words & 5.400 & 2.895 & {\bf 2.355} & 2.439 & 4.064 & 2.595 & 22.294 \\
Adiac & 0.124 & 0.103 & 0.089 & {\bf 0.069} & 0.081 & 0.071 & 0.103 \\
ArrowHead & 2.677 & 1.759 & {\bf 1.282} & 1.327 & 1.587 & 1.411 & 2.965 \\
Beef & 14.814 & 6.412 & {\bf 5.252} & 5.694 & 11.112 & 5.528 & 31.486 \\
BeetleFly & 33.082 & 20.819 & {\bf 20.781} & 22.127 & 25.554 & 21.960 & 191.285 \\
BirdChicken & 21.646 & 9.445 & {\bf 7.807} & 8.026 & 473.653 & 8.243 & 70.614 \\
CBF & 22.498 & 11.844 & {\bf 11.433} & 11.597 & 15.321 & 12.291 & 28.228 \\
Car & 1.556 & 0.932 & {\bf 0.693} & 0.901 & 1.171 & 1.079 & 2.439 \\
ChlorineConcentration & 19.239 & 10.663 & {\bf 10.434} & 10.468 & 11.370 & 10.638 & 13.549 \\
CinC\_ECG\_torso & 112.562 & 78.292 & 69.415 & 70.383 & 76.693 & {\bf 68.641} & 751.445 \\
Coffee & 1.078 & 0.657 & 0.460 & {\bf 0.393} & 0.435 & 0.399 & 0.571 \\
Computers & 172.590 & {\bf 138.605} & 144.576 & 146.409 & $\infty$ & 154.956 & 381.271 \\
Cricket\_X & 48.334 & 35.136 & {\bf 33.103} & 33.312 & 42.018 & 34.430 & 125.879 \\
Cricket\_Y & 41.804 & 31.395 & {\bf 31.044} & 31.158 & 35.957 & 31.749 & 97.393 \\
Cricket\_Z & 46.957 & {\bf 33.453} & 34.005 & 33.708 & 45.125 & 36.025 & 140.474 \\
DiatomSizeReduction & 0.039 & 0.033 & 0.028 & 0.021 & 0.024 & {\bf 0.019} & 0.032 \\
DistalPhalanxOutlineAgeGroup & 1.578 & 0.988 & 0.784 & {\bf 0.779} & 0.847 & 0.794 & 1.075 \\
DistalPhalanxOutlineCorrect & 2.878 & 2.002 & {\bf 1.751} & 1.754 & 2475.922 & 1.790 & 2.780 \\
DistalPhalanxTW & 1.377 & 0.837 & 0.655 & {\bf 0.651} & 0.773 & 0.667 & 0.997 \\
ECG200 & 7.266 & 5.608 & {\bf 5.395} & 5.424 & 5.955 & 5.494 & 9.638 \\
ECG5000 & 12.430 & 10.377 & {\bf 10.332} & 10.343 & 12.340 & 10.595 & 18.886 \\
ECGFiveDays & 8.416 & 7.452 & {\bf 7.046} & 7.101 & 145.106 & 7.145 & 23.477 \\
Earthquakes & 172.035 & 91.568 & {\bf 90.684} & 91.071 & $\infty$ & 92.126 & 335.240 \\
ElectricDevices & 30.832 & {\bf 26.480} & 27.131 & 27.076 & $\infty$ & 27.615 & 57.938 \\
FISH & 1.183 & 0.806 & 0.541 & {\bf 0.508} & 0.645 & 0.551 & 1.933 \\
FaceAll & 18.102 & 13.305 & 13.104 & {\bf 13.074} & 16.491 & 13.915 & 40.404 \\
FaceFour & 17.070 & 13.069 & {\bf 12.984} & 13.091 & $\infty$ & 13.568 & 28.203 \\
FacesUCR & 17.172 & {\bf 13.081} & 13.293 & 13.394 & 15.780 & 13.498 & 35.942 \\
FordA & 53.903 & 42.199 & {\bf 41.835} & 41.966 & 53.545 & 44.259 & 235.362 \\
FordB & 61.168 & 48.150 & {\bf 47.327} & 47.743 & 60.120 & 50.121 & 246.802 \\
Gun\_Point & 5.924 & 2.132 & 1.695 & {\bf 1.666} & 2.543 & 1.682 & 5.906 \\
Ham & 25.353 & 18.841 & 17.457 & {\bf 17.294} & $\infty$ & 17.917 & 32.456 \\
HandOutlines & 2.238 & 1.718 & 1.004 & 0.527 & $\infty$ & {\bf 0.515} & 6.452 \\
Haptics & 12.554 & 8.874 & {\bf 7.785} & 8.197 & 12.193 & 8.219 & 35.613 \\
Herring & 1.655 & 1.117 & 0.809 & {\bf 0.760} & 0.956 & 0.817 & 2.564 \\
InlineSkate & 100.849 & 46.460 & {\bf 27.248} & 35.578 & N/A & N/A & N/A \\
ItalyPowerDemand & 2.597 & 1.990 & {\bf 1.956} & 1.985 & 2.132 & 1.997 & 2.449 \\
MedicalImages & 5.719 & 4.319 & 4.145 & {\bf 4.070} & 4.791 & 4.371 & 8.047 \\
MiddlePhalanxOutlineAgeGroup & 0.870 & 0.578 & 0.427 & {\bf 0.415} & 0.464 & 0.426 & 0.552 \\
MiddlePhalanxOutlineCorrect & 0.799 & 0.609 & 0.460 & {\bf 0.443} & 0.501 & 0.461 & 0.577 \\
MiddlePhalanxTW & 0.658 & 0.466 & 0.335 & {\bf 0.321} & 0.358 & 0.332 & 0.434 \\
MoteStrain & 24.451 & {\bf 20.720} & 20.829 & 21.057 & $\infty$ & 21.273 & 26.694 \\
NonInvasiveFatalECG\_Thorax1 & 1.619 & 1.384 & 0.907 & 0.785 & {\bf 0.691} & 0.814 & 1.400 \\
NonInvasiveFatalECG\_Thorax2 & 1.624 & 1.370 & 0.932 & {\bf 0.827} & 2.163 & 0.853 & 1.409 \\
OSULeaf & 27.428 & 18.666 & {\bf 18.544} & 18.595 & 24.692 & 20.244 & 135.980 \\
OliveOil & 0.367 & 0.074 & 0.022 & 0.013 & 0.011 & {\bf 0.009} & 0.011 \\
PhalangesOutlinesCorrect & 1.172 & 0.895 & 0.699 & {\bf 0.695} & 0.766 & 0.704 & 1.002 \\
Phoneme & 135.535 & {\bf 104.971} & 105.478 & 108.031 & 126.055 & 108.513 & 254.392 \\
Plane & 0.928 & 0.600 & 0.404 & {\bf 0.399} & 0.499 & 0.430 & 1.203 \\
ProximalPhalanxOutlineAgeGroup & 0.820 & 0.502 & 0.361 & {\bf 0.346} & 0.390 & 0.356 & 0.512 \\
ProximalPhalanxOutlineCorrect & 0.816 & 0.630 & 0.463 & {\bf 0.452} & 0.517 & 0.461 & 0.669 \\
ProximalPhalanxTW & 0.637 & 0.431 & 0.313 & {\bf 0.304} & 0.341 & 0.308 & 0.471 \\
RefrigerationDevices & 154.420 & {\bf 133.321} & 135.721 & 135.300 & $\infty$ & 142.697 & 358.823 \\
ScreenType & 189.188 & 143.582 & 143.894 & {\bf 141.776} & $\infty$ & 148.464 & 325.840 \\
ShapeletSim & 231.937 & 124.443 & {\bf 122.000} & 122.506 & 154.089 & 127.977 & 284.079 \\
ShapesAll & 13.416 & 7.519 & {\bf 6.420} & 6.509 & 7.317 & 7.478 & 80.306 \\
SmallKitchenAppliances & 188.030 & 173.670 & 169.755 & {\bf 167.097} & $\infty$ & 173.004 & 505.356 \\
SonyAIBORobotSurface & 5.715 & 4.002 & 3.870 & 3.896 & $\infty$ & {\bf 3.828} & 5.444 \\
SonyAIBORobotSurfaceII & 11.300 & 8.947 & {\bf 8.853} & 8.871 & 12.651 & 8.977 & 14.225 \\
StarLightCurves & 13.581 & 6.619 & 4.054 & {\bf 3.765} & 7.247 & 4.517 & 30.354 \\
Strawberry & 2.218 & 1.413 & 1.128 & {\bf 1.070} & 1.374 & 1.156 & 2.128 \\
SwedishLeaf & 2.957 & 2.068 & {\bf 2.049} & 2.081 & 2.520 & 2.163 & 6.236 \\
Symbols & 0.762 & 0.451 & 0.412 & {\bf 0.401} & $\infty$ & 0.474 & 4.822 \\
ToeSegmentation1 & 35.832 & 26.067 & 26.337 & {\bf 25.735} & 31.157 & 27.493 & 131.683 \\
ToeSegmentation2 & 34.264 & 22.238 & {\bf 20.800} & 21.563 & $\infty$ & 23.080 & 164.101 \\
Trace & 1.737 & 1.744 & 1.508 & {\bf 1.378} & 4.170 & 1.969 & 26.814 \\
TwoLeadECG & 1.533 & 1.172 & {\bf 1.030} & 1.043 & 1.323 & 1.093 & 2.046 \\
Two\_Patterns & 10.891 & 7.505 & {\bf 6.045} & 6.079 & 18.987 & 6.584 & 66.027 \\
UWaveGestureLibraryAll & 67.549 & 38.179 & {\bf 32.894} & 33.426 & $\infty$ & 39.241 & 167.486 \\
Wine & 0.707 & 0.188 & 0.127 & 0.111 & 0.114 & {\bf 0.110} & 0.118 \\
WordsSynonyms & 9.804 & 7.282 & {\bf 6.711} & 6.785 & 8.884 & 6.868 & 39.843 \\
Worms & 101.850 & 61.067 & 58.725 & {\bf 56.793} & 244.738 & 63.234 & 415.674 \\
WormsTwoClass & 122.901 & 68.771 & {\bf 64.655} & 64.898 & 1297.616 & 72.011 & 395.088 \\
synthetic\_control & 18.147 & {\bf 9.189} & 9.307 & 9.350 & 11.520 & 9.614 & 19.237 \\
uWaveGestureLibrary\_X & 34.423 & 19.787 & 18.746 & {\bf 17.807} & $\infty$ & 24.269 & 93.839 \\
uWaveGestureLibrary\_Y & 27.744 & 14.309 & {\bf 13.010} & 13.607 & $\infty$ & 15.283 & 51.854 \\
uWaveGestureLibrary\_Z & 21.927 & 10.081 & 8.456 & {\bf 8.453} & $\infty$ & 11.040 & 47.947 \\
wafer & 32.561 & 29.197 & 28.908 & {\bf 28.820} & $\infty$ & 33.379 & 67.413 \\
yoga & 23.698 & 11.632 & 9.433 & {\bf 9.204} & 16.239 & 10.058 & 93.688 \\
\bottomrule
\end{tabular}
\end{table}

\section{$k$-means clustering: DTW loss achieved (Eq.
    (\protect\ref{eq:kmeans_obj}) with
    $\gamma=0$, log-scaled) when using random initialization}
\label{appendix:k_means_random}

\begin{table}[H]
%\scriptsize
\fontsize{8}{7}\selectfont
\centering
\begin{tabular}{r c c c c c c c}
\toprule
Dataset & Soft-DTW $\gamma=1$ & $\gamma=0.1$ & $\gamma=0.01$ & $\gamma=0.001$ & Subgradient
method & DBA & Euclidean mean \\
\cmidrule[1pt](rl){2-8}
50words & 16.294 & 16.193 & {\bf 16.125} & 16.135 & 16.163 & 16.156 & 16.205 \\
Adiac & {\bf 11.933} & {\bf 11.933} & {\bf 11.933} & {\bf 11.933} & {\bf 11.933} & {\bf 11.933} & {\bf 11.933} \\
ArrowHead & 9.020 & 8.757 & 8.699 & {\bf 8.687} & 8.732 & 8.692 & 8.958 \\
Beef & 11.215 & 11.095 & 11.069 & {\bf 11.061} & {\bf 11.061} & 11.117 & 11.215 \\
BeetleFly & 9.946 & 9.618 & {\bf 9.531} & 9.592 & 9.619 & 9.591 & 10.368 \\
BirdChicken & 9.996 & 9.652 & {\bf 9.374} & 9.515 & 9.870 & 9.585 & 10.335 \\
CBF & 10.150 & 10.065 & {\bf 10.005} & 10.006 & 10.009 & 10.009 & 10.150 \\
Car & 9.392 & 9.290 & 9.067 & {\bf 9.039} & 9.059 & 9.046 & 9.276 \\
ChlorineConcentration & 15.512 & 15.214 & 15.182 & {\bf 15.175} & 15.176 & 15.176 & 15.331 \\
CinC\_ECG\_torso & 13.134 & {\bf 12.837} & 12.848 & 12.868 & 13.621 & 12.877 & 13.621 \\
Coffee & 7.150 & 6.893 & 6.692 & {\bf 6.628} & 6.693 & 6.651 & 6.825 \\
Computers & {\bf 16.420} & 16.498 & 16.489 & 16.502 & 16.960 & 16.475 & 16.960 \\
Cricket\_X & 16.922 & 16.696 & 16.629 & {\bf 16.628} & 16.649 & 16.653 & 16.955 \\
Cricket\_Y & 16.783 & 16.588 & 16.545 & {\bf 16.533} & 16.570 & 16.570 & 16.803 \\
Cricket\_Z & 16.874 & 16.669 & 16.597 & {\bf 16.593} & 16.620 & 16.620 & 16.981 \\
DiatomSizeReduction & 5.959 & 5.907 & 5.889 & {\bf 5.739} & 5.798 & 5.758 & 5.932 \\
DistalPhalanxOutlineAgeGroup & 11.193 & 11.220 & 11.202 & 11.198 & 11.194 & 11.196 & {\bf 11.158} \\
DistalPhalanxOutlineCorrect & 12.467 & 12.373 & {\bf 12.340} & 12.342 & 12.494 & 12.350 & 12.483 \\
DistalPhalanxTW & 11.244 & 11.260 & 11.263 & 11.251 & 11.264 & 11.261 & {\bf 11.222} \\
ECG200 & 11.395 & 11.317 & 11.323 & {\bf 11.274} & 11.300 & 11.289 & 11.501 \\
ECG5000 & 16.169 & {\bf 16.084} & 16.142 & 16.136 & 16.137 & 16.136 & 16.211 \\
ECGFiveDays & 8.734 & 8.579 & 8.522 & {\bf 8.513} & 8.713 & 8.533 & 8.818 \\
Earthquakes & 14.757 & 14.727 & {\bf 14.726} & 14.728 & 14.757 & {\bf 14.726} & 14.757 \\
ElectricDevices & 22.404 & 22.428 & 22.401 & 22.398 & 22.630 & 22.399 & {\bf 22.332} \\
FISH & 10.841 & 10.740 & 10.594 & {\bf 10.514} & 10.560 & 10.566 & 10.841 \\
FaceAll & 16.272 & 16.187 & 16.185 & 16.183 & 16.197 & {\bf 16.182} & 16.291 \\
FaceFour & 10.422 & 10.318 & {\bf 10.302} & 10.316 & 10.575 & 10.321 & 10.533 \\
FacesUCR & 14.479 & 14.432 & 14.426 & {\bf 14.423} & 14.430 & 14.429 & 14.431 \\
FordA & 18.604 & 18.390 & 18.388 & 18.387 & 18.977 & {\bf 18.385} & 18.977 \\
FordB & 17.620 & 17.429 & 17.425 & 17.426 & 17.466 & {\bf 17.416} & 17.998 \\
Gun\_Point & 10.242 & 10.019 & 9.843 & 9.743 & 9.883 & {\bf 9.738} & 10.130 \\
Ham & 12.772 & 12.545 & 12.488 & {\bf 12.473} & 13.240 & 12.506 & 12.957 \\
MedicalImages & 15.081 & 14.982 & 14.985 & {\bf 14.979} & 14.986 & 14.986 & 15.032 \\
MiddlePhalanxOutlineAgeGroup & 9.909 & 9.919 & 9.856 & {\bf 9.818} & 9.824 & 9.822 & 9.856 \\
MiddlePhalanxOutlineCorrect & 11.121 & 11.088 & 10.984 & 10.951 & 10.962 & 10.961 & {\bf 10.923} \\
MiddlePhalanxTW & {\bf 10.514} & {\bf 10.514} & {\bf 10.514} & {\bf 10.514} & {\bf 10.514} & {\bf 10.514} & {\bf 10.514} \\
MoteStrain & 9.560 & 9.484 & 9.460 & 9.451 & {\bf 9.201} & 9.470 & 9.557 \\
NonInvasiveFatalECG\_Thorax1 & {\bf 17.728} & N/A & N/A & N/A & N/A & N/A & N/A \\
ProximalPhalanxTW & 11.055 & 10.993 & 10.978 & {\bf 10.958} & 10.968 & 10.965 & 10.968 \\
RefrigerationDevices & 17.391 & 17.351 & {\bf 17.311} & 17.322 & 17.758 & 17.324 & 17.758 \\
ScreenType & 17.467 & 17.388 & 17.306 & 17.297 & 18.126 & {\bf 17.289} & 17.838 \\
ShapeletSim & 11.176 & {\bf 10.896} & 10.905 & 10.906 & 10.916 & 10.915 & 11.176 \\
ShapesAll & 17.539 & 17.405 & {\bf 17.331} & 17.333 & 17.605 & 17.357 & 17.509 \\
SmallKitchenAppliances & 17.551 & 17.611 & {\bf 17.537} & 17.606 & 18.007 & 17.561 & 18.007 \\
SonyAIBORobotSurface & 8.181 & 7.959 & {\bf 7.934} & 7.943 & 8.247 & 7.958 & 8.084 \\
SonyAIBORobotSurfaceII & 9.349 & {\bf 9.265} & 9.267 & 9.267 & 9.325 & 9.277 & 9.338 \\
StarLightCurves & 19.435 & 19.110 & {\bf 19.012} & N/A & N/A & N/A & N/A \\
Trace & 14.570 & 14.570 & 14.556 & {\bf 14.550} & 14.555 & 14.556 & 14.553 \\
TwoLeadECG & 6.939 & 6.939 & 6.892 & 6.879 & 6.936 & 6.892 & {\bf 6.743} \\
Two\_Patterns & 17.416 & 17.379 & 17.317 & 17.325 & 17.524 & {\bf 17.307} & 17.524 \\
UWaveGestureLibraryAll & 18.911 & 18.641 & {\bf 18.514} & 18.531 & 19.282 & 18.537 & 19.244 \\
Wine & 7.527 & 7.297 & 6.482 & 6.358 & 6.390 & 6.353 & {\bf 6.223} \\
WordsSynonyms & 15.209 & 15.093 & {\bf 15.024} & 15.025 & 15.053 & 15.036 & 15.159 \\
Worms & 14.184 & 14.051 & {\bf 13.889} & 13.896 & 14.943 & 13.968 & 14.648 \\
WormsTwoClass & 13.727 & 13.471 & {\bf 13.429} & 13.462 & 14.944 & 13.494 & 14.944 \\
synthetic\_control & 15.338 & 15.303 & 15.292 & 15.291 & 15.303 & 15.295 & {\bf 15.278} \\
uWaveGestureLibrary\_X & 18.789 & {\bf 18.568} & N/A & N/A & N/A & N/A & N/A \\
\bottomrule
\end{tabular}
\end{table}

\section{$k$-means clustering: DTW loss achieved (Eq.
    (\protect\ref{eq:kmeans_obj}) with
    $\gamma=0$, log-scaled) when using Euclidean mean initialization}
\label{appendix:k_means_euc}

\begin{table}[H]
%\scriptsize
\fontsize{8}{7}\selectfont
\centering
\begin{tabular}{r c c c c c c c}
\toprule
Dataset & Soft-DTW $\gamma=1$ & $\gamma=0.1$ & $\gamma=0.01$ & $\gamma=0.001$ & Subgradient
method & DBA & Euclidean mean \\
\cmidrule[1pt](rl){2-8}
50words & 16.233 & 16.145 & 16.046 & {\bf 16.035} & 16.045 & 16.233 & 16.233 \\
Adiac & 12.311 & 12.311 & 12.264 & 12.241 & 12.234 & {\bf 12.233} & 12.311 \\
ArrowHead & 9.014 & 8.963 & 8.766 & {\bf 8.746} & 8.851 & 8.809 & 9.014 \\
Beef & 11.225 & 11.110 & 11.088 & 11.079 & 11.077 & {\bf 11.070} & 11.300 \\
BeetleFly & 9.895 & 9.290 & 9.268 & {\bf 9.240} & 9.512 & 10.926 & 10.926 \\
BirdChicken & 10.032 & 9.542 & 9.352 & {\bf 9.338} & 9.422 & 9.414 & 10.335 \\
CBF & 10.246 & 9.995 & 9.910 & {\bf 9.908} & 9.933 & 9.921 & 10.246 \\
Car & 9.276 & 9.229 & 8.989 & {\bf 8.910} & 8.936 & 8.935 & 9.276 \\
ChlorineConcentration & 15.331 & 15.291 & 15.254 & {\bf 15.252} & 15.270 & {\bf 15.252} & 15.331 \\
CinC\_ECG\_torso & 13.197 & 12.803 & 12.752 & {\bf 12.728} & 13.723 & 13.723 & 13.723 \\
Coffee & 6.825 & 6.825 & 6.668 & 6.599 & 6.605 & {\bf 6.591} & 6.825 \\
Computers & 16.417 & 16.346 & 16.301 & {\bf 16.289} & 17.167 & 16.342 & 17.167 \\
Cricket\_X & 16.895 & 16.719 & 16.622 & 16.623 & 16.612 & {\bf 16.600} & 16.987 \\
Cricket\_Y & 16.770 & 16.651 & 16.546 & {\bf 16.514} & 16.515 & 16.527 & 16.861 \\
Cricket\_Z & 16.924 & 16.748 & 16.670 & {\bf 16.633} & 16.653 & 16.653 & 17.028 \\
DiatomSizeReduction & 5.963 & 5.963 & 5.963 & {\bf 5.884} & 5.897 & 5.896 & 5.963 \\
DistalPhalanxOutlineAgeGroup & {\bf 11.164} & {\bf 11.164} & {\bf 11.164} & {\bf 11.164} & {\bf 11.164} & {\bf 11.164} & {\bf 11.164} \\
DistalPhalanxOutlineCorrect & 12.544 & 12.533 & 12.494 & 12.475 & {\bf 12.240} & 12.479 & 12.577 \\
DistalPhalanxTW & {\bf 11.242} & 11.259 & 11.256 & 11.243 & 11.251 & 11.245 & 11.259 \\
ECG200 & 11.462 & 11.291 & 11.239 & {\bf 11.222} & 11.231 & 11.234 & 11.503 \\
ECG5000 & 16.253 & 16.180 & 16.171 & 16.183 & {\bf 16.170} & 16.172 & 16.262 \\
ECGFiveDays & 8.738 & 8.614 & {\bf 8.543} & 8.549 & 8.709 & 8.559 & 8.818 \\
Earthquakes & 15.113 & 14.625 & 14.601 & 14.599 & 15.952 & {\bf 14.597} & 15.952 \\
ElectricDevices & 22.295 & 22.325 & 22.291 & 22.290 & 22.379 & {\bf 22.283} & 22.379 \\
FISH & 10.904 & 10.843 & 10.589 & 10.543 & {\bf 10.527} & 10.555 & 10.904 \\
FaceAll & 16.278 & 16.162 & 16.145 & 16.145 & 16.152 & {\bf 16.140} & 16.347 \\
FaceFour & 10.376 & 10.273 & 10.239 & {\bf 10.226} & 10.566 & 10.241 & 10.566 \\
FacesUCR & 14.472 & 14.434 & 14.406 & 14.407 & {\bf 14.391} & 14.481 & 14.481 \\
FordA & 18.581 & {\bf 18.354} & {\bf 18.354} & 18.360 & 20.038 & 20.038 & 20.038 \\
FordB & 17.649 & 17.443 & 17.429 & 17.436 & 17.466 & {\bf 17.427} & 19.143 \\
Gun\_Point & 10.334 & 10.027 & 9.806 & {\bf 9.751} & 9.902 & 9.833 & 10.334 \\
Ham & 12.805 & 12.603 & 12.559 & {\bf 12.558} & 12.974 & 12.561 & 12.974 \\
HandOutlines & {\bf 13.712} & N/A & N/A & N/A & N/A & N/A & N/A \\
MedicalImages & 15.082 & 14.963 & {\bf 14.940} & 14.942 & 14.950 & 14.941 & 15.091 \\
MiddlePhalanxOutlineAgeGroup & 9.856 & 9.856 & 9.855 & {\bf 9.821} & 9.823 & {\bf 9.821} & 9.856 \\
MiddlePhalanxOutlineCorrect & 10.962 & 10.962 & 10.962 & 10.959 & {\bf 10.950} & {\bf 10.950} & 10.962 \\
MiddlePhalanxTW & {\bf 10.558} & 10.587 & 10.587 & 10.569 & 10.572 & 10.570 & 10.587 \\
MoteStrain & 9.551 & 9.454 & {\bf 9.413} & 9.451 & 9.557 & 9.446 & 9.557 \\
NonInvasiveFatalECG\_Thorax1 & {\bf 17.765} & N/A & N/A & N/A & N/A & N/A & N/A \\
ProximalPhalanxTW & 10.978 & {\bf 10.973} & 10.978 & 10.978 & 10.978 & 10.978 & 10.978 \\
RefrigerationDevices & 17.260 & 17.202 & 17.093 & {\bf 17.073} & 18.140 & 17.095 & 18.140 \\
ScreenType & 17.430 & 17.359 & 17.294 & {\bf 17.292} & 17.838 & 17.323 & 17.838 \\
ShapeletSim & 11.497 & 10.864 & {\bf 10.845} & 10.853 & 10.865 & 11.608 & 11.608 \\
ShapesAll & 17.560 & 17.431 & 17.335 & {\bf 17.328} & 17.560 & 17.560 & 17.560 \\
SmallKitchenAppliances & 17.310 & {\bf 17.273} & 17.357 & 17.357 & 18.206 & 18.206 & 18.206 \\
SonyAIBORobotSurface & 8.084 & 7.980 & {\bf 7.941} & 7.947 & 8.084 & 7.948 & 8.084 \\
SonyAIBORobotSurfaceII & 9.338 & 9.207 & 9.196 & {\bf 9.195} & 9.338 & {\bf 9.195} & 9.338 \\
StarLightCurves & 19.457 & 19.178 & {\bf 19.083} & N/A & N/A & N/A & N/A \\
Trace & 14.553 & 14.553 & 14.553 & {\bf 14.549} & 14.553 & 14.553 & 14.553 \\
TwoLeadECG & 6.743 & 6.705 & 6.623 & {\bf 6.606} & 6.666 & 6.633 & 6.743 \\
Two\_Patterns & {\bf 17.084} & 17.363 & 17.242 & 17.255 & 17.518 & 17.316 & 17.942 \\
UWaveGestureLibraryAll & 18.820 & 18.613 & 18.539 & {\bf 18.488} & 19.259 & 18.508 & 19.259 \\
Wine & 6.223 & 6.223 & 6.223 & 6.223 & 6.223 & {\bf 6.205} & 6.223 \\
WordsSynonyms & 15.184 & 15.036 & {\bf 14.947} & 14.951 & 14.965 & 14.959 & 15.196 \\
Worms & 14.043 & 13.860 & 13.791 & 13.777 & 14.696 & {\bf 13.772} & 14.696 \\
WormsTwoClass & 13.699 & 13.440 & {\bf 13.322} & 13.337 & 15.076 & 13.390 & 15.076 \\
synthetic\_control & 15.472 & 15.367 & 15.337 & 15.338 & {\bf 15.330} & 15.336 & 15.472 \\
uWaveGestureLibrary\_X & 18.844 & {\bf 18.562} & N/A & N/A & N/A & N/A & N/A \\
\bottomrule
\end{tabular}
\end{table}

\section{Time-series prediction: DTW loss achieved when using random
    init}

\begin{table}[H]
%\scriptsize
\fontsize{8}{7}\selectfont
\centering
\begin{tabular}{r c c c c c}
\toprule
Dataset & Soft-DTW loss $\gamma=1$ & $\gamma=0.1$ & $\gamma=0.01$ & $\gamma=0.001$ &
Euclidean loss \\
\cmidrule[1pt](rl){2-6}
50words & 6.473 & {\bf 4.921} & 4.999 & 6.489 & 18.734 \\
Adiac & 0.094 & {\bf 0.074} & 0.078 & 0.109 & 0.103 \\
ArrowHead & 1.851 & {\bf 1.708} & 1.933 & 1.909 & 2.073 \\
Beef & 12.229 & {\bf 8.688} & 10.244 & 9.126 & 22.228 \\
BeetleFly & 35.037 & 25.439 & 27.588 & {\bf 23.494} & 50.610 \\
BirdChicken & 31.878 & 19.914 & 25.100 & {\bf 14.981} & 30.693 \\
CBF & 10.802 & {\bf 9.263} & 9.595 & 10.151 & 12.868 \\
Car & 1.724 & 2.307 & 2.202 & {\bf 1.318} & 1.588 \\
ChlorineConcentration & 7.876 & 2.108 & 2.331 & 1.735 & {\bf 0.769} \\
CinC\_ECG\_torso & 45.675 & 26.337 & {\bf 23.567} & 24.550 & 48.171 \\
Coffee & 0.914 & 0.727 & 1.662 & 1.883 & {\bf 0.660} \\
Computers & 92.584 & 84.723 & 78.953 & {\bf 75.435} & 235.208 \\
Cricket\_X & 9.394 & 8.042 & {\bf 7.123} & 7.226 & 12.080 \\
Cricket\_Y & 11.989 & 9.643 & {\bf 9.534} & 9.545 & 15.002 \\
Cricket\_Z & 9.161 & 6.889 & {\bf 6.585} & 7.200 & 11.003 \\
DiatomSizeReduction & 1.182 & 0.922 & {\bf 0.820} & 0.897 & 1.203 \\
DistalPhalanxOutlineAgeGroup & 0.426 & 0.291 & 0.541 & 0.496 & {\bf 0.231} \\
DistalPhalanxOutlineCorrect & 0.494 & 0.476 & 0.564 & 0.591 & {\bf 0.351} \\
DistalPhalanxTW & 0.441 & 0.330 & 0.305 & 1.214 & {\bf 0.231} \\
ECG200 & 1.874 & {\bf 1.716} & 1.884 & 1.734 & 1.905 \\
ECG5000 & 4.895 & 4.705 & 4.543 & {\bf 4.441} & 5.463 \\
ECGFiveDays & 1.834 & 1.944 & 1.699 & {\bf 1.642} & 2.220 \\
Earthquakes & 74.738 & 59.973 & 60.877 & {\bf 57.827} & 147.980 \\
ElectricDevices & 20.186 & {\bf 15.125} & 15.218 & 15.287 & 37.121 \\
FISH & 0.464 & 0.429 & {\bf 0.354} & 0.459 & 0.462 \\
FaceAll & 9.317 & 7.451 & 7.902 & {\bf 7.276} & 10.716 \\
FaceFour & {\bf 19.564} & 20.881 & 28.150 & 28.839 & 46.841 \\
FacesUCR & 15.359 & {\bf 14.643} & 16.143 & 17.428 & 28.576 \\
Gun\_Point & 0.896 & {\bf 0.805} & 0.923 & 0.834 & 0.858 \\
Ham & 20.154 & 17.931 & 17.786 & {\bf 17.413} & 24.340 \\
Haptics & {\bf 16.174} & 17.775 & 17.142 & 17.423 & 23.130 \\
Herring & 1.000 & 0.712 & {\bf 0.666} & 0.762 & 0.865 \\
InsectWingbeatSound & 3.460 & 2.823 & 2.458 & {\bf 2.220} & 5.437 \\
ItalyPowerDemand & 0.911 & 0.893 & {\bf 0.711} & 0.798 & 0.881 \\
LargeKitchenAppliances & 63.153 & 60.739 & {\bf 60.157} & 61.841 & 266.853 \\
Lighting2 & 73.293 & 66.341 & {\bf 65.335} & 66.881 & 147.668 \\
Lighting7 & 44.446 & 42.699 & {\bf 40.608} & 41.502 & 68.902 \\
Meat & 0.162 & 0.242 & 0.246 & 0.650 & {\bf 0.099} \\
MedicalImages & 1.023 & 0.853 & {\bf 0.708} & 0.778 & 1.211 \\
MiddlePhalanxOutlineAgeGroup & 0.343 & 0.347 & 0.570 & 0.400 & {\bf 0.312} \\
MiddlePhalanxOutlineCorrect & 0.278 & 0.204 & 0.227 & 0.202 & {\bf 0.182} \\
MiddlePhalanxTW & 0.251 & 0.153 & 0.445 & 0.314 & {\bf 0.132} \\
MoteStrain & 10.188 & {\bf 9.986} & 11.119 & 10.250 & 11.183 \\
NonInvasiveFatalECG\_Thorax1 & 1.002 & 0.920 & 0.675 & {\bf 0.634} & 1.219 \\
OSULeaf & 15.125 & 11.722 & 11.086 & {\bf 10.775} & 30.739 \\
OliveOil & 0.476 & 0.683 & 2.082 & 2.076 & {\bf 0.020} \\
PhalangesOutlinesCorrect & 0.352 & 0.216 & 0.352 & 0.338 & {\bf 0.170} \\
Phoneme & 160.536 & 150.017 & 148.175 & {\bf 145.093} & 219.704 \\
Plane & 0.619 & {\bf 0.564} & 0.834 & 0.788 & 0.630 \\
ProximalPhalanxOutlineAgeGroup & 0.134 & 0.062 & 0.105 & 0.118 & {\bf 0.046} \\
ProximalPhalanxOutlineCorrect & 0.129 & 0.047 & 0.089 & 0.128 & {\bf 0.044} \\
ProximalPhalanxTW & 0.154 & 0.077 & 0.102 & 0.150 & {\bf 0.055} \\
RefrigerationDevices & 108.421 & 93.519 & {\bf 89.370} & 89.873 & 160.361 \\
ShapeletSim & 102.413 & {\bf 70.455} & 71.156 & 72.094 & 108.936 \\
ShapesAll & 10.391 & 9.027 & 7.850 & {\bf 7.207} & 18.348 \\
SonyAIBORobotSurface & 4.453 & 4.494 & {\bf 4.318} & 4.910 & 4.388 \\
SonyAIBORobotSurfaceII & 8.072 & 8.302 & {\bf 7.758} & 8.669 & 8.628 \\
Strawberry & 0.123 & 0.088 & 0.137 & 0.100 & {\bf 0.081} \\
SwedishLeaf & 1.486 & 1.277 & 1.316 & {\bf 1.169} & 1.633 \\
Symbols & 17.963 & 14.039 & 15.172 & {\bf 13.192} & 38.268 \\
ToeSegmentation1 & 23.866 & 22.987 & 26.056 & {\bf 22.401} & 35.806 \\
ToeSegmentation2 & 41.450 & 33.100 & {\bf 30.931} & 31.106 & 61.899 \\
Trace & 0.563 & 0.379 & 0.352 & {\bf 0.279} & 0.582 \\
TwoLeadECG & 0.441 & 0.394 & {\bf 0.318} & 0.320 & 0.336 \\
Two\_Patterns & 15.035 & 10.100 & 10.588 & {\bf 8.584} & 35.923 \\
UWaveGestureLibraryAll & 40.324 & 28.975 & 26.193 & {\bf 25.897} & 93.019 \\
Wine & 0.164 & 0.203 & 1.417 & 0.958 & {\bf 0.028} \\
WordsSynonyms & 12.466 & 10.437 & {\bf 9.165} & 9.219 & 32.003 \\
Worms & 81.236 & 63.938 & 60.950 & {\bf 59.995} & 114.528 \\
WormsTwoClass & 78.455 & 66.609 & {\bf 60.207} & 61.685 & 122.619 \\
synthetic\_control & 7.709 & {\bf 5.315} & 5.390 & 5.506 & 7.690 \\
uWaveGestureLibrary\_X & 13.096 & 9.995 & 10.143 & {\bf 9.433} & 19.995 \\
uWaveGestureLibrary\_Y & 9.793 & 7.272 & 7.327 & {\bf 7.225} & 17.706 \\
uWaveGestureLibrary\_Z & 11.883 & 8.909 & 8.494 & {\bf 8.416} & 20.092 \\
wafer & 1.049 & 0.473 & {\bf 0.386} & 0.496 & 2.636 \\
yoga & 2.932 & 2.431 & {\bf 1.995} & 3.309 & 4.305 \\
\bottomrule
\end{tabular}
\end{table}

\clearpage
\section{Time-series prediction: DTW loss achieved when using Euclidean
    init}

\begin{table}[H]
%\scriptsize
\fontsize{8}{7}\selectfont
\centering
\begin{tabular}{r c c c c c}
\toprule
Dataset & Soft-DTW loss $\gamma=1$ & $\gamma=0.1$ & $\gamma=0.01$ & $\gamma=0.001$ &
Euclidean loss \\
\cmidrule[1pt](rl){2-6}
50words & 6.330 & 5.628 & 4.885 & {\bf 4.553} & 18.734 \\
Adiac & 0.082 & 0.076 & {\bf 0.064} & 0.079 & 0.103 \\
ArrowHead & 1.823 & 2.016 & {\bf 1.762} & 2.106 & 2.073 \\
Beef & 7.250 & 6.940 & 7.146 & {\bf 3.757} & 22.228 \\
BeetleFly & 32.430 & {\bf 26.600} & 27.199 & 29.003 & 50.610 \\
BirdChicken & 24.952 & 22.600 & {\bf 19.914} & 20.540 & 30.693 \\
CBF & 10.744 & 8.978 & 9.215 & {\bf 8.398} & 12.868 \\
Car & 0.906 & 0.812 & {\bf 0.709} & 0.740 & 1.588 \\
ChlorineConcentration & 6.018 & 0.979 & {\bf 0.695} & 0.698 & 0.769 \\
CinC\_ECG\_torso & 29.892 & {\bf 18.638} & 19.635 & 19.191 & 48.171 \\
Coffee & 0.870 & 0.582 & 0.511 & {\bf 0.496} & 0.660 \\
Computers & 86.619 & {\bf 79.250} & 82.215 & 81.417 & 235.208 \\
Cricket\_X & 10.954 & 8.200 & {\bf 7.932} & 8.296 & 12.080 \\
Cricket\_Y & 11.901 & 10.150 & 10.265 & {\bf 9.574} & 15.002 \\
Cricket\_Z & 9.714 & 7.760 & {\bf 7.544} & 8.041 & 11.003 \\
DiatomSizeReduction & 0.964 & {\bf 0.852} & 0.874 & 0.869 & 1.203 \\
DistalPhalanxOutlineAgeGroup & 0.403 & 0.206 & {\bf 0.175} & 0.177 & 0.231 \\
DistalPhalanxOutlineCorrect & 0.515 & 0.310 & 0.300 & {\bf 0.262} & 0.351 \\
DistalPhalanxTW & 0.468 & 0.228 & 0.186 & {\bf 0.178} & 0.231 \\
ECG200 & 1.907 & 1.541 & 1.565 & {\bf 1.536} & 1.905 \\
ECG5000 & 4.737 & 4.190 & 4.398 & {\bf 4.148} & 5.463 \\
ECGFiveDays & 1.584 & 1.396 & {\bf 1.322} & 1.335 & 2.220 \\
Earthquakes & 71.461 & {\bf 55.819} & 56.504 & 57.153 & 147.980 \\
ElectricDevices & 19.499 & 15.045 & {\bf 14.999} & 15.228 & 37.121 \\
FISH & 0.439 & 0.353 & 0.319 & {\bf 0.318} & 0.462 \\
FaceAll & 9.309 & 8.687 & {\bf 7.803} & 7.853 & 10.716 \\
FaceFour & 20.483 & {\bf 20.411} & 21.259 & 21.444 & 46.841 \\
FacesUCR & 14.984 & 14.530 & {\bf 14.403} & 14.729 & 28.576 \\
Gun\_Point & 0.447 & 0.368 & 0.300 & {\bf 0.297} & 0.858 \\
Ham & 16.152 & 14.717 & {\bf 12.252} & 13.424 & 24.340 \\
Haptics & 15.177 & 14.275 & 12.394 & {\bf 11.931} & 23.130 \\
Herring & 0.310 & 0.305 & 0.292 & {\bf 0.249} & 0.865 \\
InsectWingbeatSound & 3.104 & 2.346 & 2.186 & {\bf 2.036} & 5.437 \\
ItalyPowerDemand & 0.802 & {\bf 0.595} & 0.623 & 0.654 & 0.881 \\
LargeKitchenAppliances & 61.531 & 63.834 & 59.116 & {\bf 57.219} & 266.853 \\
Lighting2 & 65.602 & 62.240 & 61.561 & {\bf 60.826} & 147.668 \\
Lighting7 & 43.930 & 41.668 & 40.535 & {\bf 39.422} & 68.902 \\
Meat & 0.173 & 0.140 & 0.103 & {\bf 0.077} & 0.099 \\
MedicalImages & 0.932 & 0.671 & {\bf 0.615} & 0.651 & 1.211 \\
MiddlePhalanxOutlineAgeGroup & 0.283 & 0.200 & {\bf 0.134} & 0.154 & 0.312 \\
MiddlePhalanxOutlineCorrect & 0.269 & 0.169 & {\bf 0.133} & 0.148 & 0.182 \\
MiddlePhalanxTW & 0.225 & 0.126 & 0.111 & {\bf 0.094} & 0.132 \\
MoteStrain & 9.704 & {\bf 9.321} & 9.385 & 10.156 & 11.183 \\
NonInvasiveFatalECG\_Thorax1 & 0.921 & 0.736 & 0.540 & {\bf 0.536} & 1.219 \\
OSULeaf & 16.168 & 14.372 & 13.354 & {\bf 12.350} & 30.739 \\
OliveOil & 0.179 & 0.298 & 0.034 & 0.026 & {\bf 0.020} \\
PhalangesOutlinesCorrect & 0.328 & 0.231 & 0.161 & {\bf 0.153} & 0.170 \\
Phoneme & 173.501 & {\bf 158.036} & 159.293 & 158.776 & 219.704 \\
Plane & 0.454 & 0.343 & {\bf 0.311} & 0.370 & 0.630 \\
ProximalPhalanxOutlineAgeGroup & 0.137 & 0.056 & 0.039 & {\bf 0.036} & 0.046 \\
ProximalPhalanxOutlineCorrect & 0.118 & 0.046 & 0.033 & {\bf 0.030} & 0.044 \\
ProximalPhalanxTW & 0.164 & 0.067 & 0.054 & {\bf 0.045} & 0.055 \\
RefrigerationDevices & 107.693 & 95.458 & {\bf 90.328} & 91.687 & 160.361 \\
ShapeletSim & 106.805 & 73.626 & {\bf 72.985} & 73.987 & 108.936 \\
ShapesAll & 11.946 & 7.724 & {\bf 7.127} & 7.522 & 18.348 \\
SonyAIBORobotSurface & 4.068 & 3.619 & {\bf 3.400} & 3.737 & 4.388 \\
SonyAIBORobotSurfaceII & 6.954 & 6.585 & {\bf 6.410} & 6.593 & 8.628 \\
Strawberry & 0.110 & 0.082 & 0.071 & {\bf 0.069} & 0.081 \\
SwedishLeaf & 1.438 & 1.151 & {\bf 1.095} & 1.108 & 1.633 \\
Symbols & 14.717 & {\bf 12.040} & 15.229 & 16.199 & 38.268 \\
ToeSegmentation1 & 24.293 & {\bf 20.905} & 22.914 & 22.566 & 35.806 \\
ToeSegmentation2 & 44.439 & 36.333 & {\bf 35.804} & 41.121 & 61.899 \\
Trace & 0.578 & 0.331 & 0.272 & {\bf 0.250} & 0.582 \\
TwoLeadECG & 0.157 & 0.131 & {\bf 0.129} & 0.149 & 0.336 \\
Two\_Patterns & 14.843 & 11.616 & 11.622 & {\bf 11.059} & 35.923 \\
UWaveGestureLibraryAll & 42.336 & 30.864 & 27.572 & {\bf 26.573} & 93.019 \\
Wine & 0.069 & 0.263 & 0.045 & 0.029 & {\bf 0.028} \\
WordsSynonyms & 12.654 & 10.089 & 9.887 & {\bf 8.946} & 32.003 \\
Worms & 74.589 & 71.946 & 70.245 & {\bf 67.669} & 114.528 \\
WormsTwoClass & 81.311 & {\bf 61.360} & 62.281 & 74.672 & 122.619 \\
synthetic\_control & 7.455 & 5.509 & 5.374 & {\bf 5.369} & 7.690 \\
uWaveGestureLibrary\_X & 14.151 & 10.065 & 9.231 & {\bf 9.197} & 19.995 \\
uWaveGestureLibrary\_Y & 9.852 & 7.285 & 7.066 & {\bf 7.036} & 17.706 \\
uWaveGestureLibrary\_Z & 12.019 & 8.893 & {\bf 8.453} & 8.513 & 20.092 \\
wafer & 1.125 & 0.838 & {\bf 0.765} & 0.818 & 2.636 \\
yoga & 2.510 & 2.209 & {\bf 1.807} & 1.851 & 4.305 \\
\bottomrule
\end{tabular}
\end{table}

\end{document}